\documentclass[11pt]{article}

\usepackage{latexsym,mathrsfs}
\usepackage{amsmath,amssymb} 
\usepackage{amsthm,enumerate,verbatim}
\usepackage{amsfonts}
\usepackage{graphicx}
\usepackage{algorithm}
\usepackage{algorithmic}
\usepackage{url}
\usepackage[dvipsnames]{xcolor}
\usepackage{pifont}
\usepackage{mathtools}
\usepackage{makecell}
\usepackage[normalem]{ulem}
\usepackage{hyperref}
\usepackage{tabularx}
\usepackage{multirow}
\usepackage{lineno}

\newtheorem{theorem}{Theorem}

\newtheorem{definition}{Definition}
\newtheorem*{definition*}{Definition}

\newtheorem{remark}{Remark}

\DeclareMathOperator{\conv}{conv} 
\DeclareMathOperator{\argmax}{argmax} 
\DeclareMathOperator{\argmin}{argmin}

\DeclareMathOperator{\tr}{tr} 
\DeclareMathOperator{\diag}{diag} 
\DeclareMathOperator{\sign}{sign}

\definecolor{brightpink}{rgb}{1.0, 0.0, 0.5}

\title{Revisiting data augmentation for subspace clustering} 
\date{}

\author{Maryam Abdolali$^{*}$ \qquad Nicolas Gillis\thanks{Emails: \{maryam.abdolali, nicolas.gillis\}@umons.ac.be. The authors acknowledge the support by the Francqui Foundation, 
and by the 
Fonds de la Recherche Scientifique - FNRS and the Fonds Wetenschappelijk Onderzoek - Vlanderen (FWO) under EOS Project no O005318F-RG47. 
} \\   
	Department of Mathematics and Operational Research \\ 
	University of Mons, 
	Rue de Houdain 9, 7000 Mons, Belgium
}

\begin{document}

\maketitle

\begin{abstract}
Subspace clustering is the classical problem of clustering a collection of data samples that approximately lie around several low-dimensional subspaces. 
The current state-of-the-art approaches for this problem are based on the self-expressive model which represents the samples as linear combination of other samples. However, these approaches require sufficiently well-spread samples for accurate representation which might not be necessarily accessible in many applications. In this paper, we shed light on this commonly neglected issue and argue that data distribution within each subspace plays a critical role in the success of self-expressive models. Our proposed solution to tackle this issue is motivated by the central role of data augmentation in the generalization power of deep neural networks. We propose two subspace clustering frameworks for both unsupervised and semi-supervised settings that use augmented samples as an enlarged dictionary to improve the quality of the self-expressive representation. 
We present an automatic augmentation strategy using a few labeled samples for the semi-supervised problem relying on the fact that the data samples lie in the union of multiple linear subspaces. Experimental results confirm the effectiveness of data augmentation, as it significantly improves the performance of general self-expressive model.
\end{abstract}

\textbf{Keywords}: subspace clustering, data augmentation, auto-augmentation, sparse representation

\section{Introduction}

With the advancements in data acquisition technology, high-dimensional data is widespread in many areas of machine learning and signal processing. 
Data clustering plays a vital role in analyzing and understanding  high-dimensional data. Many techniques have been proposed for this challenging problem. Their common goal is to aggregate the data into several groups based on a similarity/proximity measure. Among these approaches, subspace clustering (SC) is a prominent clustering approach that groups the data points according to their proximity to underlying latent low-dimensional subspaces~\cite{vidal2011subspace}. 

SC relies on the assumption that the data samples are distributed around multiple low-dimensional subspaces, rather than being uniformly spread across the whole ambient space.
\begin{definition}[Subspace Clustering] \label{def:SC} Let the matrix $X \in \mathbb{R}^{d \times n}$ consist of $n$ samples in $\mathbb{R}^d$ that are drawn from a union of $p$ unknown linear subspaces, $\{\mathcal{S}_i\}_{i=1}^p$, with unknown dimensions,  $\{d_j\}_{j=1}^p$ where $d_j \ll d$ for $j=1,\dots,p$. 
Without any prior information about the orientation or the distribution of subspaces, the goal is to recover the underlying low-dimensional structures by assigning each sample to a subspace/cluster. 
\end{definition}

In the past two decades, a wide variety of SC approaches have been proposed; see \cite{vidal2011subspace, lu2018subspace, rahmani2017innovation, ding2021dual, abdolali2021beyond} and the references therein. 
Among the many proposed techniques, the ones that are based on the concept of \emph{self-expressive representations} are widely accepted to be the current state-of-the-art SC approaches~\cite{elhamifar2013sparse,liu2012robust,soltanolkotabi2012geometric}. Self-expressive representations rely on the fact that each sample can be expressed as a linear combination of other samples from the same subspace, that is, $X(:,j) = \sum_{i \neq j} C(i,j) X(:,i)$, where $X(:,j)$ indicates the $j$-th column of the data matrix $X$, and $C(i,j)$ is the $(i,j)$-th entry of the self-expressive coefficient matrix $C \in \mathbb{R}^{n \times n}$. The nonzero entries in the square matrix $C$ indicate a pairwise affinity between the corresponding samples. 

However, there are typically infinitely many solutions to a self-expressive model, because the number of samples in a low-dimensional subspace is usually larger  than its intrinsic dimension. Ideally, the coefficient matrix should respect the \emph{subspace preserving} property, which requires each sample to be represented by the data points from the \emph{same} subspace. In other words, the entries in the coefficient vector $C(:,j)$ that correspond to the samples from different subspaces should be equal to zero. 
Several regularization functions are proposed in the literature to enforce the subspace preserving property in self-expressive representations; including the sparsity-inducing $\ell_1$ norm~\cite{elhamifar2013sparse}, the low-rank promoting nuclear norm~\cite{liu2012robust}, and the Forbenius norm~\cite{lu2012robust}. 
Once the self-expressive coefficient matrix is computed, the samples are clustered by applying the classic spectral clustering algorithm~\cite{von2007tutorial} on the affinity matrix $A = |C| + |C^\top| \in \mathbb{R}^{n \times n}$. 

Self-expressive based SC algorithms have been extensively studied from a theoretical standpoint~\cite{elhamifar2013sparse,soltanolkotabi2012geometric,you2019affine,li2018geometric,wang2013noisy}, 
and have been used successfully in many machine learning applications~\cite{cui2021new,zhai2016new,breloy2018robust}. 
However, the success of self-expressiveness heavily depends on the existence of well-spread samples in each subspace~\cite{soltanolkotabi2012geometric}. 
To the best of our knowledge, the majority of the literature has overlooked the role of the data distribution within each subspace on the performance of the SC algorithms, and has treated the given data points as a fixed input. 
In fact, the major focus of most SC approaches is to obtain a higher quality coefficient matrix, that is, strengthening the connections between the data points within each subspace while reducing the wrong connections between the samples from different clusters. This was tackled using different regularizations~\cite{lu2013correlation,wang2011efficient,you2016oracle}, post-processing approaches~\cite{yang2019subspace,ji2014efficient,peng2016constructing}, or proposing more robust techniques~\cite{heckel2015robust,lu2013correntropy}. However, the key role of the distribution of the samples across the subspaces  is usually not discussed.   

In this paper, we focus on the role of the data ``quality" and its underlying impact on the ``quality" of the coefficient matrix. Inspired by the remarkable influence of data augmentation on the performance and \emph{generalization ability} of neural networks~\cite{shorten2019survey,xie2020unsupervised}, we employ data augmentation to increase the diversity of the data without collecting new samples.

\paragraph{Contribution and outline of the paper} 

In this paper, we propose a general framework to incorporate data augmentation within SC. First, in Section~\ref{sec:prior work}, we review the self-expressive based model and the existing algorithms which are the foundations of the proposed models. 
Then, the main contributions are presented in the next sections: 
\begin{enumerate}
	\item In Section~\ref{sec:geometry}, we provide a geometric interpretation to illustrate the impact of augmented samples on the performance of SC, with the focus on sparsity regularized SC.
	
	\item In Section~\ref{sec:proposed}, we consider unsupervised and semi-supervised SC, and propose a framework for integrating data augmentation within the general optimization model of SC. 
	We first revisit our prior work~\cite{abdolali2022subspace} and generalize it for three representative self-expressive unsupervised SC algorithms. 
	We then adapt it to the semi-supervised setting with a few labeled samples available. Using the labeled samples, we present an auto-augmentation strategy. 
	We also provide a simple practical strategy to improve the scalability of the proposed approaches for large real-world data sets.
	
	\item In Section~\ref{sec:numerical}, we illustrate the different properties of the proposed approaches on several numerical experiments, and show their  effectiveness compared to the state of the art using several synthetic and real-world data sets. 
\end{enumerate}

This paper is the extension of our recent conference paper~\cite{abdolali2022subspace} which only considers the unsupervised setting with classic augmentation strategies and only for sparsity regularized SC.  

\section{Related Works} \label{sec:prior work}

\begin{figure}[ht!]
	\begin{center}
		\includegraphics[width=1\textwidth]{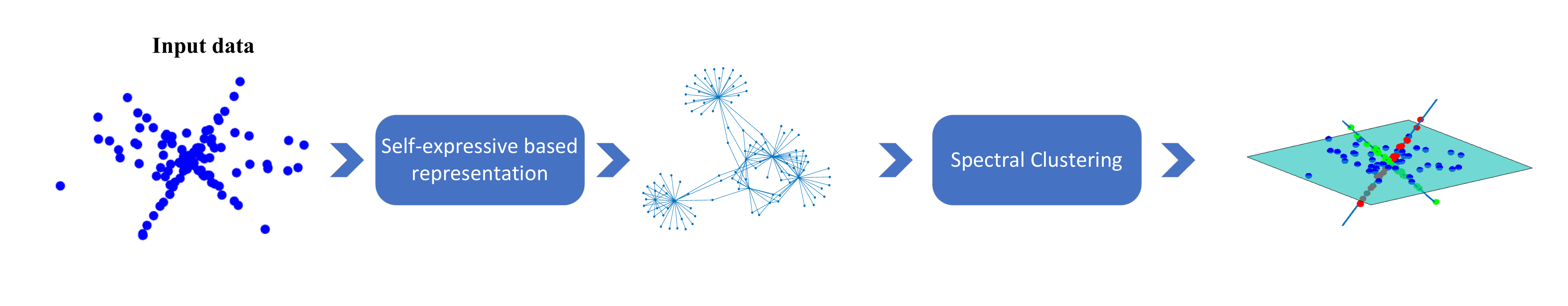}   
		\caption{The overall procedure of self-expressive based SC approaches. In the first step, a pair-wise affinity matrix using self-expressive representations is constructed. In the second step, the clusters are obtained by applying spectral clustering on the affinity matrix.} 
		\label{SC}
	\end{center}
\end{figure}

As illustrated in Figure~\ref{SC}, self-expressive based SC approaches consist of two main steps: 
\begin{enumerate}
	\item Represent the samples using other samples and construct a coefficient matrix, $C$, which encodes the pairwise similarity between samples.
	
	\item Cluster the samples via spectral clustering on the affinity matrix, $A = |C|+|C|^\top$.  
\end{enumerate}

Correct partitioning of samples by spectral clustering depends on the quality of the coefficient matrix computed in the first step. The self-expressive coefficient matrix 
is generally obtained by solving the following optimization problem:
\begin{align}
	\min_{C \in \mathbb{R}^{n \times n}} \  \mathcal{R}(C) + \lambda \ \mathcal{L}(X-XC) \quad \text{such that} \; \ C(i,i) =0 \quad \text{for }i=1,\dots,n, 
\end{align}
where $\mathcal{R}(C)$ is a regularizer, $\lambda >0$ is a regularization parameter, and $\mathcal{L}$ is the loss function which measures the self-expressive representation error and is typically the squared Frobenius norm. 
Three regularizers are the most widely used: 
(1)~the component-wise $\ell_1$ norm, $\|C\|_{\ell_1}=\sum_{i,j}|C(i,j)|$, in sparse subspace clustering (SSC)~\cite{elhamifar2013sparse}, 
(2)~the nuclear norm, $\|C\|_*=\sum_{i=1}^n \sigma_C(i)$ where $\sigma_C(j)$ is the $j$th singular value of $C$, in low-rank representation (LRR)~\cite{liu2012robust}; and 
(3)~the Frobenius norm, $\|C\|_F^2$, in 
least square regression (LSR)~\cite{lu2012robust}. 
Out of the existing regularizers, the $\ell_1$ norm has the strongest theoretical guarantees~\cite{soltanolkotabi2012geometric,wang2013noisy}. In fact, the $\ell_1$ norm based SC is subspace preserving when the underlying subspaces are independent, disjoint and even intersecting/overlapping. Whereas, the subspace preserving guarantees for SC algorithms based on the nuclear and Frobenius norms are limited to independent subspaces~\cite{liu2012robust,lu2012robust}.

\paragraph{Semi-supervised SC} Several approaches in the literature have extended unsupervised SC to the semi-supervised setting where a few labeled samples are available. 
Most approaches use the labels to construct a graph with explicit label smoothness constraint. 
In particular, graph-based label propagation (LP) approaches, such as the Locally and Globally Consistency based method~\cite{zhou2003learning}, are combined with self-expressive based graph construction in SC. The goal is to yield a higher quality graph that respects the label consistency between the adjacent nodes~\cite{zhuang2012non, li2015learning, fang2015robust}. These approaches mainly differ in the regularizers for the coefficient matrix, the label propagation approach, and the optimization scheme; see~\cite{chong2020graph} and the references therein. 

Improving the quality of the coefficient matrix in challenging scenarios such as dealing with data contaminated by noise, outliers or missing entries~\cite{lane2019classifying,heckel2015robust,lu2013correntropy}, 
coming from multiple views/sources~\cite{gao2015multi,kang2020partition}, and 
with nonlinear structures (see~\cite{abdolali2021beyond} and the references therein) have formed the main body of the previous studies on SC. 
However, almost all of the past studies have considered the data matrix as a fixed input and attempted to improve the performance (under various settings and assumptions) without altering the input data. 

In this paper, we focus on the effect of data distribution on the quality of the coefficient matrix. Motivated by the significant role of data augmentation in the performance of neural networks,
we will propose an SC paradigm relying on data augmentation leading to representations that better respect the subspace preserving property.

\paragraph{SC and data augmentation} To the best of our knowledge, there are only a handful number of SC algorithms that incorporate augmented samples for learning the representations. Moreover, these approaches tackle \emph{nonlinear} SC. The common backbone of these approaches is integrating self-expressive representation as a linear layer within a conventional autoencoder network~\cite{ji2017deep}. These approaches typically use data augmentation to obtain consistent representations for different augmentation strategies applied on the data. In other words, the augmented samples are used as \emph{different views} from the given input data and the goal is to achieve consistent representations among these views. For example, 
a subspace consistency loss is proposed in~\cite{abavisani2020deep} to enforce consistent subspaces for original samples and their corresponding transformed samples. Later in~\cite{zhang2021triplet}, the concept of \emph{contrast learning} was brought in neural network based SC and data augmentation was used to generate \emph{positive pairs} that are assumed to share the same subspace/cluster.

In this paper, we will combine the original and augmented samples as a unified enlarged dictionary, and argue that this provides more information to ensure subspace preserving representations. 

\section{Geometric perspective on the impact of data augmentation in sparse subspace clustering} \label{sec:geometry}

In this section, we focus on SSC, and provide geometric intuitions on how data augmentation affects the performance of SC. 
Although the focus of the study in this section is on SSC, we will numerically show in Section~\ref{sec:numerical} that data augmentation is beneficial for other self-expressive based SC approaches, namely LSR and LRR.

\subsection{The effect of data augmentation on subspace preserving guarantees}

Let us first review the conditions under which SSC obtains subspace preserving coefficients, for which the arrangement of the subspaces plays a key role \footnote{For a summarization of subspace preserving guarantees for different SC algorithms see Table~1 in~\cite{abdolali2021beyond}.}. 
For the simplest subspace arrangement, that is, \emph{independent} subspaces, SSC is always subspaces preserving~\cite{elhamifar2013sparse,soltanolkotabi2012geometric}. 
For more complex subspace arrangements, two notions play a crucial role:  
\emph{subspace incoherence} and \emph{inradius}.
Subspace incoherence is a measure of the separation amongst the subspaces. 
As the angle between subspaces decreases, the incoherence increases, and, subsequently, the SC problem becomes more challenging. Given a data point, its corresponding subspace incoherence, denoted $\mu(X(:,j)) \in [0,1]$, 
provides a ``distance'' between $X(:,j)$ and the subspaces it does not belong to.  
In particular, $\mu(X(:,j)) = 1$ if the subspaces are orthogonal to $X(:,j)$ while $\mu(X(:,j)) = 0$ if $X(:,j)$ belongs to one of these subspaces.
For completeness, a rigorous definition of subspace incoherence, based on the dual of the SSC optimization problem, is provided in the supplementary material~\ref{suppl:defpres}. 
For a given convex hull $\mathcal{P}$, its inradius, denoted $r(\mathcal{P})$, is defined as the largest Euclidean ball that is inscribed in $\mathcal{P}$. 
After the normalization of the data points in a subspace, the corresponding inradius  captures how well the samples are spread in a subspace,  in terms of spanning the different dimensions.   
Given these two concepts, we have the following theorem that guarantees SSC to be subspace preserving. 
 \begin{theorem}[Subspace preserving condition for SSC~\cite{soltanolkotabi2012geometric}]
 	Suppose all the samples are normalized to have unit $\ell_2$  norm, that is, $\|X(:,i)\|_2 = 1$ for all $i$. 
 	The optimal solution to 
 	\begin{align} \label{SSC}
	C(:,j) \ = \ \argmin_{c\in \mathbb{R}^n} \|c\|_1 \quad \text{such that} \; \; X(:,j) = Xc \; \text{and} \; c(j)=0, 
 \end{align} 
 	is subspace preserving for the sample $X(:,j) \in \mathcal{S}_\ell$ if the following condition holds:
 	\begin{align} \label{ssc_condition}
 		\mu(X(:,j)) \; < \; r(\mathcal{P}^\ell_{-j}),
 	\end{align}
 	where $\mathcal{P}^\ell_{-j}$ is the convex hull of the samples in the $\ell$-th subspace without $X(:,j)$. %and $\mu(X(:,j))$ is the subspace incoherence of the sample $X(:,j)$ with respect to samples from other subspaces.
 \end{theorem}

%According to this condition, there are two important factors for the success of SSC. The subspace incoherence quantifies the closeness between the dual direction of a sample and the samples from the other subspaces. 
Based on this inequality, as the incoherence increases, the inradius should increase as well to obtain a subspace preserving solution. 
Hence, as the latent subspaces get closer to each other, the value of inradius plays a significant role in the success of SSC. 
However, note that the subspace incoherence relies on the intrinsic structure of the data and changing this quantity with no prior information on the latent structure of samples in the high-dimensional space is nontrivial (if not impossible). 
But as we will argue in this paper, augmenting the samples can increase the  inradius, and hence benefits SSC in the challenging case of clustering nearby subspaces.

\subsection{The effect of data augmentation on the coefficient matrix}

In this section, we present the geometric interpretation on how changing the inradius using data augmentation can enhance SSC. 
Let us consider the geometric interpretation of SSC, which was first reported in the compressed sensing literature~\cite{chen2001atomic,nasihatkon2011graph}. 

Let ${X^{\pm}_{-j}=[\pm X(:,1),\dots,\pm X(:,j-1),\pm X(:,j+1),\dots,\pm X(:,n)] \in \mathbb{R}^{d \times 2n-2}}$ be the matrix which is obtained by combining the negative samples with the entire original samples in $X$, except the $j$-th sample, $X(:,j)$. Using $X^{\pm}_{-j}$ as the dictionary,~\eqref{SSC} can be reformulated as: 
\begin{align} \label{SSC2}
	\min_{a \in \mathbb{R}^{2n-2}} e^\top a \quad \text{such that} \; \;  X(:,j)=X^{\pm}_{-j} a, \ a \geq 0,
\end{align}
where $a \geq 0$ indicates the element-wise nonnegativity constraint, and $e$ is the vector of all ones of appropriate dimension.   
The first constraint in~\eqref{SSC2} can be written as  
\[
X(:,j)=X^{\pm}_{-j} a = X^{\pm}_{-j} \underbrace{\frac{a}{e^\top a}}_{b} \ e^\top a,\]
where $b = \frac{a}{e^\top a} \in \mathbb{R}^{2n-2}$. The vector $X^{\pm}_{-j} b$ belongs to 
the convex hull of the set of points in $X^{\pm}_{-j}$, that is, $\conv(X^{\pm}_{-j})$, and by setting $\beta = \frac{1}{e^\top a}$, we write the problem in~\eqref{SSC2} as:
\begin{align}
	\max_\beta \ \beta \quad \text{such that} \; \; \beta X(:,j) \in \conv(X^{\pm}_{-j}). 
\end{align}
This means that~\eqref{SSC2} looks for the vectors with the maximum length on the ray generated by $X(:,j)$ and that belongs to $\conv(X^{\pm})$. The optimal solution will belong to a face of $\conv(X^{\pm})$, and the vertices of that face will correspond to the samples with nonzero entries in the sparse coefficient vector $C(:,j)$. 
The distribution of samples within each subspace influences the convex hull of the samples, and hence the inradius of the subspace. 
Changing the distribution of data points within each subspace affects the nonzero entries of the coefficient matrix, and, consequently, might affect the correctness of the clustering. 

\begin{figure*}[!htbp]
	\begin{minipage}[b]{0.5\linewidth}
		\centering
		\centerline{\includegraphics[width=7.5cm]{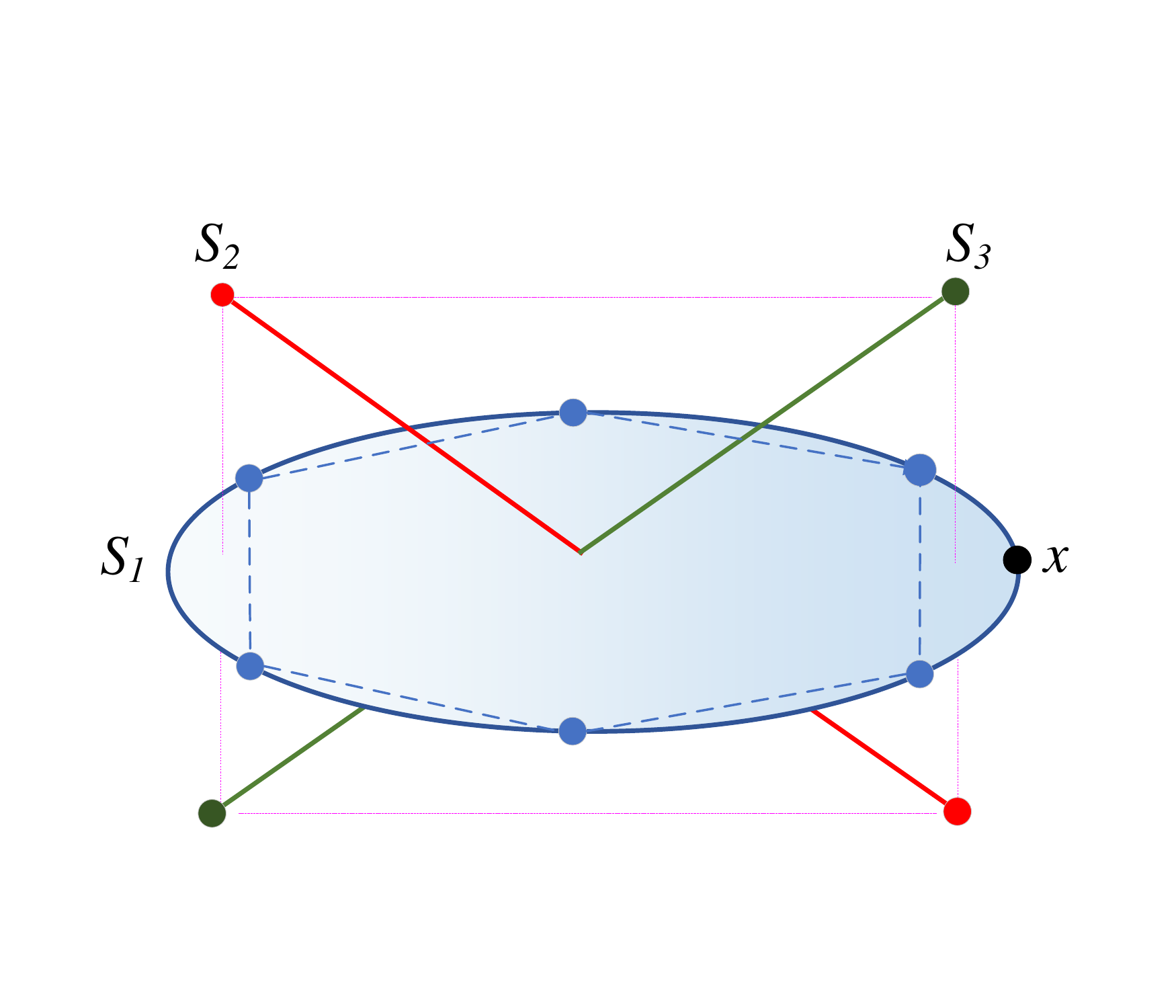}}
		%  \vspace{1.5cm}
		\centerline{(a) without data augmentation}\medskip
	\end{minipage}
	\hfill
	\begin{minipage}[b]{0.5\linewidth}
		\centering
		\centerline{\includegraphics[width=7.5cm]{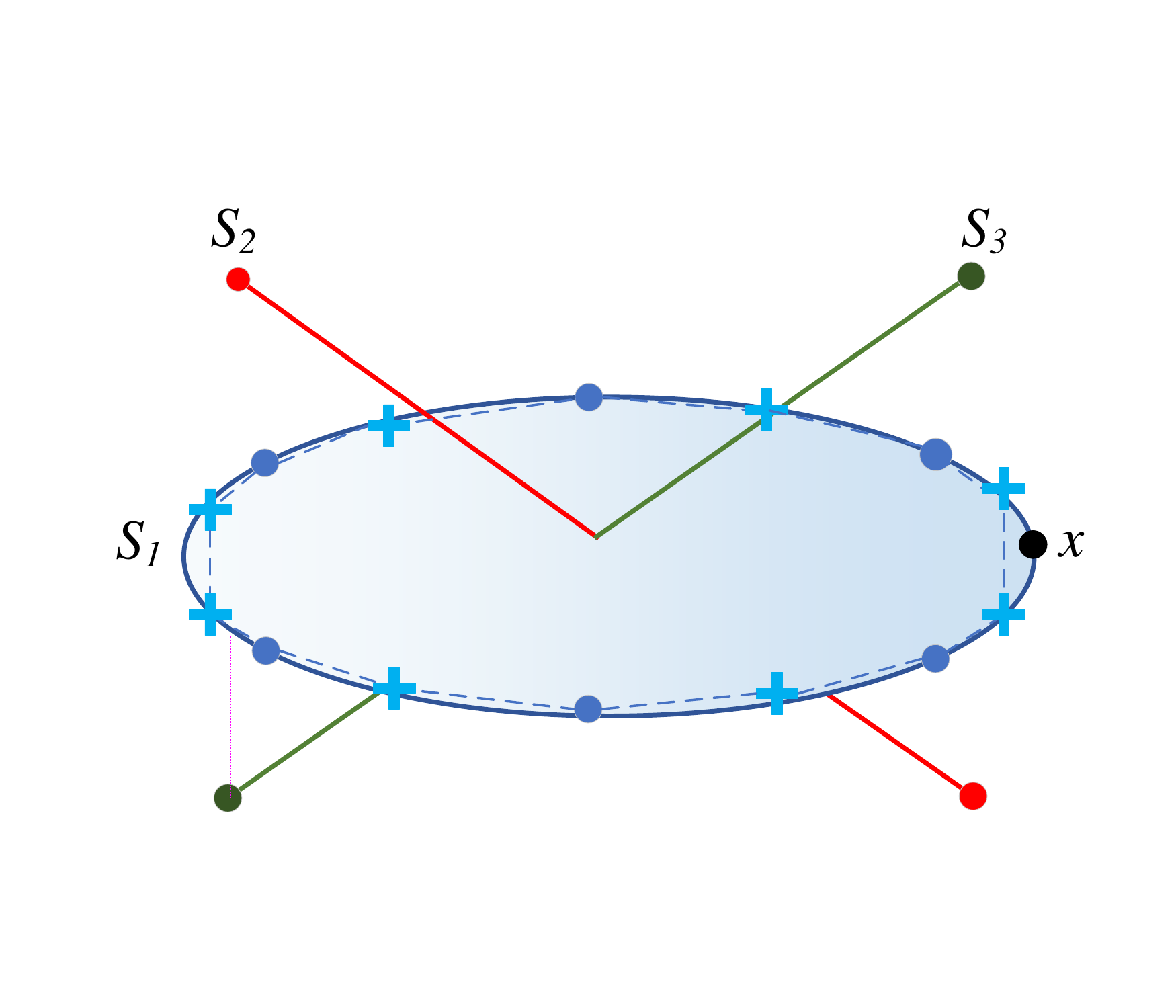}}
		%  \vspace{1.5cm}
		\centerline{(b) with data augmentation}\medskip
	\end{minipage}
	\caption{The effect of augmentation on the correctess of SSC. 
	With no data augmentation, see (a), the sample $x \in \mathcal{S}_1$ (black dot) is represented by the samples from the wrong subspaces, $\mathcal{S}_2$ and $\mathcal{S}_3$. Adding augmented samples (crosses) to the given samples, see (b), allows $x$ to be represented by the augmented samples from the same subspace.} 
	\label{aug_illus}
\end{figure*}

Let us show this using a simple example in three dimensions. 
Suppose we are given samples from three disjoint subspaces,  $\{\mathcal{S}_i\}_{i=1}^3$, in three dimensions ($d=3$); one plane and two lines; see Figure~\ref{aug_illus}~(a). 
The samples are shown with full circles. Let $x \in \mathcal{S}_1$ (black circle) be a sample that lies on the intersection of $\mathcal{S}_1$ and $\mathcal{S}_2 \bigoplus \mathcal{S}_3$, where $\bigoplus$ indicates the sum of two subspaces. 
The ray generated by $x$ intersects the boundary of the convex hull of the other  samples, $\conv(X^\pm_{-x})$, on a face with vertices belonging to the subspaces $\mathcal{S}_2$ and $\mathcal{S}_3$. Hence, the nonzero entries in the coefficient vector corresponding to $x$ correspond to samples from wrong subspaces. Now, suppose that using an augmentation strategy a few new samples (shown by cross signs) are generated and added to the existing samples; see  Figure~\ref{aug_illus}~(b). Now, the ray generated by $x$ intersects the convex hull on the face with vertices belonging to the same subspace, $\mathcal{S}_1$. 
This is, in fact, due to the increase in the value of the inradius of the subspace $\mathcal{S}_1$. 

\section{Subspace clustering with data augmentation} \label{sec:proposed}

In the next section, we explain how we integrate the the augmented samples within the general optimization model of SC. 

Data augmentations are widely used in the neural network literature as an explicit regularization technique to improve the generalization capability and reduce overfitting. 
Augmentation refers to synthetic data generation strategies that are applied on a given existing data set in order to increase the number of available samples and the quality/diversity of the data. The major common assumption for augmentation strategies across different data domains is that they should be \emph{label preserving}. An augmentation transformation is said to be label preserving if it does not change the cluster (label) of the generated samples. 
Examples of classic data augmentation strategies for images include primitive image processing functions such as flipping, rotation, scaling and  color space shifting. 

Inspired by the significant role of augmentation in training neural networks with limited samples for various domains, we revisit this concept for the problem of SC and use data augmentation to avoid degenerate subspaces and generate well-spread samples within each subspace. 
In this section, we first focus on the more direct approach of applying classic predefined augmentation strategies in the unsupervised scenario. In section~\ref{sec:a-sc}, we propose a framework, dubbed augmented subspace clustering (A-SC) that benefits from a synthetically enlarged data set. 
We extend this framework in Section~\ref{sec:as-sc} in the semi-supervised scenario based on label propagation and argue that the increase in the quality of the coefficient matrix due to the augmented data can benefit the label propagation. Moreover, under the model of union of multiple subspaces, we propose a strategy to \emph{automatically} generate augmented samples. Hence, we provide a semi-supervised methodology that uses few labeled samples to both \emph{learn} the augmented samples and \emph{refine} the connectivity graph in an iterative algorithm.

\subsection{Unsupervised subspace clustering with data augmentation}\label{sec:a-sc}

In this section, we focus on unsupervised data augmentation in order to increase the diversity of given samples by forming an enlarged overcomplete dictionary. This approach, named A-SC, is the extension of our prior work recently presented in \cite{abdolali2022subspace}. We generalize this work for the three most common regularizations for the coefficient matrix $C$ and introduce a general framework for combining self-expressive based SC approaches with synthetic augmented data.

A-SC relies on unsupervised \emph{instance-based} data augmentation. For instance-based strategies, the transformation function is applied on each sample \emph{individually} and \emph{independently}. Examples of such strategies are classic image augmentation techniques such as flipping, scaling and rotations. Suppose we are given $m$ different augmentation functions that are assumed to be label preserving. 
Let $\hat{X} = [\hat{X}_1,\dots,\hat{X}_m] \in \mathbb{R}^{d \times nm}$ be the $m$ sets of augmented samples, where $\hat{X}_j \in \mathbb{R}^{d \times n}$ is the synthetic samples generated by applying the $j$-th transformation function on the data matrix $X$, for $j=1,\dots,m$. 
The matrix $\hat{X}$ contains extra information that is implicitly available in the original data $X$. The final overcomplete dictionary, $\tilde{X}$, is formed by concatenating original samples with the augmented ones as follows: 
$\tilde{X}=[X \ | \ \hat{X}] \in \mathbb{R}^{d \times (nm+n)}$. Using the newly available information in $\tilde{X}$,we propose the following optimization model: 
\begin{equation}\label{eq:A-SC} 
	\tilde{C} \quad = \quad  
	\underset{C \in \mathbb{R}^{n(m+1)\times n}}{\argmin} \ \mathcal{R}(C)+\frac{\lambda}{2} \ \|X-\tilde{X}C\|_F^2 
	 \quad \text{such that} \; \; C(j,\Omega(j))=0 \ \text{ for } \ j=1,\dots,n, 
\end{equation}
where $\Omega(j)$$=$$\{j+(k-1)n\}_{k=1}^{m+1}$ contains the indices of the augmented samples corresponding to the sample $X(:,j)$. The constraint $C(j,\Omega(j))=0$ makes sure that each sample is not represented using itself or its corresponding augmented samples. In fact, this is crucial to avoid the possible \emph{oversegmentation} problem, that is, segmenting a cluster (subspace) into several excessive mini-clusters. Representing a given sample $X(:,j)$ only by its corresponding augmented samples reveals nothing about the relationship and connectivity between the samples in the original data matrix. 
%If no other sample uses the particular sample $X(:,j)$ for its representation, this will isolate the node corresponding to $X(:,j)$. Hence this results in a oversegmentation of the graph with $X(:,j)$ as a individual cluster containing only a single node/sample. \ngc{This is not specific to our model, I think this is rather a distraction for the reader} 
We consider three representative regularization functions for $\mathcal{R}(C)$, namely, $\|C\|_1$, $\|C\|_*$ and $\|C\|_F^2$ which eventually leads to three algorithms, Augmented SSC (A-SSC), Augmented LRR (A-LRR) and Augmented LSR (A-LSR), respectively. 

For obtaining the clusters by spectral clustering, we take advantage of the specific block-wise structure in the rectangular coefficient matrix $\tilde{C}$. In particular, the matrix $\tilde{C}$ is composed of vertical concatenation of several square matrices: $\tilde{C}= \ [\tilde{C}_1 \ ; \ \dots \ ; \ \tilde{C}_{m+1}]$ where $\tilde{C}_1 \in \mathbb{R}^{n \times n}$ corresponds to the representation coefficients using original samples in $X$, and, similarly, $\{\tilde{C}_j\}_{j=2}^{(m+1)} \in \mathbb{R}^{n \times n}$ indicates the coefficient representations using the augmented samples in $\{\tilde{X}_j\}_{j=1}^m$. 
We therefore summarize the matrix $\tilde{C}$ into a square matrix $C_f \in \mathbb{R}^{n \times n}$ as follows: 
$C_f = \sum_{i=1}^{m+1} \ |\tilde{C}_i|$. 
Lastly, the final clusters are obtained by applying spectral clustering on the  affinity matrix $A_f \ = \ |C_f| \ + \ |C_f^\top|$. 

However, increasing the number of augmented samples is a double edged sword. On one hand, it can potentially improve the performance, due to the increase in the inradius of the subspaces, see~\eqref{ssc_condition}. 
On the other hand, it also increases the computational cost. 
Solving~\eqref{eq:A-SC} using ADMM~\cite{gabay1976dual} (which is standard and efficient strategy which we adopt in this paper) costs  $O(\tilde{n}^3)$ operations which might limit the applicability of A-SC for real-world data sets, and limits the number of augmented samples. 

Fortunately, the proposed formulation in~\eqref{eq:A-SC} allows us to use several already available scalable SC algorithms~\cite{you2016scalable, abdolali2019scalable, chen2020stochastic}. 
For simplicity, in this paper we use a simple strategy proposed in \cite{zhuang2016locality}: instead of representing every sample $X(:,j)$ as a linear combination of the entire large dictionary, $\tilde{X}$, it uses its k-nearest neighbors (kNN). In other words, the self-expressive dictionary is limited to the $k$ samples that are closest to the sample $X(:,j)$ (discarding its augmented samples). 
The corresponding class of models is dubbed augmented kNN SC (Ak-SC), while we denote Ak-SSC, Ak-LRR and Ak-LSR, the SC algorithms for the corresponding regularizers.

\subsection{Semi-supervised subspace clustering with data augmentation}\label{sec:as-sc}

In many applications, there are often a few labeled samples available among the many unlabeled ones. Extracting the latent low-dimensional subspaces by taking advantage of the existing labeled samples is the goal of semi-supervised SC algorithms. The supervisory information from the few labeled samples can be beneficial in learning a higher quality coefficient matrix. %and subsequently more subspace preserving connections between the samples. 
In this section, we first introduce the proposed framework for combining the augmented samples within the semi-supervised SC problem, and then we suggest a possible auto-augmentation strategy according to the model of union of multiple subspaces. We eventually discuss the shortcoming of this strategy for extracting nonlinear structures.

\subsubsection{Combining data augmentation and label propagation in semi-supervised SC}

Our proposed approach will use the 
the information from the labeled and augmented samples in order to 
(i)~construct a more accurate coefficient matrix $C$,
and 
(ii)~propagate the available label information across the graph {corresponding to the coefficient matrix $C$}. 

Let the data $X \in \mathbb{R}^ {d \times n}$ be divided into labeled samples $X_l \in \mathbb{R}^{d \times n_l}$ and unlabeled samples $X_u \in \mathbb{R}^{d \times n_u}$, such that $n_l + n_u = n$. 
For simplicity, we assume that there are equal number of labeled samples for each category, namely, $\frac{n_l}{p}$. Let $Y = \{0,1\}^{n \times p}$ be the label indicator matrix which contains the available labels of the samples and is defined as follows: 
\[Y(i,j)= \left\lbrace \begin{array}{lc}
	1, & \ \text{if $X(:,i)$ is a labeled sample from the $j$th subspace},\\
	0, &  \ \text{otherwise}.
\end{array}\right.
\] 
The goal of the proposed semi-supervised approach, referred to as Augmented semi-supervised SC (AS-SC) is to achieve a \emph{label-consistent} coefficient matrix using the augmented samples. A coefficient matrix is label consistent if for any two samples with different labels, the corresponding entry in the coefficient matrix is zero. Let $F \in [0,1]^{\tilde{n} \times p}$ be the estimated membership of the  $\tilde{n}$ samples to the $p$ clusters, where the entries in each each row sum to one. Each row of $F$ %represents a probability vector which 
encodes the estimated probability of the corresponding sample belonging to each of the $p$ clusters, that is, $Fe=e$ and $F \geq 0$; 
see Appendix~\ref{app:F_stochastic}.  
%\ngc{This makes sense, but we do not enforce these constraints in the optimization problem... Should we discuss that? My intuition tells me that $F$ eventually satisfies this constraints because of the regularization terms, right?}
%\ngc{F is not defined... should it be binary with one non-zero per column? In theory yes, but this is not enforced in the optimization so this might be confusing... We should define and clarify} 
The matrix $\tilde{Y} \in \{0,1\}^{\tilde{n} \times p}$ is the extended label matrix which is constructed by zero-padding the original label matrix: $\tilde{Y}=[Y;0_{(\tilde{n}-n)\times p}]$ where $0_{(\tilde{n}-n)\times p}$ is zero matrix of dimension $(\tilde{n}-n)\times p$. Moreover, we keep the relationship between augmented samples and their corresponding original samples in the binary matrix $S \in \{0,1\}^{\tilde{n} \times n}$, defined as: 
\begin{equation} \label{define_S}
	S(i,j)= \left\lbrace \begin{array}{lc}
	1, & \ \text{ if } \tilde{X}(:,i) \text{ is an augmented sample generated from the sample }X(:,j),\\
	0, &  \ \text{otherwise}.
\end{array}\right.
\end{equation}
Using the generated augmented samples in combination with label information, AS-SC optimizes the following model: 
\begin{align}\label{eq:AS-SC}
	(\tilde{C},F)=\argmin_{C,F} & \ \mathcal{R}(C) \ + \ \frac{\lambda}{2} \underbrace{\|X-\tilde{X}C\|_F^2}_{\text{loss function}} + \underbrace{\lambda_2 \sum_{i=1}^{\tilde{n}} \sum_{j=1}^n \|F(i,:)-F(j,:)\|_2^2 \ |C(i,j)|}_{\text{label consistency}} \nonumber \\
	&+ \underbrace{\gamma_1 \ \tr\left((F-\tilde{Y})^\top U (F-\tilde{Y})\right)}_{\text{preserving initial labels}} + \underbrace{\gamma_2 \ \sum_{i=1}^{\tilde{n}} \sum_{j=1}^n \|F(i,:)-F(j,:)\|_2^2 \ S(i,j)}_{\text{label preserving augmentation}}, \nonumber \\ 
	\text{such that} & \ C(\Phi_j,j)=0 \ \text{ for } \ j=1,\dots,n, \text{ and } 
	Fe=e, F \geq 0, 
\end{align}
where $\Phi_j$ is a set containing the \emph{cannot-links} information, that is, the edges which should not exist for the sample $X(:,j)$. In particular, $\Phi_j$ contains: (i) the indices of labeled samples with different labels from the sample $X(:,j)$ to avoid connection between samples from different clusters, (ii) the index $j$ and the indices of the corresponding augmented samples from the sample $X(:,j)$ to avoid trivial connections.
The diagonal binary matrix $U \in \{0,1\}^{\tilde{n} \times \tilde{n}}$ is defined as:
\begin{equation}\label{define_U}
	U(i,i)= \left\lbrace \begin{array}{ll}
		1, & \ \text{if $\tilde{X}(:,i)$ is labeled},\\
		0, &  \ \text{otherwise}.
	\end{array}\right.
\end{equation} 
There are three additional vital penalty terms in~\eqref{eq:AS-SC} compared to the unsupervised problem in~\eqref{eq:A-SC}: 
\begin{itemize}
	\item The term $\sum_{i=1}^{\tilde{n}} \sum_{j=1}^n \|F(i,:)-F(j,:)\|_2^2 \ |C(i,j)|$ plays the role of enforcing the consistency between the coefficient matrix $C$ and the estimated label matrix $F$. 
	If two samples, say $\tilde{X}(:,i)$ and $\tilde{X}(:,j)$, have different estimated labels, then the corresponding entry in the coefficient matrix, $C(i,j)$, is encouraged to have a small value. This encourages the removal of connections/links that do not follow the label information 
	within the matrix $F$. 
	Similarly, the matrix $F$ is enforced to follow the implicit graph structure in the coefficient matrix $C$. In particular, large values in the entries of the matrix $C$ encourage the labels of the corresponding samples to be close to each other in the matrix $F$. In other words, this term enforces \emph{label smoothness} over the graph structure induced by $C$. 
	
	\item  The term $\tr\big((F-\tilde{Y})^\top U (F-\tilde{Y})\big)$ ensures that the estimated labels in $F$ are equal to the initial given labels in the matrix $\tilde{Y}$. This term, in combination with the term above, is known as \emph{locally and globally consistent} based label propagation~\cite{zhou2003learning} in the semi-supervised learning literature (see also Section~\ref{sec:prior work}).  
	
	\item The term $\sum_{i=1}^{\tilde{n}} \sum_{j=1}^n \|F(i,:)-F(j,:)\|_2^2 \ S(i,j)$ ensures that the augmented samples share the same label with the  original samples from which they were generated. 
	%In other words, if $\tilde{X}(:,i)$ is an augmented sample generated from the sample $\tilde{X}(:,j)$, they should have the same labels (estimated in $F(:,i)$ and $F(:,j)$).
	
\end{itemize}

%The matrix $F$ in each iteration can be interpreted as the current \emph{confidence} over the estimated labels. In fact, the matrix $F$ is an stochastic matrix with entries between zero and one while the sum of the entries in each row is equal to one. The closer the entries are to one, the higher the confidence over the estimated labels are. 
The regularization parameters $\gamma_1 \text{ and } \gamma_2 >0$ are set to a high value, since we want to preserve the given label information. We have used $\gamma_1=\gamma_2=1000$ for all the experiments in this paper.

\begin{remark}[Avoiding spectral clustering]
	The square coefficient matrix $C_f \in \mathbb{R}^{n \times n}$ can be calculated from the matrix $\tilde{C}$ using $C_f(i,j)=\sum_{k \in \Omega(i)} |\tilde{C}(k,j)|$, for $i,j=1,\dots,n$ where $\Omega(i)$ contains the indices of the augmented samples from the sample $X(:,i)$, as in Section~\ref{sec:a-sc}. One can then apply 
	 spectral clustering on the affinity matrix corresponding to $C_f$ to obtain the clustering labels. 
	 However, for semi-supervised SC, we can avoid the computationally expensive spectral clustering step by using the estimated label matrix $F$ for identifying the final labels. In other words, the label of each sample $X(:,j)$ can be obtained using $\argmax F(j,:)$ for $j=1,\dots,n$. We have used this approach in all numerical experiments of the paper. 
\end{remark}

\begin{remark}[Paths in the induced graph] \label{graph_powers}
	Ideally, there should be $p$ separated connected components in the graph induced by $C_f$, and labeled samples of each cluster should be separated from other samples with different labels. 
This means that there should be no connecting ``path" in the graph between the labeled samples of different clusters. 
It is possible to penalize such paths using ``graph powers"~\cite{bondy2008graph}, but this is rather computationally heavy. 
However, we can interpret the Locally and Globally consistent-based label propagation as an efficient enforcement of one-hop level consistency between the (estimated) labels of the nodes in the graph.
\end{remark}

\paragraph{Solving the AS-SC model~\eqref{eq:AS-SC}}

The optimization problem in~\eqref{eq:AS-SC} is nonconvex. 
We use a two-block coordinate decent approach to solve it: 
We iteratively optimize over one of the two matrices, $C$ or $F$, while keeping the other fixed. 
We use ADMM to estimate $C$ for $F$ fixed, while the optimal $F$ is the solution of a linear system of equations for $C$ fixed.   
%(i) first using ADMM, we optimize the problem with respect to the matrix $C$, while keeping the matrix $F$ fixed, (ii) second, by assuming the matrix $C$ as fixed, the matrix $F$ is updated by solving a linear system of equations. T
This iterative algorithm is terminated when the matrix $F$ does not change considerably between two consecutive iterations, or when the number of iterations exceeds a predefined value (we set it as 10 in this paper as this iterative process converges rather quickly; see Section~\ref{sec:syntexp} for numerical experiments). 
 We have observed in our numerical experiments that updating $F$ and $C$ simultaneously in a unified ADMM algorithm, as in~\cite{wang2018unified}, is very sensitive to the parameter of $\lambda_2$ which controls the label propagation speed. We use zero matrices to initialize both matrices of $F$ and $C$. 
Since our alternating scheme first optimizes $C$, this means that the value of $C$ after the first iteration is the coefficient matrix that would be obtained without using the label information. The details can be found in Appendix~\ref{app:as-ssc}.  

 To reduce the computational burden of using augmented samples, we use the same strategy as unsupervised A-SC, namely using the $k$ nearest neighbors of each sample. 
 This leads to the a class of algorithms dubbed as Augmented kNN Semi-supervised Subspace Clustering (AkS-SC). 
 Depending on the regularizer used for $\mathcal{R}(C)$, we have three algorithms: AkS-SSC, AkS-LRR and AkS-LSR. 
 The detailed optimization procedures for these three algorithms are provided in the supplementary material~\ref{suppl:optsemiAkS-SC}.

\subsubsection{Generating augmented samples within the union of multiple subspaces}

There is a large body of work devoted to designing better and richer augmentation strategies and a wide variety of strategies are available for data augmentation in various domains~\cite{perez2017effectiveness,shorten2019survey,shorten2021text}. As we discussed earlier, augmentation strategies should not change the category (subspace) of the samples but with not enough prior knowledge, a critical question remains: 

\emph{Which strategies should be selected for data augmentation in different domains and applications, and how the parameters of these strategies should be set to generate label-preserving samples?} 

Hence, a big obstacle in using data augmentation for SC is generating the augmented samples that are label preserving. In this section, we propose an auto-augmentation approach under the union of multiple subspaces assumption which we refer to as  ``linear interpolation". 

As the augmentation should not alter the category of samples, the augmented samples should belong to one of the subspaces. 
Hence, for each cluster, we generate the new samples by \emph{linear combinations} of given labeled samples from the same cluster. 
Let $\bar{X} = [\breve{X}_1,\dots,\breve{X}_p] \in \mathbb{R}^{d \times (pn_a)}$ be the generated auto-augmented samples, where $n_a$ is the number of augmented samples per cluster. In particular, for the subspace $S_j$, $n_a$ augmented samples are generated as follows: for $k=1,\dots,n_a$, 
\[ 
\breve{X}_j(:,k)= \{Xa \ | \ a(i) = 0 \ \text{ when } Y(i,j) \neq 1\} , 
\]  
where $a \in \mathbb{R}^n$ is a vector with randomly generated entries (see below for more detail). 
As the augmented samples are generated using the labeled samples from the same cluster, they are guaranteed to belong to the same subspace, and hence this is a label-preserving augmentation strategy under the SC model (Definition~\ref{def:SC}).   
There are several options for randomly generating $a \in \mathbb{R}^n$. For each cluster, we first (randomly) select $q$ samples among $\frac{n_l}{p}$ labeled samples, where $2 \leq q \leq \frac{n_l}{p}$. The nonzero entries are then randomly generated following two strategies: 
\begin{itemize}
	\item Uniform distribution in the interval $[0,1]$ where the vector $a$ is normalized to have unit $\ell_1$ norm. %, that is, $\sum_j a(j)=1$. 
	This augmentation strategy is useful for the applications where the samples are nonnegative, so that the augmented samples are nonnegative as well. 
	
	\item Gaussian distribution of mean 0 and variance 1. This random number generation is useful where the data is not necessarily nonnegative and spans the entire low-dimensional subspace.
\end{itemize} 
The parameter $q$ controls the \emph{locality} of the augmented samples. 
A smaller $q$ might generate more realistic samples (to humans) in data domains such as image data. On the other hand, setting $q$ as the maximum possible value,  $\frac{n_l}{p}$, generates samples that better cover the intrinsic low-dimensional subspace.

\subsubsection{Non-linear SC and manifold intrusion} \label{sec:manfold_intrusion}

The linearity assumption in self-expressive based SC approaches might be restrictive in many real-world applications where the data lies on nonlinear low-dimensional manifolds. 
Fortunately, focusing on \emph{nearby} samples for the representations using kNN not only reduces the computational cost, but also leads to \emph{locality preserving} representations. 
This is an important concept in analyzing smooth nonlinear manifolds in both the single and  multiple manifold learning literature; see, e.g., ~\cite{abdolali2021beyond} and the references therein. %Approaches relying on kNN are reliable to preserve many of the existing structures. 
Hence, our proposed kNN-based algorithms can also be used for clustering nonlinear manifolds. 

\begin{figure*}[!ht]
	\begin{minipage}[b]{0.5\linewidth}
		\centering
		\centerline{\includegraphics[width=10cm]{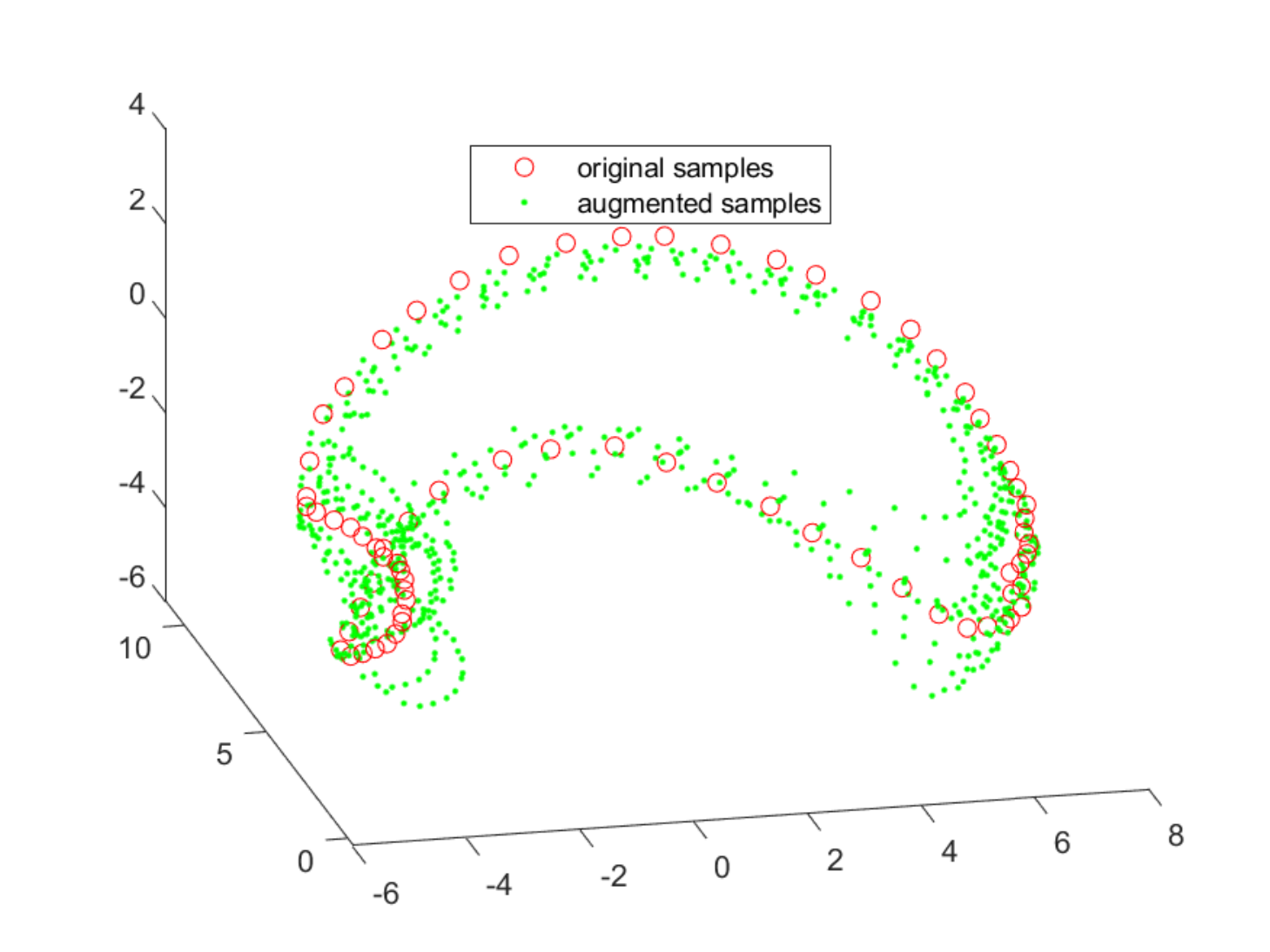}}
		%  \vspace{1.5cm}
		\centerline{(a) Using classic strategies.}\medskip
	\end{minipage}
	\hfill
	\begin{minipage}[b]{0.5\linewidth}
		\centering
		\centerline{\includegraphics[width=10cm]{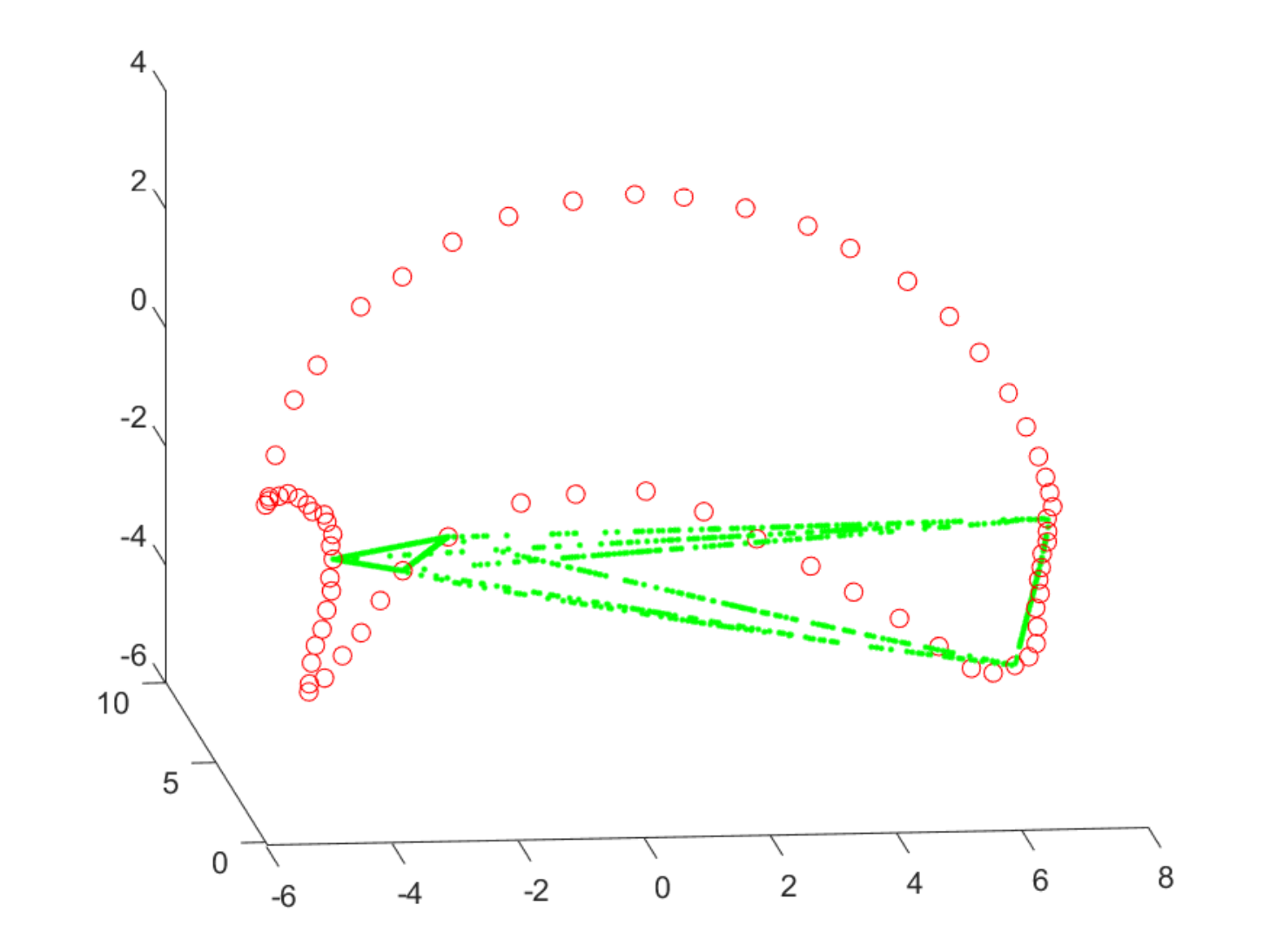}}
		%  \vspace{1.5cm}
		\centerline{(a) Using linear interpolation.}\medskip
	\end{minipage}
	\caption{Illustrating the effect of nonlinearity in generating augmented samples for COIL-20.}
	\label{fig:manifold_intrusion_coil20}
\end{figure*}

Under the SC model, 
generating augmented data by \emph{linear interpolation} will better cover the  subspaces, and hence lead to larger inradiuses. 
However, in the absence of the linearity assumption for the underlying subspaces, linear interpolation might not be desirable and can give rise to a potential phenomenon known as ``manifold intrusion", that is, generating out-of-distribution samples~\cite{guo2019mixup}.
Figure~\ref{fig:manifold_intrusion_coil20} illustrates this phenomenon for samples from a real-world dataset. This figure displays 3-dimensional projection of samples from the first category of the widely used COIL-20 data set which contains images of various objects taken from different angles; see
Section \ref{sec:numerical} for more details. We generate augmented samples following two different schemes: (i) classical instance-based image transformation using rotation and scaling and, (b) linear combination of five labeled samples using linear interpolation based augmentation. For the first scheme, the augmented samples are produced by applying five random strategies on each individual sample, and for the second scheme 100 augmented samples are generated by convex linear combination of pairs of labeled samples. In these figures, the original samples are plotted with red circle and the augmented samples are shown with green dots. Comparing these two figures, we observe that carefully selected classic augmentation strategies can generate samples that \emph{follow} the intrinsic manifold structure accurately. However, generating samples using linear interpolation for nonlinear manifolds might result in samples that do not follow the data distribution nor span the manifold structures. Increasing the number of labeled samples might improve spanning the manifold structure but can potentially worsen the ineffective out-of-distribution phenomenon. 

In summary, linear combination of samples is a \emph{parameter-free} and \emph{domain-independent} augmentation strategy, but it is typically not an effective augmentation strategy for data with nonlinear structures; see Table~\ref{tab:coil:as-sc} for a numerical comparison.

\begin{remark}
	In neural networks literature, prominent data augmentation strategies can be divided into two categories: 
	(i)~instance-based methods, which apply transformations on samples individually~\cite{shorten2019survey}, and 
	(ii)~mixed-example based methods, which combine several samples using various techniques for generating new samples~\cite{zhang2017mixup,inoue2018data}. 
The proposed linear interpolation can be considered as special case of mixed-example augmentations. Based on our numerical results, carefully selected and tuned instance-based strategies often lead to higher performance boost compared to mixed-example ones. 
This is due to: (i) the limited diversity in the samples generated using linear interpolations, and (ii) the aforementioned manifold intrusion phenomenon for data on nonlinear manifolds. 
Moreover, instance-based strategies do not need any label information for the data generation, and hence can be used for unsupervised tasks as well. 
However, designing and selecting appropriate label preserving strategies with no prior knowledge is still an ongoing research direction. On the other hand, mixed-example based strategies have less parameters to tune and can be easily used for data with latent linear structures, independent from the data domain. 
\end{remark}

\section{Numerical results} \label{sec:numerical}

In this section, we evaluate the proposed approaches on both synthetic and real-world data sets. All experiments are implemented in Matlab (R2019b), and run on a laptop with Intel Core i7-9750H, @2.60 GHz CPU and 16 GB RAM. The code is available from \url{https://sites.google.com/site/nicolasgillis/code}. 

\paragraph{Evaluation criteria}
We use two metrics to evaluate the clustering algorithms:
\begin{itemize}
	\item Error rate (err) which belongs to the interval $[0-100]$ and the lower values correspond to better clustering performance:
	\begin{align*}
		\text{err} = \min_{\pi, \text{ a permutation}}\frac{\sum_{i=1}^{n} \mathbf{1}\big(\ell(i) \neq 
			\hat{\ell}_\pi(i) \big) }{n} \times 100,
	\end{align*}
	where $\ell \in \{1,\dots,p\}^n$ and $\hat{\ell} \in \{1,\dots,p\}^n$ are the ground-truth and estimated labels by a clustering algorithm, respectively. In addition, $\mathbf{1}(.)$ is the indicator function which returns one if the input condition is satisfied.
	
	\item Normalized Mutual Information (NMI) which we scale between 0 and 100. A larger NMI value indicates a more accurate clustering. 
		\begin{align*}
			\text{NMI}  = \frac{I(\ell;\hat{\ell})}{\sqrt{H(\ell)H(\hat{\ell})}} \times 100,
		\end{align*}
	where $I( \cdot ; \cdot )$ and $H(\cdot)$ are the mutual information metric 
	and the entropy function, respectively.		
\end{itemize}

\paragraph{Tested algorithms} 

We compare the proposed approaches with state-of-the-art unsupervised and semi-supervised SC approaches. 
For unsupervised SC, we selected popular SC algorithms including the classic algorithms of SSC~\cite{elhamifar2013sparse}, LRR~\cite{liu2012robust}, and LSR~\cite{lu2012robust}; as well as SC with elastic net regularization in EnSC~\cite{you2016oracle}, greedy SC using OMP~\cite{you2016scalable}, nonlinear kernel based SC approach in KSSC~\cite{patel2014kernel}, and scalable kNN based SC algorithms, namely kNN-SSC~\cite{zhuang2016locality}, kNN-LRR and kNN-LSR. 
For semi-supervised SC, in addition to the state-of-the-art algorithms of NNLRR~\cite{fang2015robust}, NNLRS~\cite{zhuang2012non}, $S^3R$ and $S^2LRR$ in~\cite{li2015learning}, 
we compare the proposed AkS-SC with SSC-LP, LRR-LP, LSR-LP, which are based on the proposed AkS-SC without augmentation but using the iterative label propagation scheme. 
The parameters of these approaches are selected and tuned according to the recommendations in the corresponding papers.

\paragraph{Data sets}

We use two real-world data sets: %, namely COIL-20  and MNIST, for evaluating and comparing the performance of proposed algorithms with the current state-of-the-art SC approaches.
\begin{itemize}
	\item The COIL-20 data set (Columbia Object Image Library)~\cite{nene1996columbia} contains 1440 images of 20 objects. Each category has 72 images of an object which were captured from different views/poses. All the images are cropped and processed to $128 \times 128$ pixels. Following the common approach in SC literature, we downsampled the images to to $32 \times 32$ pixels~\cite{ji2017deep}. 
	
	\item The MNIST data set contains 70000 images of 10 handwritten digits with the size 28$\times$28 pixels. This data set is divided into 60000 training images and 10000 testing images. We selected the first 50 images per digit from the MNIST test data set. Based on the common practice introduced in~\cite{you2016scalable}, we use a scattering convolution network~\cite{bruna2013invariant} to extract feature vectors of dimension 3472, and then project them to dimension 100 using PCA.
\end{itemize}
In addition to this, we will use a carefully designed synthetic data generation model to analyze in detail the effect of augmentation in the semi-supervised setting; see Section~\ref{sec:syntexp}. 

%to evaluate the proposed semi-supervised SC algorithm and investigate the effect of data augmentation in details.

\subsection{Analyzing the unsupervised A-SC}

In this Section, we evaluate the proposed scalable unsupervised Ak-SC algorithms: Ak-SSC, Ak-LSR and Ak-LRR, on two real-world data sets of COIL-20 and MNIST.

\subsubsection{Applying Unsupervised SC on COIL-20}

\paragraph{Comparing performances}

Table~\ref{tab:coil20} provides the error rates and NMI on the tested  unsupervised SC algorithms on the COIL-20 data set. 
For our proposed augmented based SC algorithms, we used the standard predefined augmentation strategies of \emph{flipping left to right}, 5 \emph{random rotations} within the range $[-10^\circ,10^\circ]$, and 5 \emph{random scalings} within the range $[0.9,1.1]$. Furthermore, we set $k=20$ and $\lambda=\frac{\mu}{\max_{i\neq j}|X(:,i)^\top X(:,j)|}$, with $\mu=30$. Due to the randomness of generated augmented samples, the average of error rate and NMI over 10 trials are reported for Ak-SC algorithms. 
\begin{center}
	\begin{table}[!ht]
		\begin{center} \small
			\caption{Comparison of unsupervised SC algorithms on the COIL-20 data set.}
			\label{tab:coil20} 
			\resizebox{\textwidth}{!}{ 
				\begin{tabular}{c||ccc|ccc|ccc|ccc}
					\hline
					Evaluation & SSC & LRR & LSR & OMP & EnSC & KSSC & kNN-SSC & kNN-LRR & kNN-LSR  & Ak-SSC & Ak-LRR & Ak-LSR \\ \hline
					err & 25.90 & 35.63 & 35.76 & 33.89 & 22.08 & 14.93 & 23.33 & 24.37 & 22.29 & 0.31$\pm$0.23 & 0.48$\pm$0.20 & 0.20$\pm$0.27\\
					NMI & 88.67 & 75.98 & 71.86 & 78.86 & 88.30 & 95.71 & 93.03 & 86.34 & 86.41 & 99.64$\pm$0.26 & 99.48$\pm$0.20 & 99.78$\pm$0.29\\ \hline
			\end{tabular}}
		\end{center}
	\end{table}
\end{center}

We observe that:
\begin{itemize}
	\item All augmented SC algorithms significantly outperform all other unsupervised SC approaches. 	There is a significant gap between the performance of classic algorithms and the corresponding augmented ones: from 15\% error rate for the best classical algorithm (KSSC) to less than 0.5\% for all augmented algorithms.
	
	\item Augmented SC algorithms outperform the kNN-based algorithms. 
	This shows that the considerable improvement in Ak-SC algorithms is due to the addition of augmented samples, not to the use of kNN.  
	
	\item The improvement in the performance of augmented SC algorithms is not limited to the sparsity-promoting SSC. 
	In fact, Ak-LRR and Ak-LSR perform similarly. 
	Hence, although we have only provided the geometrical motivations for using augmented samples for SSC (Section~\ref{sec:geometry}),  augmentation improves the performance of other SC models as well. 
\end{itemize}
In fact, the additional implicit information in the augmented samples helps to improve the ``quality" of the coefficient matrix by strengthening the correct connections between the samples from same clusters and simultaneously weakening/eliminating the wrong connections between the samples from different clusters. This is illustrated for a specific sample from the COIL-20 dataset in the supplementary material~\ref{suppl:addnumexpqualcoef}.

\begin{remark}[Influence of $k$ and of augmentation strategies] 
The numerical experiments above do not provide information on the sensitivity with respect to the value of $k$ in kNN, and to the choice of the augmentation strategies. In the supplementary material~\ref{suppl:addnumexpkaug}, we provide extensive numerical experiments about this question. We observe that generating augmented samples using any augmentation strategy within the set \{ \{flipping\}, \{rotation, scaling\}, \{flipping, rotation, scaling\}\} results in a notable improvement for all the  unsupervised Ak-SC algorithms, independently of the value of $k$. Moreover, as the augmentation \{flipping, rotation, scaling\} provides the most implicit information in the self-expressive dictionary, it leads to the highest performance improvement among the other strategies. 
\end{remark}

\subsubsection{Applying Unsupervised SC on MNIST}

\paragraph{Comparing performances}

We compared the performance of Ak-SC algorithms with other state-of-the-art SC algorithms on the MNIST-test data set as well. For our proposed Ak-SC algorithms, we have set $k=30$ and $\lambda=\frac{\mu}{\max_{i\neq j}|X(:,i)^\top X(:,j)|}$, with $\mu=100$. We use 5 random scaling within the range $[0.8,1.2]$ and 5 random rotations with the rotation angle chosen randomly from the range $[-30^\circ,30^\circ]$. We report the average results of 10 trials for Ak-SC algorithms. The error rate and NMI corresponding to the selected SC algorithms are summarized in Table~\ref{tab:mnist:unsup}. 

For more than 6 digits, augmented based SC algorithms significantly outperform other SC algorithms. All three augmented  SC algorithms, Ak-SSC, Ak-LSR and Ak-LRR, perform similarly in most of the cases and outperform the corresponding kNN-based algorithms. This confirms that the improvement in the performance is due to the augmented samples. Moreover, the performance of Ak-SC is not sensitive to the parameter of $k$; see the supplementary material~\ref{suppl:addnumexpkMNIST} for numerical experiments.
\begin{center}
	\begin{table}[!ht]
		\begin{center} \small
			\caption{Comparison of unsupervised SC algorithms on the MNIST-test data set.}
			\label{tab:mnist:unsup} 
			\resizebox{\textwidth}{!}{ 
				\begin{tabular}{c|c||ccc|ccc|ccc|ccc}
					\hline
					Digits & Evaluation & SSC & LRR & LSR & OMP & EnSC & KSSC & kNN-SSC & kNN-LRR & kNN-LSR  & Ak-SSC & Ak-LRR & Ak-LSR \\ \hline
					\multirow{2}{*}{[0:1]}& err & 0 & 49.00 & 0 & 0 & 1.00 & 0 & 0 & 0 & 0 & 0$\pm$0 & 0$\pm$0 & 0$\pm$0 \\
					& NMI & 100 & 3.54 & 100 & 100 & 92.91 & 100 & 100 & 100 & 100 & 100$\pm$0 & 100$\pm$0 & 100$\pm$0 \\\hline
					\multirow{2}{*}{[0:2]}& err & 0 & 50.00 & 0 & 0.67 & 0 & 0 & 0.66 & 1.33 & 3.33 & 2.80$\pm$0.61 & 1.46$\pm$1.32 & 3.06$\pm$0.46 \\
					& NMI & 100 & 13.02 & 100 & 97.02 & 100 & 100 & 97.01 & 94.87 & 89.96 & 90.61$\pm$1.98 & 94.75$\pm$4.41 & 89.55$\pm$1.48 \\\hline
					\multirow{2}{*}{[0:3]}& err & 32.50 & 38.00 & 1.50 & 5.00 & 34.00 & 1.50 & 3.5 & 5.00 & 6.50 & 2.00$\pm$0.40 & 1.50$\pm$0.47 & 2.45$\pm$0.15 \\
					& NMI & 80.87 & 41.94 & 95.87 & 85.92 & 80.51 & 95.20 & 88.64 & 86.76 & 82.00 & 93.8$\pm$1.15 & 95.22$\pm$1.41 & 92.59$\pm$0.59 \\\hline
					\multirow{2}{*}{[0:4]}& err & 27.60 & 4.00 & 1.20 & 4.00 & 27.20 & 1.60 & 2.00 & 4.40 & 4.80 & 2.20$\pm$0.33 & 1.92$\pm$0.59 & 2.88$\pm$0.36 \\
					& NMI & 85.44 & 90.41 & 96.69 & 89.68 & 86.82 & 95.47 & 94.26 & 89.68 & 89.18 & 93.99$\pm$0.82 & 94.75$\pm$1.39 &92.23$\pm$1.12 \\\hline
					\multirow{2}{*}{[0:5]}& err & 22.33 & 12.67 & 9.33 & 23.33 & 23.67 & 7.67 & 9.66 & 10.66 & 11.66 & 3.33$\pm$0.27 & 3. 03$\pm$0.29 & 4.40$\pm$1.23 \\
					& NMI & 83.87 & 80.61 & 85.76 & 80.66 & 83.61 & 88.80 & 83.12 & 81.53 & 78.92 & 92.06$\pm$0.73 & 92.55$\pm$0.72 & 90.33$\pm$1.33 \\\hline
					\multirow{2}{*}{[0:6]}& err & 20.29 & 10.00 & 8.00 & 23.14 & 20.86 & 7.71 & 9.42 & 9.71 & 10.54 & 3.68$\pm$0.92 & 3.80$\pm$0.78 & 4.14$\pm$0.20 \\
					& NMI & 85.18 & 86.13 & 87.67 & 79.28 & 85.62 & 88.24 & 84.28 & 83.90 & 82.42 & 91.95$\pm$1.17 & 91.55$\pm$1.18 & 90.83$\pm$0.50 \\\hline
					\multirow{2}{*}{[0:7]}& err & 19.50 & 8.00 & 6.75 & 24.75 & 20.25 & 8.00 & 9.50 & 9.75 & 9.50 & 3.80$\pm$0.28 & 3.95$\pm$0.34 & 4.70$\pm$1.12 \\
					& NMI & 84.56 & 87.02 & 88.43 & 77.08 & 83.88 & 88.24 & 84.80 & 84.55 & 84.49 & 92.08$\pm$0.58 & 91.79$\pm$0.78 & 90.92$\pm$1.10  \\\hline
					\multirow{2}{*}{[0:8]}& err & 18.44 & 21.33 & 22.22 & 22.22 & 18.67 & 14.22 & 20.80 & 18.86 & 16.24 & 6.57$\pm$0.83 & 6.44$\pm$0.77 & 7.22$\pm$0.77 \\
					& NMI & 84.47 & 78.53 & 79.43 & 75.77 & 87.59 & 79.97 & 71.87 & 74.89 & 75.89 & 87.58$\pm$0.94 & 87.87$\pm$0.67 & 86.68$\pm$0.75 \\\hline
					\multirow{2}{*}{[0:9]}& err & 18.80 & 21.40 & 21.80 & 27.80 & 20.00 & 15.80 & 20.36 & 19.32 & 19.00 &8.40$\pm$0.76 & 8.32$\pm$1.05 & 8.40$\pm$0.44 \\
					& NMI & 82.66 & 77.82 & 79.51 & 70.72 & 81.63 & 77.89 & 71.61 &72.81 & 72.38 &85.32$\pm$0.75 &  85.41$\pm$1.03 & 85.09$\pm$0.71  \\\hline
			\end{tabular}}
		\end{center}
	\end{table}
\end{center}

\subsection{Analyzing the semi-supervised AS-SC}

In this section, we evaluate the effect of augmentation in our proposed semi-supervised algorithms. We first use a synthetic data generation model which is inspired by~\cite{soltanolkotabi2012geometric} to illustrate the geometric  insights provided in Section~\ref{sec:geometry} for augmented semi-supervised SSC (AS-SSC). We then compare the performance of the proposed approaches with the state-of-the-art semi-supervised SC algorithms on the two real-world data sets.

\subsubsection{Synthetic data sets} \label{sec:syntexp}

%In this section, we illustrate the impact of data augmentation using a set of experiments on a carefully designed synthetic data set. 
As we discussed in Section~\ref{sec:geometry}, the success of SSC depends on two factors: inradius and subspace incoherence. In order to control the subspace incoherence between synthetic subspaces, we follow the common semi-random model in \cite{soltanolkotabi2012geometric}, and consider three disjoint subspaces with intrinsic dimension $d_1 = d_2 = d_3 = 3$ in the ambient space of six dimensions, $d=6$. The bases of the three subspaces, $\{U_j\}_{j=1}^3 \in \mathbb{R}^{6 \times 3}$, are generated according to the following deterministic model:
\[
U_1 
= \binom{\cos(\theta)  I_{3}}{\sin(\theta)  I_{3}}, 
U_2 
= \binom{\cos(\theta)  I_{3}}{-\sin(\theta)  I_{3}}, 
U_3= \binom{I_{3}}{0_{3}},
\] 
where $\theta \in [0, \frac{\pi}{2}]$ controls the affinity between the three subspaces. The lower value of $\theta$ corresponds to higher subspace incoherence and closer subspaces which leads to a more challenging SC problem.
We set $n=60$ and for each subspace, we generate $n/3$ random samples by linear combinations of $U_j$ ($j=1,2,3$). The weights in the linear
combinations are randomly chosen using the Gaussian distribution with mean 0 and variance 1. The samples are normalized to have unit $\ell_2$ norm. The parameters of the optimization problem for AS-SSC are set to $\lambda=\frac{\mu}{\max_{i\neq j}|X(:,i)^\top X(:,j)|}$, with $\mu=50$ and $\lambda_2=1$. In fact, AS-SSC is not sensitive to the value of $\lambda_2$ and the results are almost the same for a large range of $\lambda_2 \in [1,10]$. 
The supplementary material~\ref{suppl:addnumexpsens} provides a detailed parameter sensitivity analysis.

\paragraph{Illustration of the label propagation}

%in matlab test_sample.m
Let us illustrate the effect of label propagation on the coefficient matrix and the corresponding implicit graph structure. 
%We generate samples from the three disjoint subspaces using the aforementioned synthetic data model. 
The parameter $\theta$ is set to $10^\circ$, hence, the subspaces are close to each other. We randomly select four samples from each subspace as the given labeled samples and produce 50 augmented samples for each subspace using linear combination of the labeled samples, with the linear weights set by random Gaussian distribution. 

\begin{figure*}[!ht]
	\begin{minipage}[b]{0.5\linewidth}
		\centering
		\centerline{\includegraphics[width=8cm]{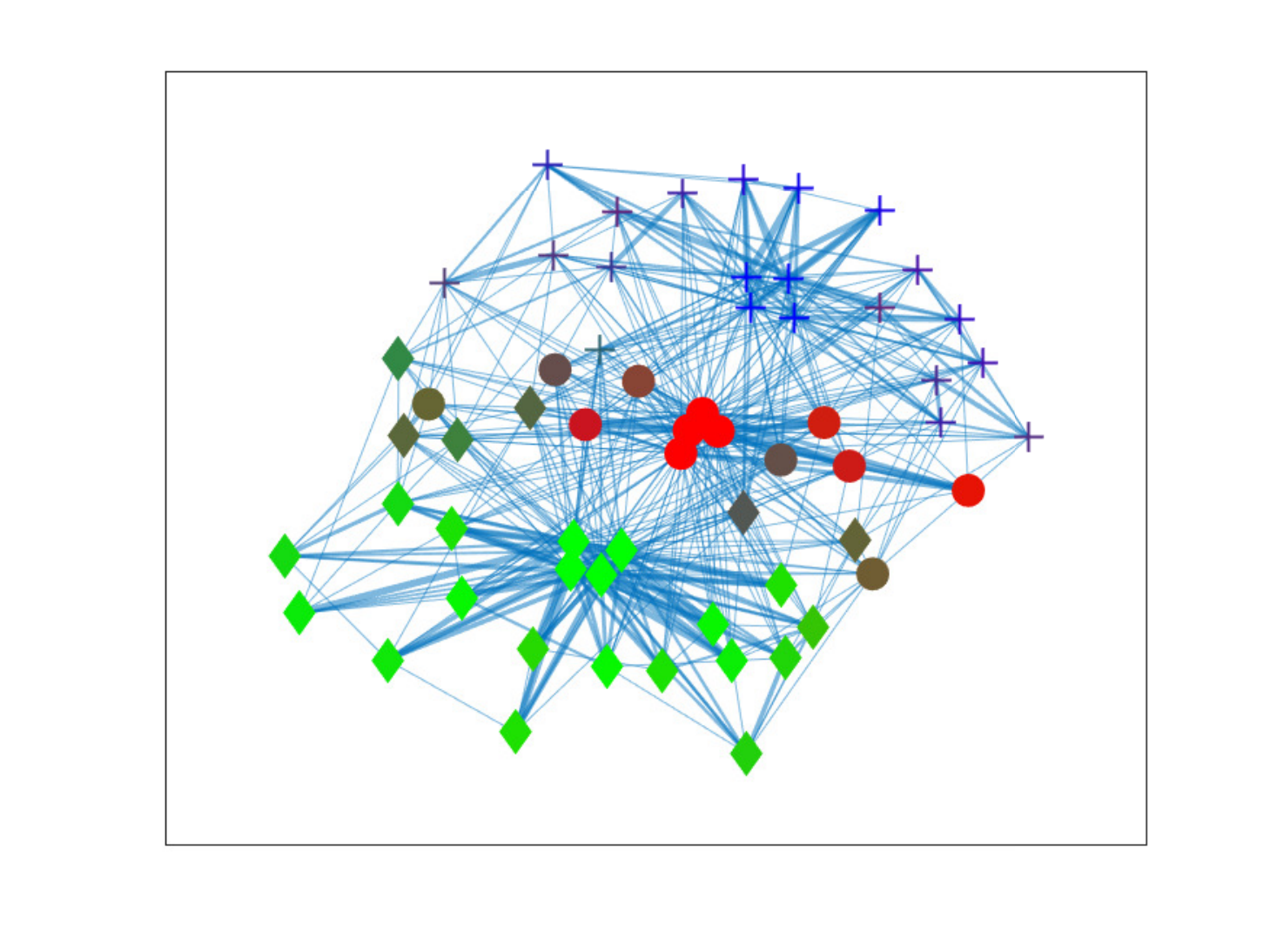}}
		%  \vspace{1.5cm}
		\centerline{(a) iteration 1, err = 11.67\%}\medskip
	\end{minipage}
	\hfill
	\begin{minipage}[b]{0.5\linewidth}
		\centering
		\centerline{\includegraphics[width=8cm]{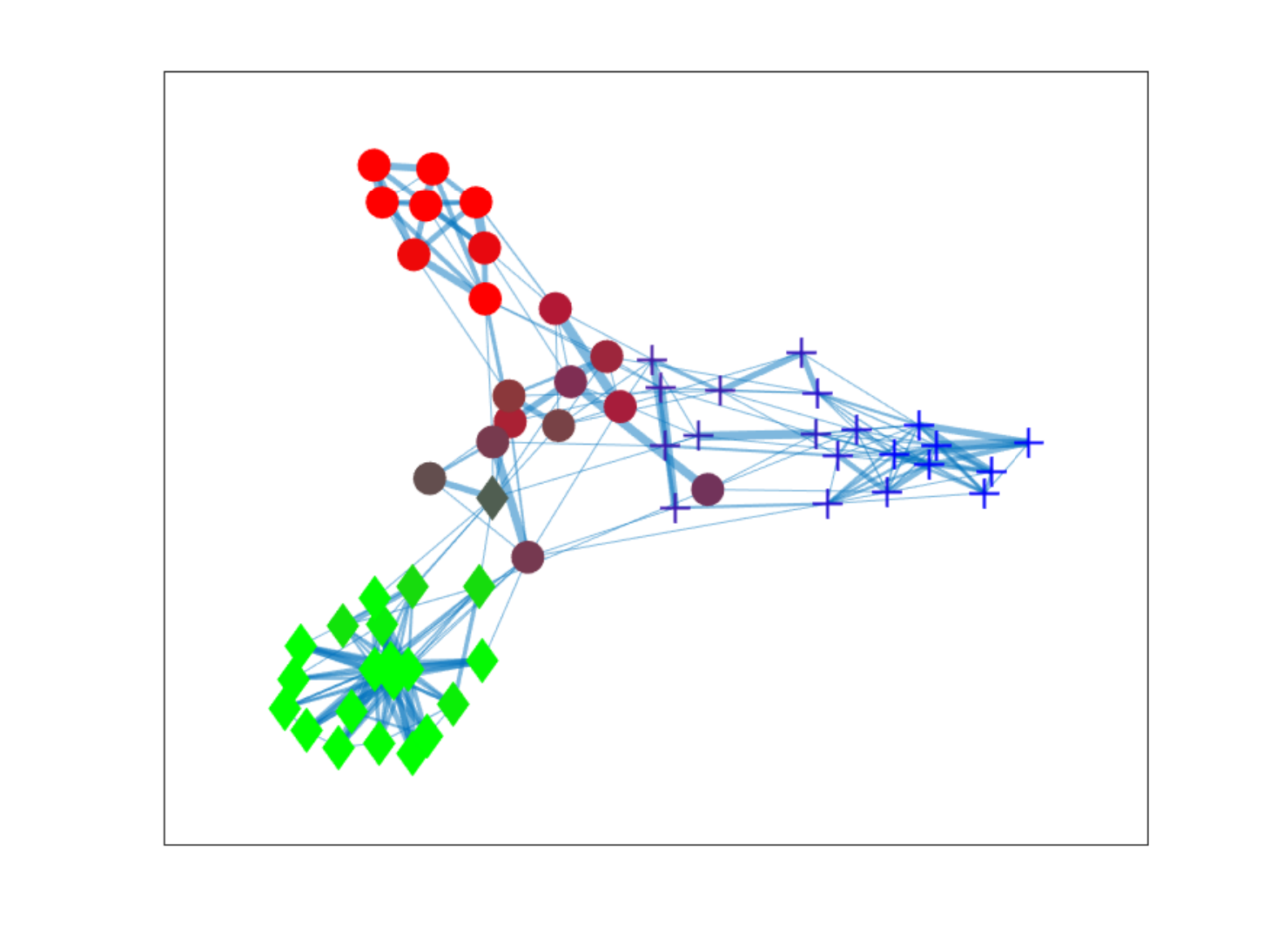}}
		%  \vspace{1.5cm}
		\centerline{(b) iteration 2, err = 1.67\%}\medskip
	\end{minipage}
	\hfill
	\begin{minipage}[b]{0.5\linewidth}
		\centering
		\centerline{\includegraphics[width=8cm]{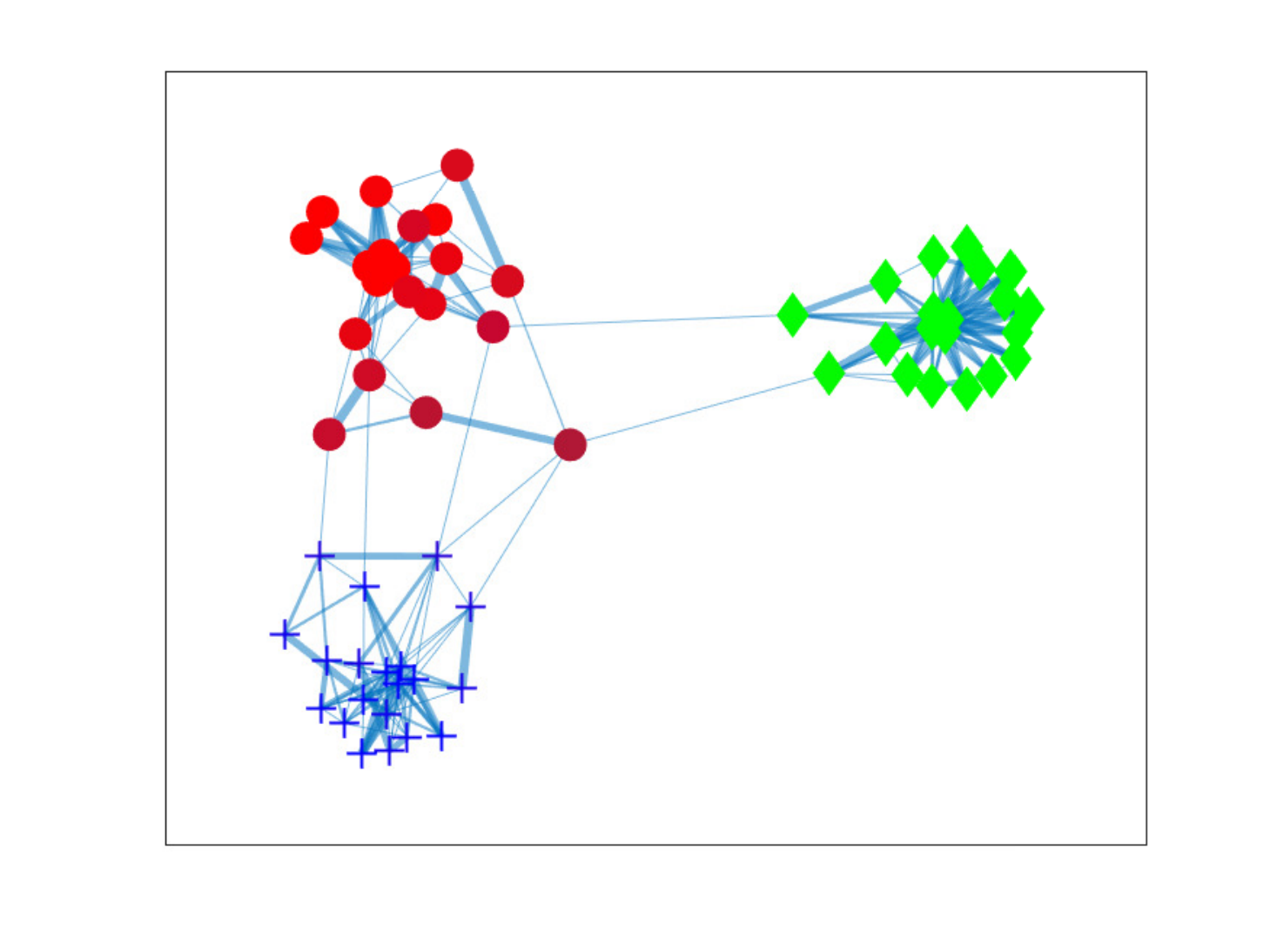}}
		%  \vspace{1.5cm}
		\centerline{(c) iteration 3, err =0\%}\medskip
	\end{minipage}
	\hfill
	\begin{minipage}[b]{0.5\linewidth}
		\centering
		\centerline{\includegraphics[width=8cm]{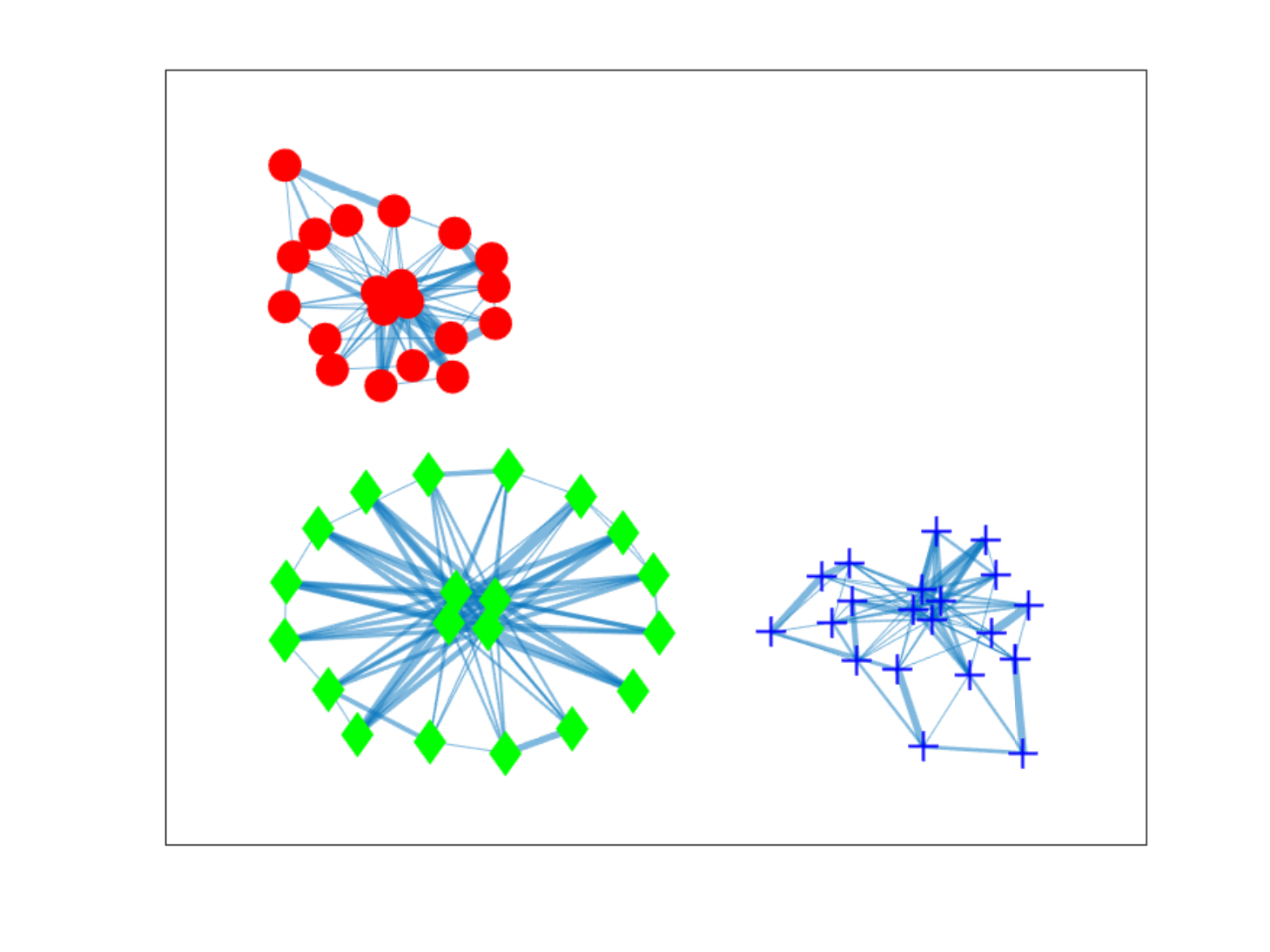}}
		%  \vspace{1.5cm}
		\centerline{(d) iteration 4, err = 0\%}\medskip
	\end{minipage}
	\caption{Illustration of the impact of label propagation on the graph connectivity through different iterations in AS-SSC.}
	\label{LP_exp}
\end{figure*}

The connectivity graph corresponding to the coefficient matrix $C_f$ for each iteration of the AS-SSC is shown in Figure~\ref{LP_exp}. The estimated labels in the matrix $F \in \mathbb{R}^{210 \times 3}$ at  each iteration are used as the colors of the nodes in the graph. Additionally, three symbols of \{cross, circle, diamond\} are used to indicate the assigned labels to the samples of each cluster, that is, $\argmax F(j,:)$ for $j=1,\dots,n$. In particular, each row of the matrix $F$ is a vector with three entries which can be used as a RGB triplet to color the corresponding node. The graph obtained after the first iteration is shown in (a). We observe many wrong connections between the samples of different clusters, and, consequently, the error rate corresponding to this graph is 11.67\%. 
Regardless, the labels of several samples are already estimated with high confidence which is reflected in their corresponding node color: there are several nodes with absolute green, red and blue colors which are equivalent to (1,0,0), (0,1,0) and (0,0,1) in the label matrix $F$. The obtained label information is used in the next iteration to reduce the connections between the samples with  different estimated labels which reduces the error rate to 1.67\%. Eventually, the error rate reduces to 0\% after the third iteration and the iterative algorithm converges after five iterations. 

\paragraph{Number of labeled samples vs the number of augmented samples} %in matlab: num_lab_vs_num_aug.m

Recall that according to~\eqref{ssc_condition} the value of inradius has a key role for successful clustering of nearby subspaces, that is, for high values of subspace incoherence. 
Here, we use a numerical example to analyze the impact of subspace incoherence on the performance of AS-SSC and investigate the effect of the number of labeled samples and the number of the augmented samples on the inradius which is indirectly measured by the clustering error rate. 

\begin{figure*}[!ht]
	\begin{minipage}[b]{0.3\linewidth}
		\centering
		\centerline{\includegraphics[width=5.8cm]{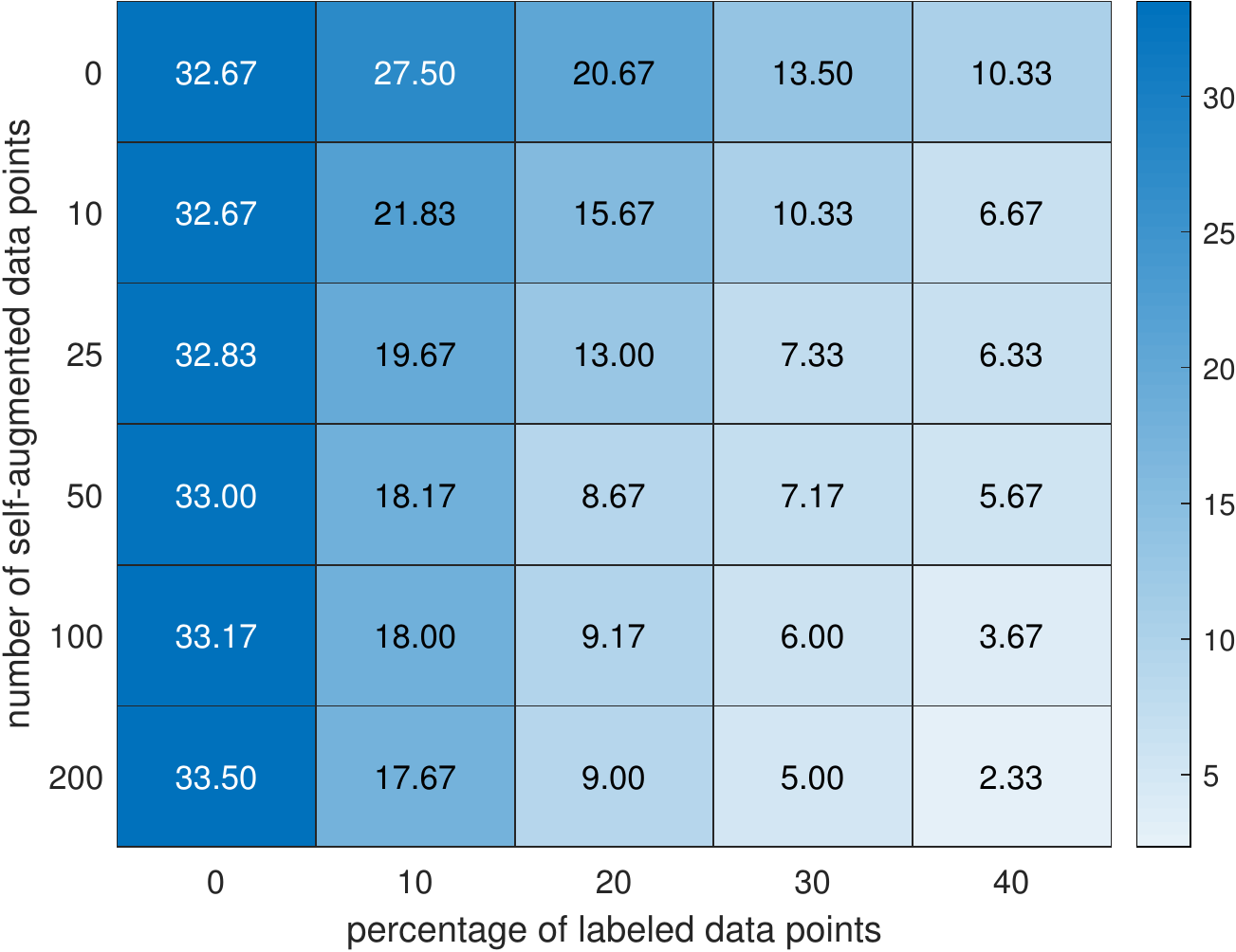}}
		%  \vspace{1.5cm}
		\centerline{(a) $\theta = 10^\circ$, one iteration}\medskip
	\end{minipage}
	\hfill
	\begin{minipage}[b]{0.3\linewidth}
		\centering
		\centerline{\includegraphics[width=5.8cm]{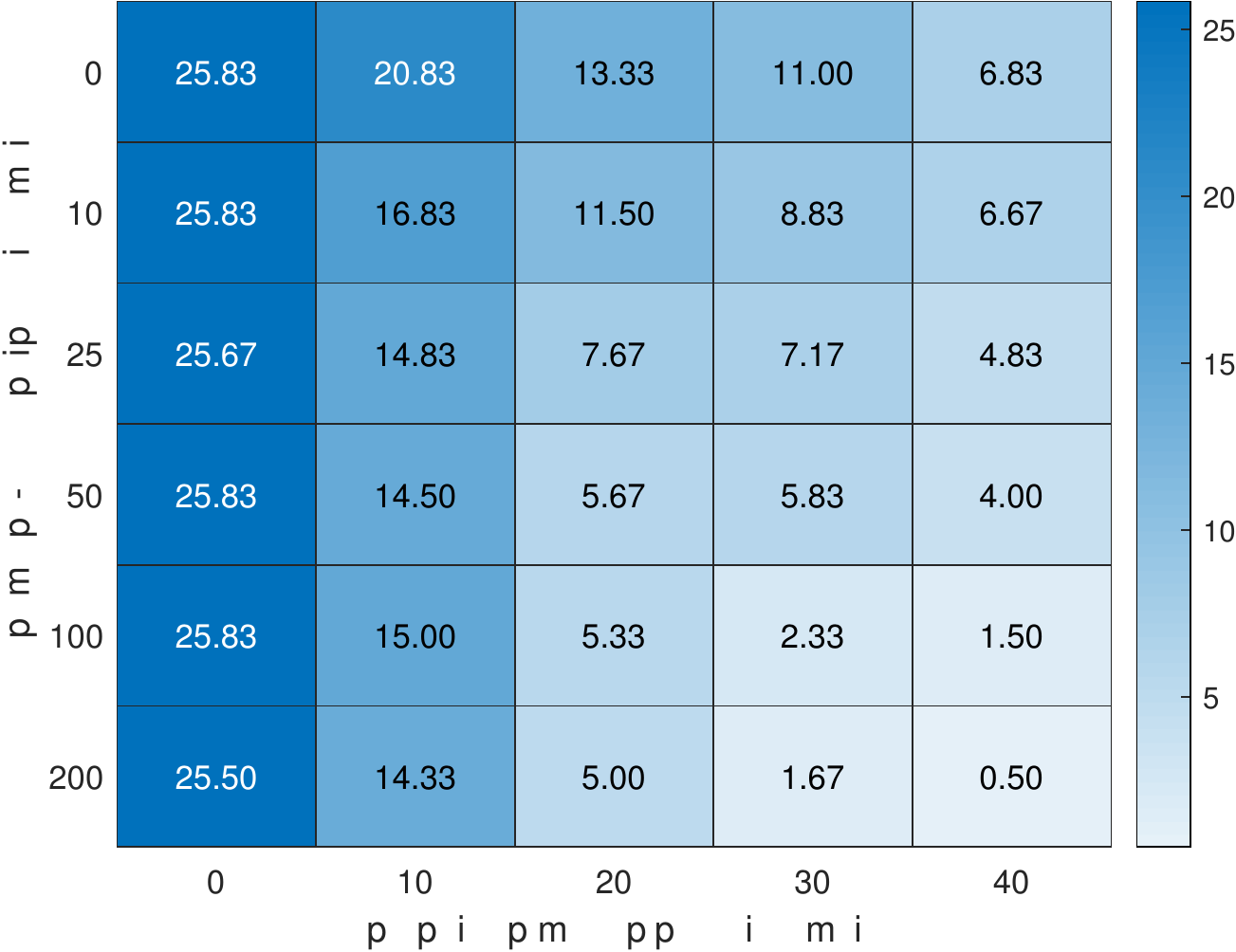}}
		%  \vspace{1.5cm}
		\centerline{(b) $\theta = 15^\circ$, one iteration}\medskip
	\end{minipage}
	\hfill
	\begin{minipage}[b]{0.3\linewidth}
		\centering
		\centerline{\includegraphics[width=5.8cm]{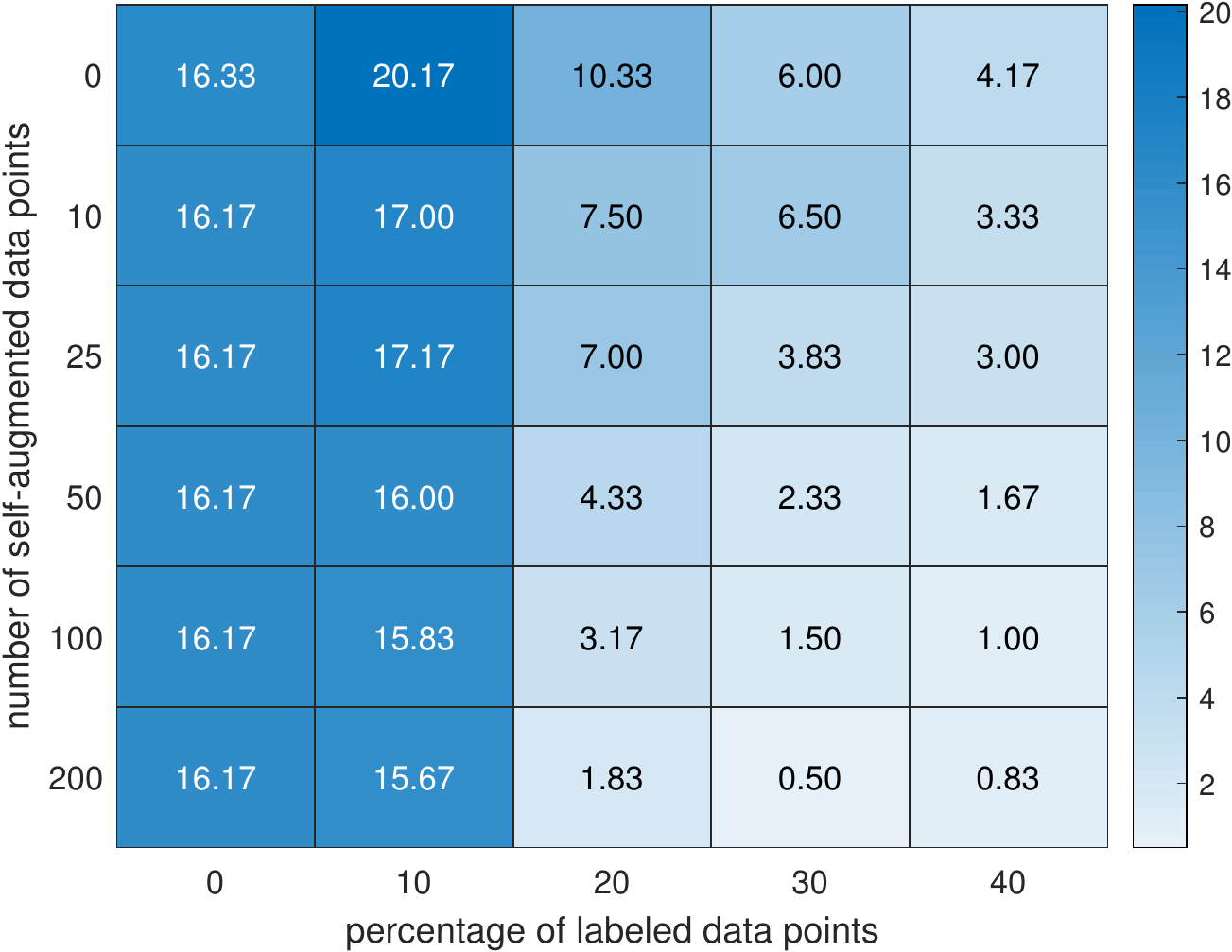}}
		%  \vspace{1.5cm}
		\centerline{(c) $\theta = 20^\circ$, one iteration}\medskip
	\end{minipage}
	\hfill
	\begin{minipage}[b]{0.3\linewidth}
		\centering
		\centerline{\includegraphics[width=5.8cm]{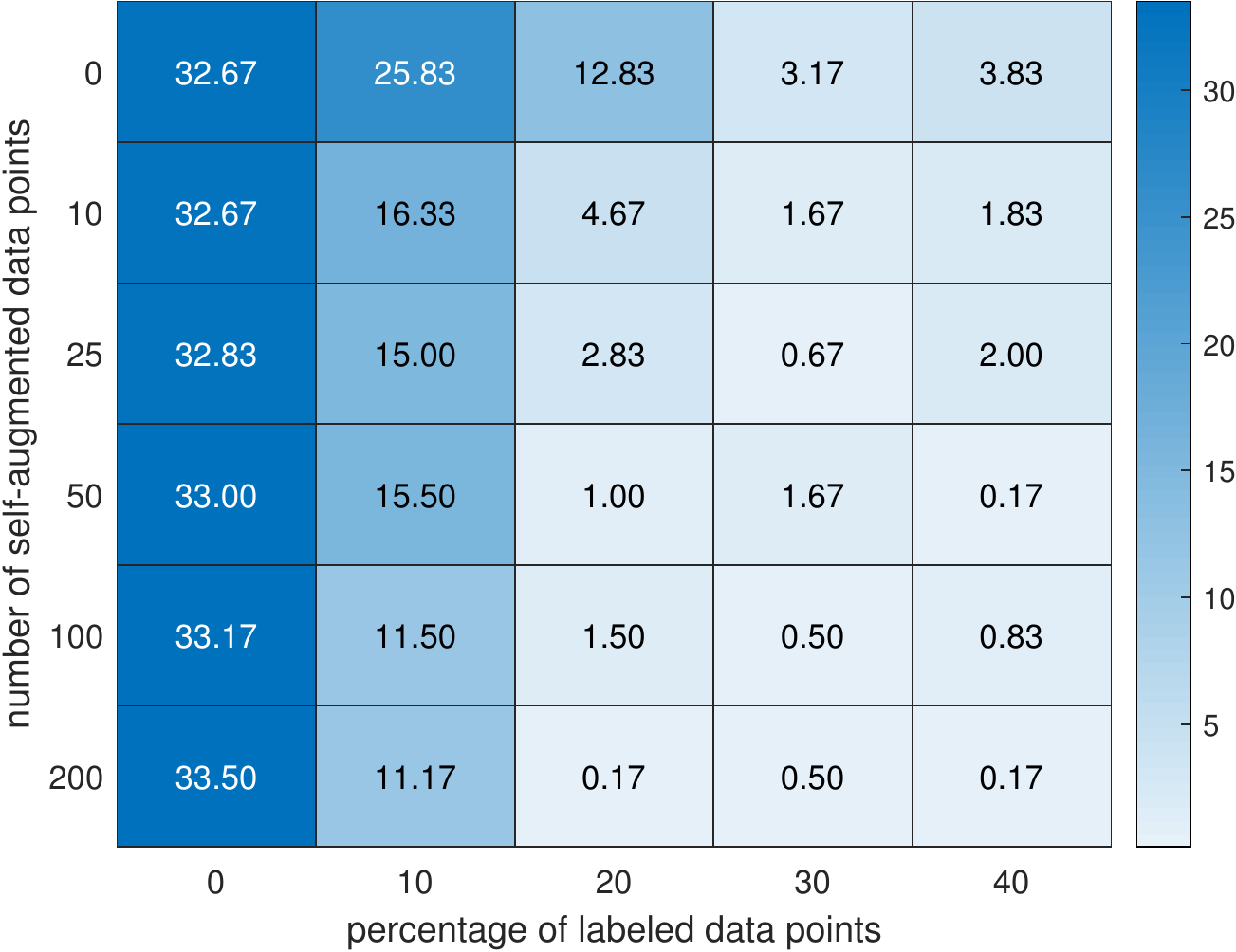}}
		%  \vspace{1.5cm}
		\centerline{(d) $\theta = 10^\circ$, iterative LP}\medskip
	\end{minipage}
	\hfill
	\begin{minipage}[b]{0.3\linewidth}
		\centering
		\centerline{\includegraphics[width=5.8cm]{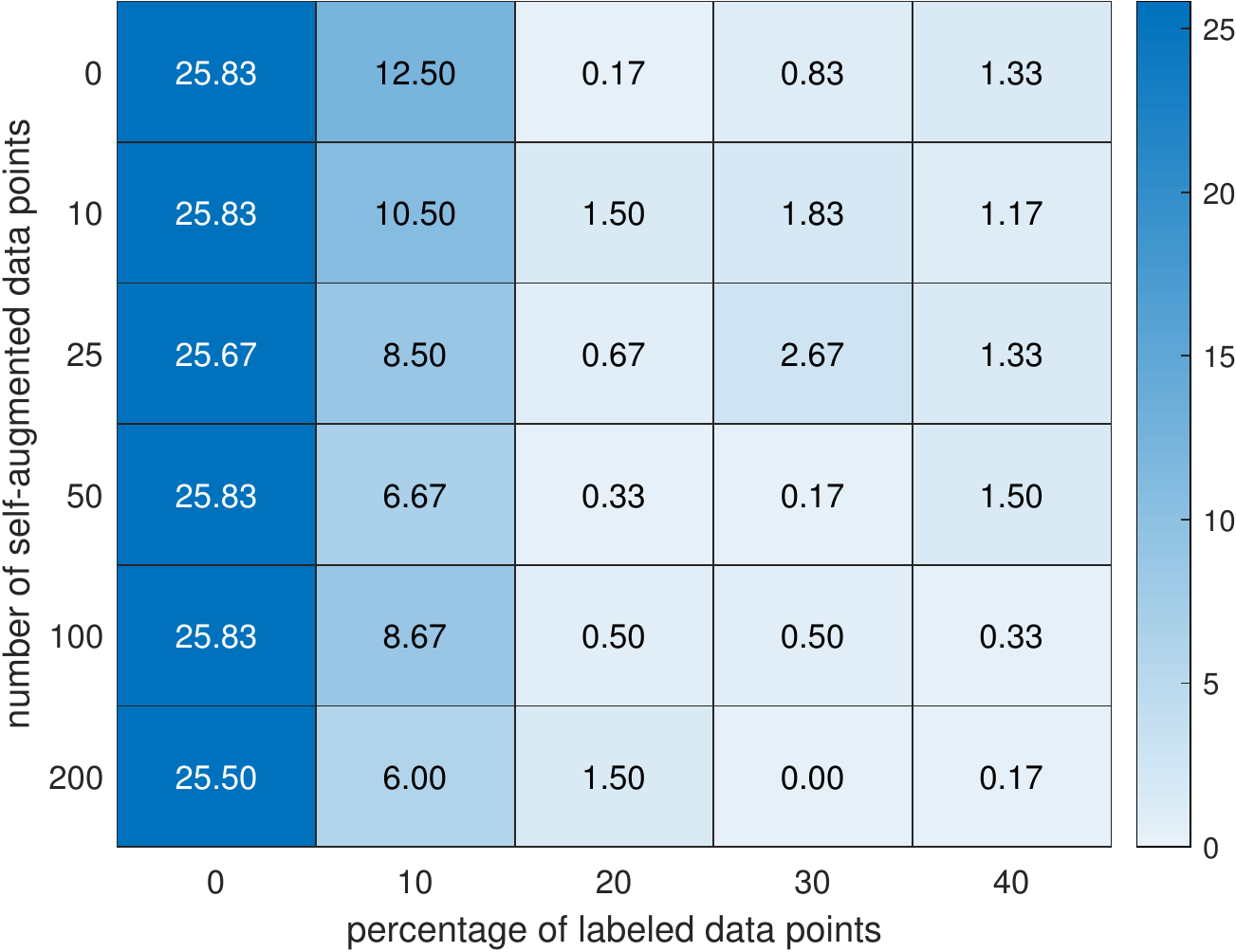}}
		%  \vspace{1.5cm}
		\centerline{(e)  $\theta = 15^\circ$, iterative LP}\medskip
	\end{minipage}
	\hfill
	\begin{minipage}[b]{0.3\linewidth}
		\centering
		\centerline{\includegraphics[width=5.8cm]{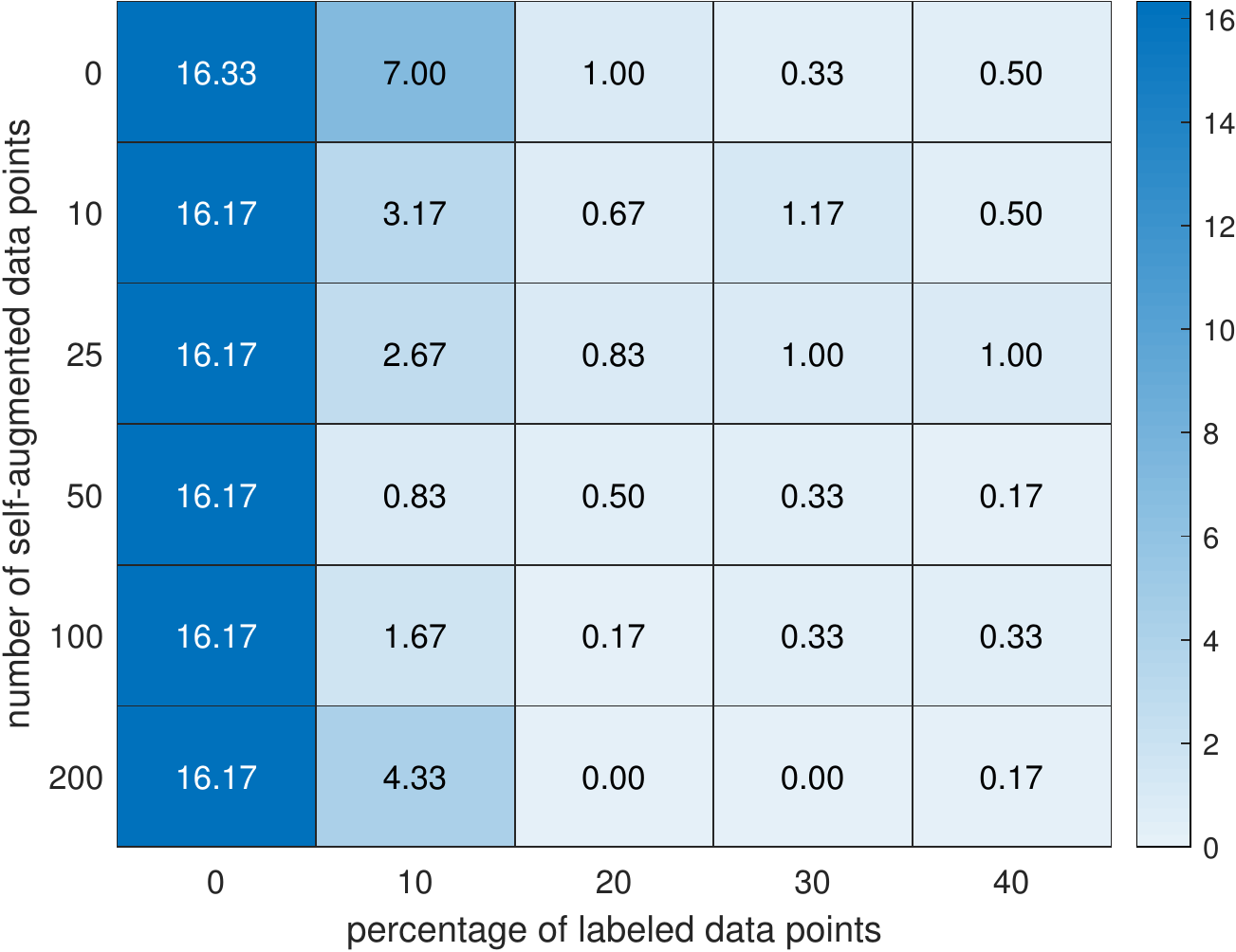}}
		%  \vspace{1.5cm}
		\centerline{(f)  $\theta = 20^\circ$, iterative LP}\medskip
	\end{minipage}
	\caption{The effect of number of augmented samples vs percentage of labeled samples on the average of error rate of AS-SSC over 10 trials for the synthetic data set. The experiment is implemented for various subspace incoherences which is controlled by the subspace angle parameter, $\theta$. %\ngc{The graphs on the top are different than the ones at the bottom (which look better), 	it would be nice to homogenize.} \textcolor{blue}{--done--} 
	} 
	\label{num_lab_vs_num_aug}
\end{figure*}

We consider three different angles between subspaces in the synthetic data generation model, namely $\theta \in \{10,15,20\}^\circ$. 
A smaller $\theta$ corresponds to a higher value for subspace incoherence. The average error rate over 10 trials for both the first iteration of the AS-SSC (merely using the augmented data) and after applying iterative AS-SSC using label propagation are reported in Figure~\ref{num_lab_vs_num_aug}. Several percentages of labeled samples (\%) are considered, namely \{0, 10, 20, 30, 40\}, and the number of augmented samples are chosen in the set \{0, 10, 25, 50, 100, 200\}.

We observe that: 
\begin{itemize}
	\item AS-SSC using iterative LP always decreases the average error rate and improves the quality of clustering. 
	\item Increasing the number of labeled samples generally improves the performance as it increases the diversity of the augmented samples and introduces more restrictions on the possible connections between the labeled samples from different clusters.
	\item A significant decrease in the error rate is observed by the addition of 50 augmented samples (per cluster). Increasing the number of augmented samples to 100 and 200 leads to further slight decrease in the error rate. 
	\item The error rate corresponding to classic SSC is the entry on the top left corner of each table. AS-SSC significantly decreases this error rate using any number of augmented samples. 
	\item The intrinsic dimension of each subspace is set to three. The 20\% percentage labeled samples is equal to 4 labeled samples for this synthetic example. We observe that as soon as the number of labeled samples is higher than the intrinsic dimension of subspaces, AS-SSC results in a major improvement in the clustering performance. 
	\item As the angles between subspaces increase and,  consequently, the subspace incoherence decreases, the SC problem gets less challenging and the amount of decrease in the error rate due to AS-SSC becomes less evident. This is completely in line with the theoretical condition in~\eqref{ssc_condition}. As the incoherence decreases, the value of the inradius does not need to be as large to obtain subspace preserving coefficients, and hence the data augmentation has less impact on the corresponding SC problem.
\end{itemize}

\paragraph{Convergence analysis}

\begin{figure*}[!ht]
	\begin{minipage}[b]{0.3\linewidth}
		\centering
		\centerline{\includegraphics[width=6cm]{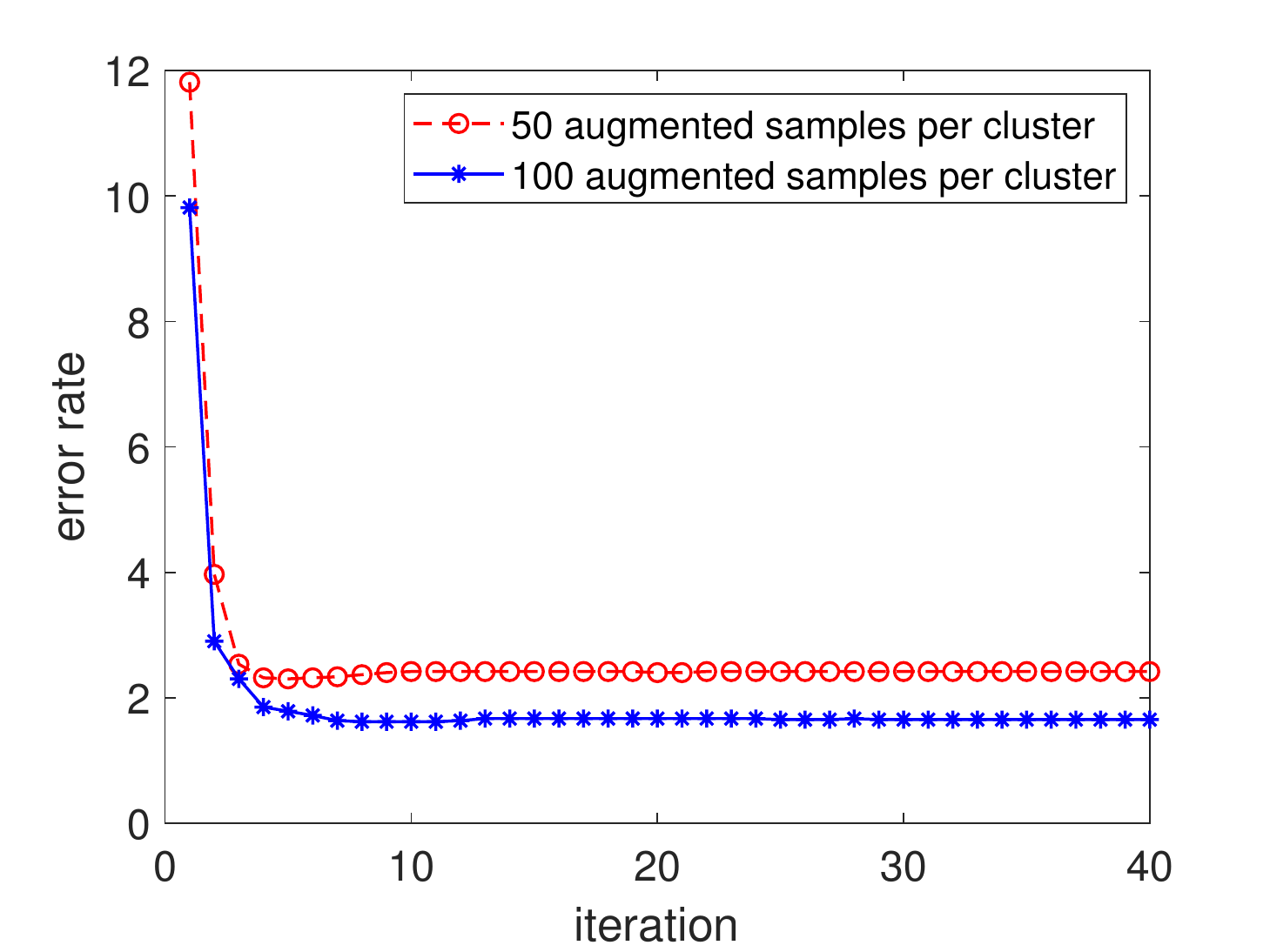}}
		%  \vspace{1.5cm}
		\centerline{(a) $\theta = 10^\circ$}\medskip
	\end{minipage}
	\hfill
	\begin{minipage}[b]{0.3\linewidth}
		\centering
		\centerline{\includegraphics[width=6cm]{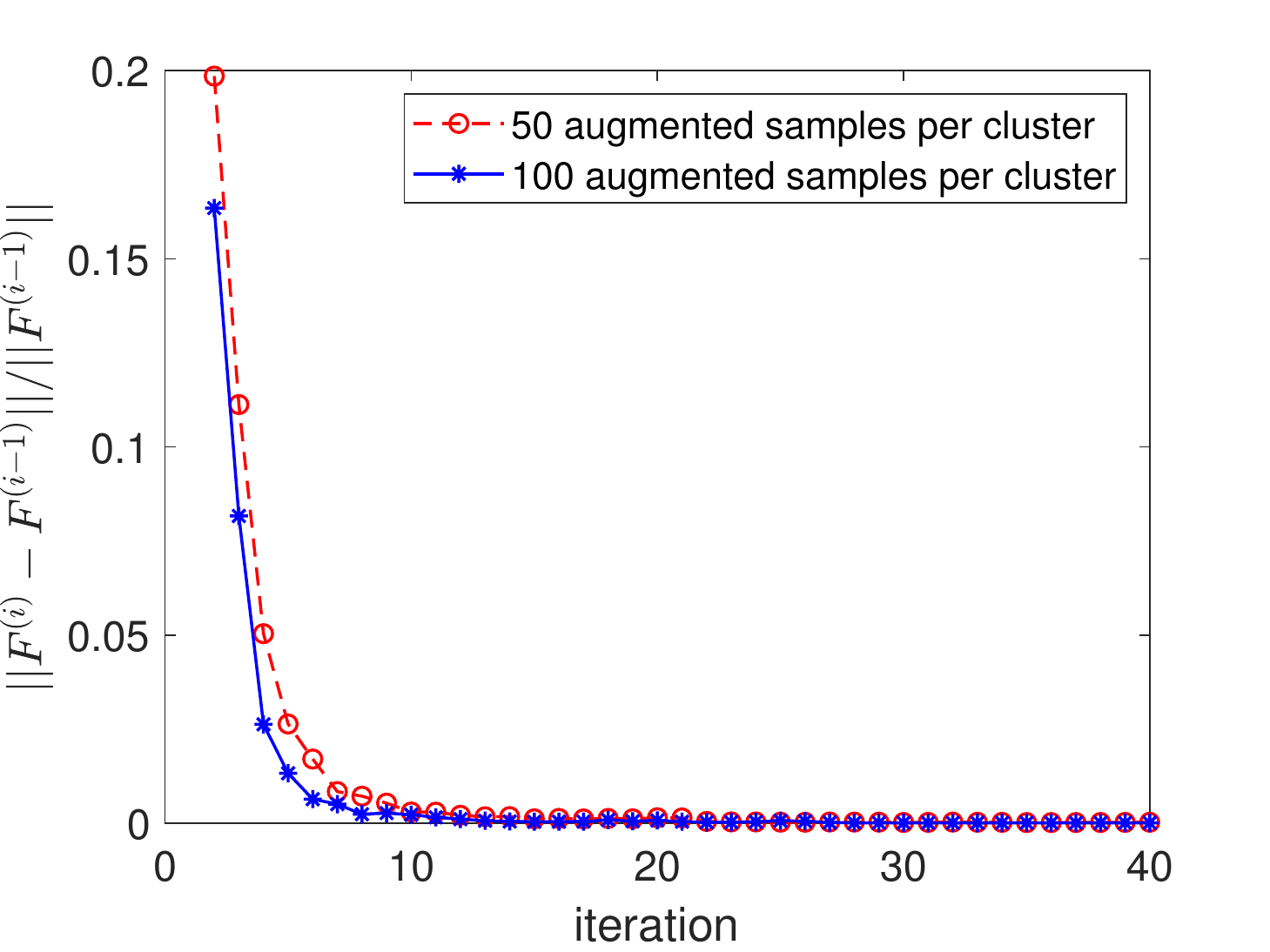}}
		%  \vspace{1.5cm}
		\centerline{(b) $\theta = 10^\circ$}\medskip
	\end{minipage}
	\hfill
		\begin{minipage}[b]{0.3\linewidth}
		\centering
		\centerline{\includegraphics[width=6cm]{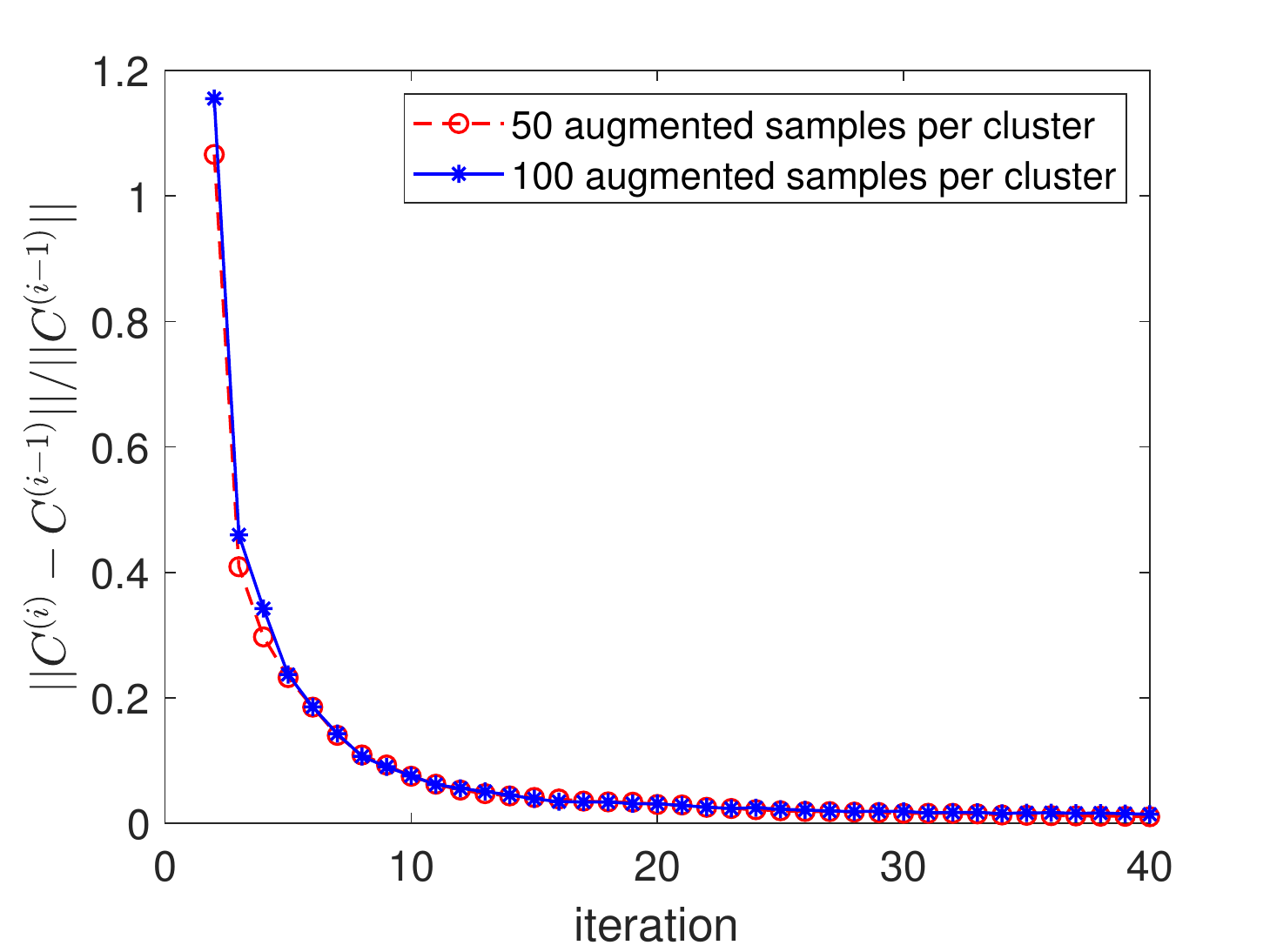}}
		%  \vspace{1.5cm}
		\centerline{(c) $\theta = 10^\circ$}\medskip
	\end{minipage}
	\hfill
	\begin{minipage}[b]{0.3\linewidth}
		\centering
		\centerline{\includegraphics[width=6cm]{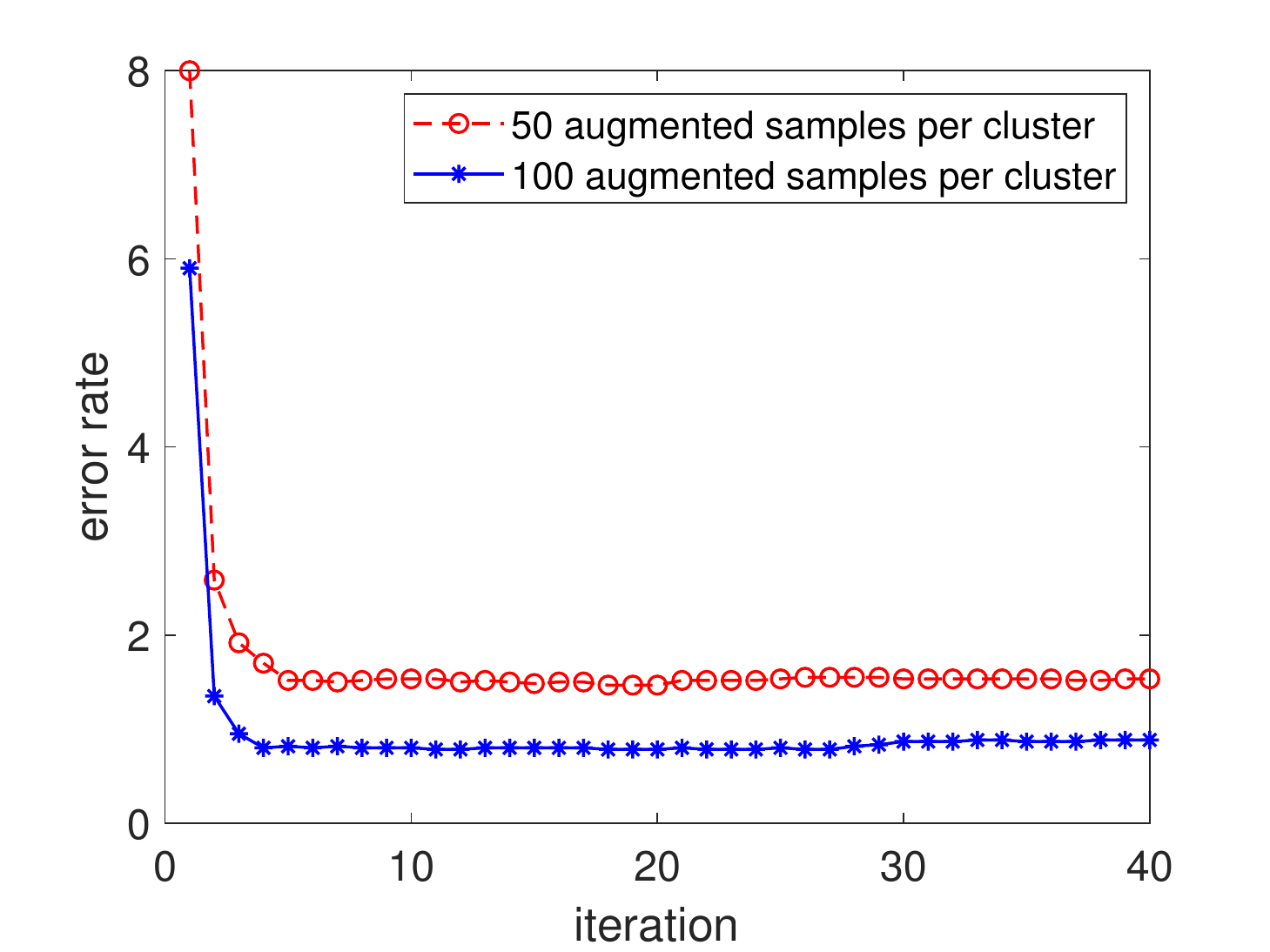}}
		%  \vspace{1.5cm}
		\centerline{(d) $\theta = 15^\circ$}\medskip
	\end{minipage}
	\hfill
	\begin{minipage}[b]{0.3\linewidth}
		\centering
		\centerline{\includegraphics[width=6cm]{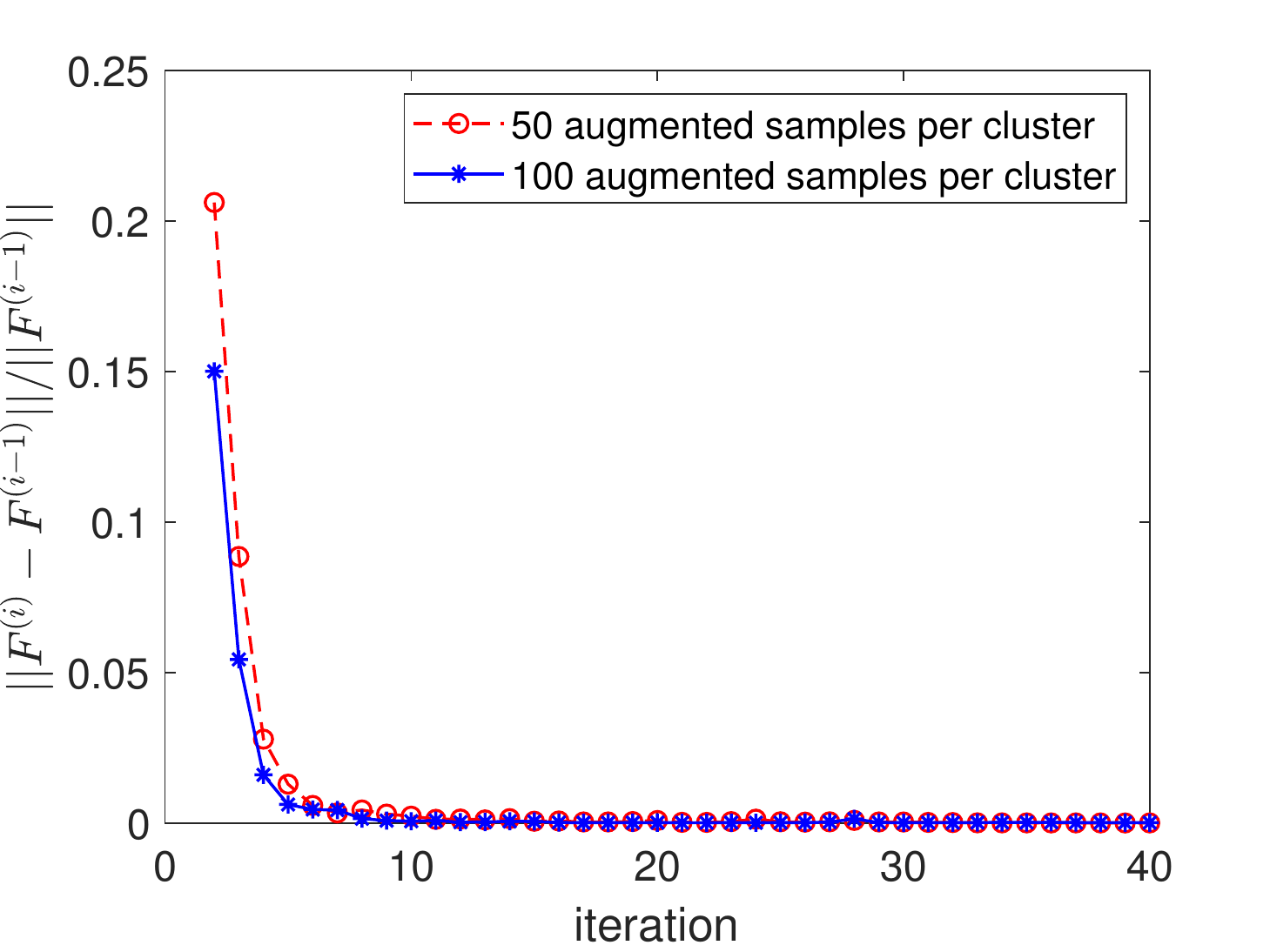}}
		%  \vspace{1.5cm}
		\centerline{(e) $\theta = 15^\circ$}\medskip
	\end{minipage}
	\hfill
	\begin{minipage}[b]{0.3\linewidth}
		\centering
		\centerline{\includegraphics[width=6cm]{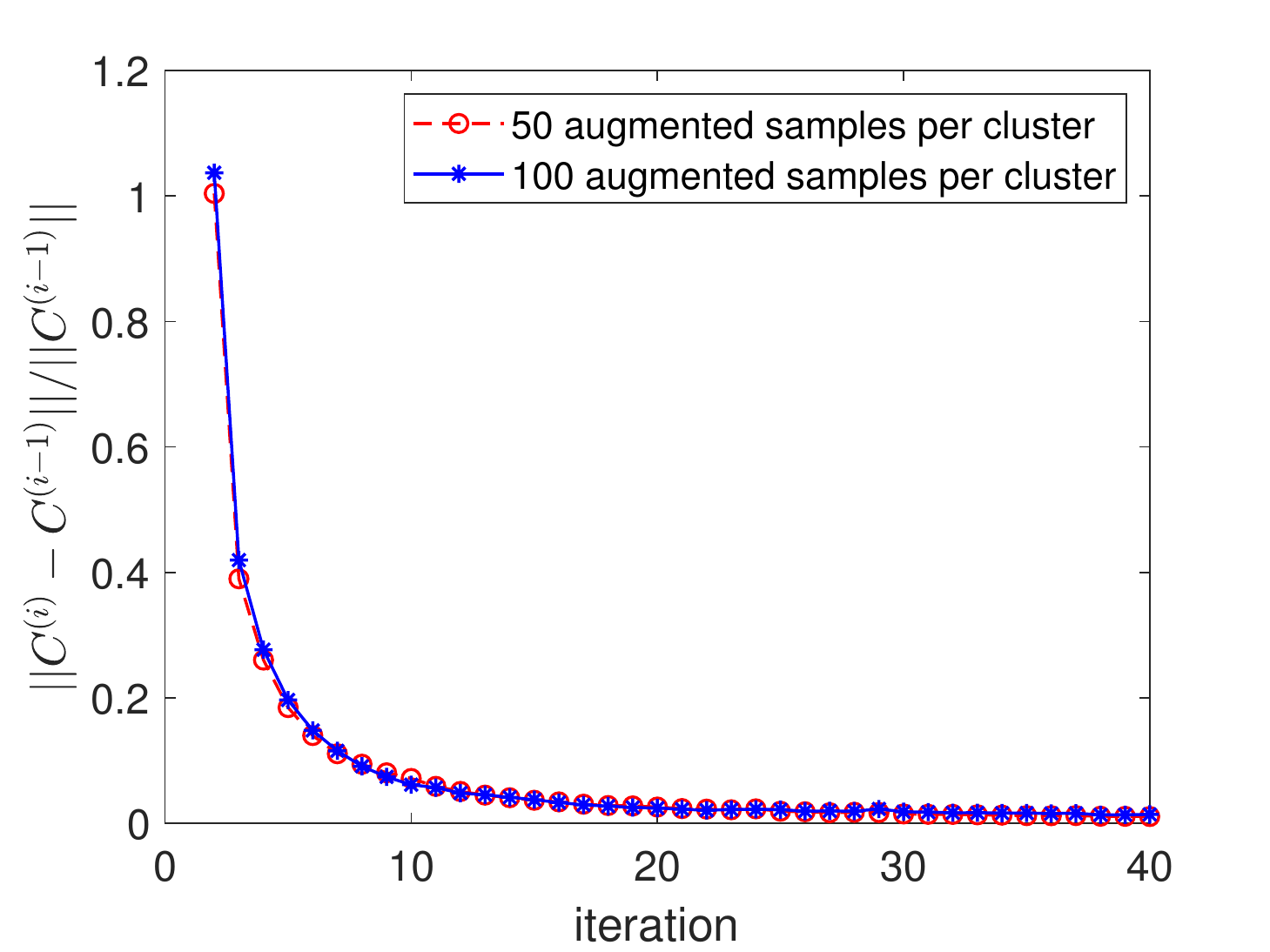}}
		%  \vspace{1.5cm}
		\centerline{(f) $\theta = 15^\circ$}\medskip
	\end{minipage}
	\caption{Convergence analysis of AS-SSC. Average of error rate, $\frac{\|F^{(i)}-F^{(i-1)}\|}{\|F^{(i-1)}\|}$ and $\frac{\|C^{(i)}-C^{(i-1)}\|}{\|C^{(i-1)}\|}$ over 100 trials for different iterations.}
	\label{convergence_err_F}
\end{figure*}

%in matlab: convergence_analysis.m
The proposed AS-SSC is an iterative approach that updates the two matrices, $C$ and $F$, alternatively. We study the convergence of AS-SSC depending on two essential factors: number of augmented samples and the angle between subspaces. Let $F^{(i)}$ and $C^{(i)}$ denote the estimated matrix $F$ and $C$ in the iteration $i$, respectively. We compute the average of error rate, $\frac{\|F^{(i)}-F^{(i-1)}\|}{\|F^{(i-1)}\|}$ and $\frac{\|C^{(i)}-C^{(i-1)}\|}{\|C^{(i-1)}\|}$ over 100 trials for different iterations and reported the results in Figure~\ref{convergence_err_F} (a)-(f). 

We observe that in all four cases, the error rate stabilizes in less than 10 iterations. Moreover, with stabilization of  $\frac{\|F^{(i)}-F^{(i-1)}\|}{\|F^{(i-1)}\|}$, the corresponding error rate does not change significantly between consequent iterations as well.

\subsubsection{Real-World data sets}

We now evaluate the performance of the proposed algorithms AkS-SSC, AkS-LSR and AkS-LRR, on the two considered real data sets, COIL-20 and MNIST.

\paragraph{COIL-20: Impact of the augmentation strategy on  AkS-SC}  

We first investigate the effect of augmentation strategies on the performance of AkS-SC algorithms. We consider three possible sets of augmentation strategies that are: \{linear interpolation\}, \{rotation, scale\}, \{flip, rotation, scale\}. Identical to the setting for the unsupervised algorithms, we set $k=20$, {$\lambda=\frac{\mu}{\max_{i\neq j}|X(:,i)^\top X(:,j)|}$, with $\mu=30$} and consider 5 random transformations for scaling and rotation. %\ngc{propagation speed is a bit odd... we can remove or use something more explicit, like: "the parameter that couples the coefficient matrix ($C$) and the labels ($F$)"} 
{The parameter $\lambda_2$, which couples the coefficient matrix ($C$) and the labels ($F$), is set to $1$.} We consider three values for the number of labeled samples, \{4,6,10\}, and randomly select the labeled samples over 10 trials. 
For the linear interpolation augmentation strategy, we set the number of augmented samples for each class as 720, which is equal to the total number of augmented samples for each class by applying 5 random rotations and 5 random scaling on each sample (for a fair comparison) and use random nonnegative weightings for the  pairwise linear combinations. 
Table~\ref{tab:coil:as-sc} reports the average error rate and NMI of AkS-SSC, AkS-LRR and AkS-LSR for each  of augmentation strategy. 
\begin{center}
	\begin{table}[!ht]
		\begin{center} \small
			\caption{Evaluation of AkS-SC on COIL-20  with respect to different augmentation strategies.}
			\label{tab:coil:as-sc}  
			\begin{tabular}{c||c|c|ccc}
				\hline
				\# Labeled & Augmentation strategy & Evaluation & AkS-SSC & AkS-LRR & AkS-LSR \\ \hline
				\multirow{6}{*}{4}& \multirow{3}{*}{\{Linear Interpolation\}} & err & 7.70$\pm$1.74 & 8.79$\pm$1.97 & 10.63$\pm$0.64 \\
				& & NMI & 94.85$\pm$0.81 & 93.97$\pm$0.91 & 91.96$\pm$0.64 \\\cline{2-6}
				& \multirow{3}{*}{\{rotation,scale\}} & err & 0$\pm$0 & 0$\pm$0 & 0.36$\pm$0.50\\
				& & NMI & 100$\pm$0 & 100$\pm$0 & 99.62$\pm$0.53 \\\cline{2-6}
				& \multirow{3}{*}{\{flip,rotation,scale\}} & err & 0$\pm$0 & 0$\pm$0 & 0.16$\pm$0.37 \\
				& & NMI & 100$\pm$0 & 100$\pm$0 & 99.84$\pm$0.34 \\\hline \hline
				\multirow{6}{*}{6} & \multirow{3}{*}{\{Linear Interpolation\}} & err & 5.20$\pm$1.175 & 5.94$\pm$0.79& 8.58$\pm$1.26 \\
				& & NMI & 96.01$\pm$96.01 & 95.83$\pm$0.58 & 93.98$\pm$1.00 \\\cline{2-6}
				& \multirow{3}{*}{\{rotation,scale\}} & err & 0$\pm$0 & 0$\pm$0 & 0$\pm$0 \\
				& & NMI & 100$\pm$0 & 100$\pm$0 & 100$\pm$0 \\\cline{2-6}
				& \multirow{3}{*}{\{flip,rotation,scale\}} & err & 0$\pm$0 & 0$\pm$0 & 0.11$\pm$0.24\\
				& & NMI & 100$\pm$0 & 100$\pm$0 & 99.88$\pm$0.26 \\\hline \hline
				\multirow{6}{*}{10} & \multirow{3}{*}{\{Linear Interpolation\}} & err & 1.65$\pm$0.84 & 3.30$\pm$0.61 & 4.77$\pm$0.66\\
				& & NMI & 98.50$\pm$0.61 & 97.23$\pm$0.38 & 95.82$\pm$0.20\\\cline{2-6}
				& \multirow{3}{*}{\{rotation,scale\}} & err & 0$\pm$0 & 0$\pm$0 & 0$\pm$0 \\
				& & NMI & 100$\pm$0 & 100$\pm$0 & 100$\pm$0 \\\cline{2-6}
				& \multirow{3}{*}{\{flip,rotation,scale\}} & err & 0$\pm$0 & 0$\pm$0 & 0$\pm$0 \\
				& & NMI & 100$\pm$0 & 100$\pm$0 & 100$\pm$0 \\\hline 
			\end{tabular}
		\end{center}
	\end{table}
\end{center}

We observe that in contrast to the unsupervised setting in Table~\ref{tab:coil20}, the sensitivity to the augmentation strategies is less evident and all AkS-SC algorithms have very good clustering performance (above 90\% in all cases). 
For a few labeled samples, as few as 4, 
%\ngc{This information is in Table~6, right? Actually, I do not see how many labelled samples you used for 'Linear Interpolation' in Table 5. Also, it is not clear which linear interpolation you used. These need to be clarified.} \textcolor{blue}{==The best result in this table is repeated in Table 6, this table is merely for showing the effect of augmentation strategy. The number of labeled samples are mentioned in the Table, it's the same for the linear interpolation too. I added above that I use pairwise nonnegative weighting.==}  
the clustering error rate can be reduced to zero. However, the linear interpolation augmentation strategy which is based on the linearity assumption for generating augmented samples does not perform as well as other strategies. This indicates that for COIL-20 data set, the \emph{global} linearity assumption is not a valid assumption. Furthermore, the results confirm that constructing a high-quality graph is essential for reliable label propagation.

\paragraph{COIL-20: comparing semi-supervised SC algorithms}

Table~\ref{tab:coil20-semi} compares the performance of the proposed AkS-SC algorithms with state-of-the-art semi-supervised SC algorithms. 
For a clear evaluation of the effect of augmentation on the performance of AkS-SC algorithms, the performance of the AkS-SC algorithms using LP but without any augmented samples are also reported as SSC-LP, LRR-LP and LSR-LP. The error rate of NNLRR, NNLRS, $S^3R$ and $S^2LRR$ are borrowed from~\cite{wang2018unified} (the NMI was not reported for these approaches). We observe that:
\begin{itemize}
	\item AkS-SC algorithms outperform other state-of-the-art algorithms by a large margin. This is due to the explicit impact of augmentation and implicit nonlinearity consideration using kNN-based dictionary construction. 
	
	\item The proposed iterative block coordinate descent algorithm for label propagation works well for this data set, and the kNN-based algorithms SSC-LP, LRR-LP and LSR-LP perform well on this data set.	
\end{itemize}
\begin{center}
	\begin{table}[!ht]
		\begin{center} \small
			\caption{Comparison of semi-supervised SC algorithms on the COIL-20 data set.}
			\label{tab:coil20-semi} 
			\resizebox{\textwidth}{!}{ 
				\begin{tabular}{c|c||ccc|cccc|ccc}
					\hline
					\# Labeled &Evaluation & SSC-LP & LRR-LP & LSR-LP & NNLRR & NNLRS & $S^3R$ & $S^2LRR$  & AkS-SSC & AkS-LRR & AkS-LSR \\ \hline
					\multirow{2}{*}{4} & err & 6.00$\pm$1.86 & 11.52$\pm$1.45 & 11.52$\pm$1.19 &14.55$\pm$1.35 & 16.71$\pm$2.67 & 13.36$\pm$2.68 & 23.57$\pm$2.54 & 0$\pm$0 & 0$\pm$0 & 0.16$\pm$0.37 \\ 
					& NMI & 95.57$\pm$1.01 & 92.46$\pm$1.22 & 92.19$\pm$0.72 & - & - & - & - & 100$\pm$0 & 100$\pm$0 & 99.84$\pm$0.34\\\hline
					\multirow{2}{*}{6} & err & 3.95$\pm$0.91 & 8.12$\pm$057 & 9.80$\pm$0.65 & 11.52$\pm$1.22 & 15.50$\pm$2.81 & 11.74$\pm$3.45 & 20.00$\pm$2.46 & 0$\pm$0 & 0$\pm$0 & 0.11$\pm$0.24\\ 
					& NMI & 96.50$\pm$0.59 & 94.30$\pm$0.65 & 93.16$\pm$0.36 & - & - & - & - & 100$\pm$0 & 100$\pm$0 & 99.88$\pm$0.26 \\\hline
					\multirow{2}{*}{10} & err & 0.93$\pm$0.86 & 5.01$\pm$1.40 & 7.23$\pm$0.82 & 8.67$\pm$1.24 & 10.57$\pm$2.99 & 7.18$\pm$0.32 & 15.81$\pm$2.32 & 0$\pm$0 & 0$\pm$0 & 0$\pm$0\\ 
					& NMI & 99.01$\pm$0.82 & 95.80$\pm$0.93 & 94.48$\pm$0.36 & - & - & - & - & 100$\pm$0 & 100$\pm$0 & 100$\pm$0\\\hline
			\end{tabular}}
		\end{center}
	\end{table}
\end{center}

\paragraph{MNIST: comparing semi-supervised SC algorithms}

We compare the performance of proposed scalable AkS-SC algorithms on the processed MNIST-test data set. 
For this experiment, we considered 50 samples for each digit from $[0:9]$. For the proposed semi-supervised approaches, we used the general augmentation strategies of scaling and rotation. For each sample, we applied 5 random scaling operators with the scaling parameter chosen randomly among the range $[0.8,1.2]$ and 5 random rotations with the rotation angle chosen randomly within $[-30^\circ,30^\circ]$. The parameters of all proposed approaches are set as $\lambda=\frac{\mu}{\max_{i\neq j}|X(:,i)^\top X(:,j)|}$, with $\mu=100$ and $\lambda_2=1$. We considered various number of randomly selected labeled samples and reported the average results over 10 trials in Table~\ref{tab:mnist-semi}. We observe that the proposed augmented based algorithms significantly outperform other approaches, including the kNN-based algorithms. This highlights the essential role of augmentation in improving the performance. The augmented based approaches perform similarly on this data set and the improvement in the performance was independent of the regularization function for the coefficient matrix. Hence, the augmentation is capable of significantly improving the SC performance regardless of the regularization function.
\begin{center}
	\begin{table}[!ht]
		\begin{center} \small
			\caption{Comparison of semi-supervised SC algorithms on the MNIST data set.}
			\label{tab:mnist-semi} 
			\resizebox{\textwidth}{!}{ 
				\begin{tabular}{c|c||ccc|cccc|ccc}
					\hline
					\# Labeled &Evaluation & SSC-LP & LRR-LP & LSR-LP & NNLRR & NNLRS & $S^3R$ & $S^2LRR$  & AkS-SSC & AkS-LRR & AkS-LSR \\ \hline
					\multirow{2}{*}{4} & err & 13.60$\pm$2.46 & 17.52$\pm$3.51 & 14.14$\pm$2.59 & 45.57$\pm$0.31 & - & 31.84$\pm$3.35 & 37.43$\pm$8.38 &8.04$\pm$0.65 & 7.90$\pm$0.69 & 7.98$\pm$0.62\\ 
					& NMI &77.59$\pm$3.24 & 74.24$\pm$3.13 & 76.90$\pm$3.48 & 49.52$\pm$0.30 & - &-  & - & 85.78$\pm$0.84 & 86.09$\pm$0.97 & 85.91$\pm$0.78 \\\hline
					\multirow{2}{*}{6} & err & 10.98$\pm$0.74 & 12.94$\pm$1.05 &  11.92$\pm$1.03 & 45.68$\pm$0.17 & - & 26.72$\pm$1.83 & 28.18$\pm$4.81 & 6.26$\pm$0.71 & 6.14$\pm$0.84 & 6.28$\pm$0.73 \\ 
					& NMI & 80.80$\pm$1.07 & 78.81$\pm$1.16 & 79.45$\pm$1.25 & 49.40$\pm$0.27 & - & - & - & 88.50$\pm$1.03 & 88.68$\pm$1.28 & 88.45$\pm$1.05\\\hline
					\multirow{2}{*}{10} & err & 8.24$\pm$0.84 & 8.18$\pm$1.48 & 8.28$\pm$1.20 & 46.02$\pm$0.34 & - & 22.07$\pm$1.37 & 22.82$\pm$3.00 & 5.80$\pm$0.71 & 5.72$\pm$0.68 & 5.78$\pm$0.75 \\ 
					& NMI & 85.04$\pm$1.52 & 85.32$\pm$2.22 & 84.8$\pm$1.97 & 49.10$\pm$0.35 & - & - & - & 89.28$\pm$1.18 & 89.43$\pm$1.21 & 89.39$\pm$1.26 \\\hline
			\end{tabular}}
		\end{center}
	\end{table}
\end{center}

%\ngc{I think the experimental part is too long, although everything is useful and nice. I focus on the experiments comparing to the state of the art, and putting the rest (e.g., example with the cars, effect of the parameter $k$, etc.) in the supplementary. We can discuss how to arrange that.} 

%\ngc{I suggest the following: Move Table 2, Figure 4, Table 4, Figure 8 to the supplementary, but mention in the text, e.g., our algorithms are not sensitive to the value of $k$; see the supplementary material.}

%\ngc{I like Table 5, it shows something rather nice, we could keep it in the paper.} 

%\ngc{ I have another suggestion: We could remove Figure 3 and replace it with Figure 9. Around Figure 3, we could say something like: Figure 3 displays 3-dimensional projection of the widely used COIL-20 data set which contains pictures of various objects taken from different angles; see Section~\ref{} for some detail.   Although SC has been shown very effective to cluster COIL-20 objects (see Section~\ref{} for some results), it does not follow the linearity assumption globally. Hence linear interpolation does not perform as well as instance-based methods (here, rotations, flipping, scaling of images); see Table~\ref{} in Section~\ref{} for a numerical comparison. }

\section{Conclusion}

In this paper, we have proposed a general framework for combining data augmentation with the state-of-the-art  self-expressive SC models using an enlarged dictionary; see the model~\eqref{eq:A-SC} for the unsupervised setting, and the model~\eqref{eq:AS-SC} for the semi-supervised setting.  
In particular, we have incorporated data augmentation within three representative state-of-the-art SC algorithms, namely SSC, LRR and LSR. We have provided geometric arguments explaining why data augmentation can  significantly benefit SC. 
For the data augmentation, we relied on two common  strategies, (i)~instance-based using classical data transformation functions, 
and (ii)~mixed-example based relying on the linearity of the SC problem, the former being preferable if available. 
%The advantages and disadvantages of both strategies were discussed and 
%We observed that instance-based are preferable. %carefully selected instance based augmentation methods is superior to the mixed-example strategy. 
Finally, we have illustrated on synthetic and real date sets the significant improvement data augmentation brings to SC algorithms. 
%Furthermore, we showed numerically that data augmentation can considerably improve the performance of all three representative algorithms, regardless of the regularization function. 

%The overall proposed framework was based on using augmented samples as an enlarged dictionary. We presented scalable practical approaches by constructing local based dictionaries for each sample using kNN. To further enhance the applicability, we considered the 
%Numerical results demonstrate that data augmentation can be significantly beneficial for the self-expressive based SC algorithms under both unsupervised and semi-supervised scenarios. 

An interesting and crucial research direction is to provide strategies for selecting effective augmentations, in particular in the absence of prior knowledge. In parallel, it would also be important to provide theoretical guarantees for augmented SC approaches.

\appendix

\section{Appendix}

\subsection{Optimizing the semi-supervised AS-SSC}\label{app:as-ssc}
%\section{Appendix: Optimizing the semi-supervised AS-SSC}\label{app:as-ssc}

In this section, we provide the details for solving~\eqref{eq:AS-SC} when $\mathcal{R}(C)=\|C\|_1$, the other cases are presented in the supplementary material~\ref{suppl:optAkSC}.  

We rely a two-block coordinate decent method to tackle AS-SSC. Our iterative approach alternates between updating $C$ while keeping $F$ fixed, and vice versa.

\paragraph{Updating the coefficient matrix $C$}

We use ADMM to optimizing the problem in ~\eqref{eq:AS-SC} with respect to the matrix $C$, while keeping the matrix $F$ fixed. By introducing the auxiliary matrix $A \in \mathbb{R}^{\tilde{n} \times n}$, this problem can be rewritten as:
\begin{align}
	\min_{C,A} & \ \|A\|_1 + \frac{\lambda}{2} \|X-\tilde{X}C\|_F^2 + \lambda_2 \sum_{i=1}^{\tilde{n}} \sum_{j=1}^n \|F(i,:)-F(j,:)\|_2^2 \ |A(i,j)|, \nonumber \\ 
	& \text{s.t. } \ A=C, \ \text{and } A(\Phi_j,j)=0 , \ \text{for} \ j=1,\dots,n.
\end{align}
We add the equality constraint $A=C$ into the objective function as:
\begin{align}
	\min_{C,A,\Delta} & \ \|A\|_1 + \frac{\lambda}{2} \|X-\tilde{X}C\|_F^2 + \lambda_2 \sum_{i=1}^{\tilde{n}} \sum_{j=1}^n \|F(i,:)-F(j,:)\|_2^2 \ |A(i,j)| \nonumber \\
	& + \frac{\rho}{2}\|C-A\|_F^2 + \tr(\Delta^\top (C-A)) , \nonumber \\ 
	& \text{s.t. } \ A(\Phi_j,j)=0, \ \text{for} \ j=1,\dots,n, 
\end{align}
where $\Delta \in \mathbb{R}^{\tilde{n} \times n}$ is the matrix of Lagrange multipliers. Optimizing this problem using ADMM consists of an iterative procedure with three updating steps at each iteration:
\begin{enumerate}
	\item Optimizing with respect to the matrix $C$ while keeping the rest of variables fixed. By taking the derivative with respect to $C$ and setting it to zero, we obtain:
	\[\big(\lambda \tilde{X}^\top \tilde{X} + \rho I_{\tilde{n}}\big) C = \lambda \tilde{X}^\top X + \rho A - \Delta, \]
	where $I_{\tilde{n}}$ is an identity matrix with size $\tilde{n} \times \tilde{n}$. The obtained problem is a system of linear equations.
	
	\item Optimizing with respect to the matrix $A$ with the rest of variables assumed as fixed: 
	\begin{align}\min_A & \ \|A\|_1 + \lambda_2 \sum_{i=1}^{\tilde{n}} \sum_{j=1}^n \|F(i,:)-F(j,:)\|_2^2 \ |A(i,j)| + \frac{\rho}{2} \|C-A\|_F^2 + \tr(\Delta^\top (C-A)) \nonumber \\ 
		& \text{s.t. }\ A(\Phi_j,j)=0 \ \text{for} \ j=1,\dots,n.
	\end{align}
	We rewrite the problem as:
	\begin{align} \label{admm_A_step}
		\min_A  & \ \|\bar{W} \odot A\|_1 + \frac{\rho}{2}\|A-(C+\frac{\Delta}{\rho})\|_F^2  
		\quad \text{s.t. } \ A(\Phi_j,j)=0 \ \text{for }j=1,\dots,n, 
	\end{align}
	where $\bar{W}(i,j) = 1 + \lambda_2 \|F(i,:)-F(j,:)\|_2^2$. This problem has the following closed-form solution:
	\begin{align*}
		J &= \mathcal{T}\left(C+\frac{\Delta}{\rho},\frac{\bar{W}}{\rho}\right), \ \text{and} \\
		A(i,j) & =\left\lbrace \begin{array}{ll}
			J(i,j), & \ i \notin \Phi_j\\
			0, &  \ i \in \Phi_j
		\end{array}\right. \ \text{for }i=1,\dots,\tilde{n} \ \text{ and } \ j=1,\dots,n, 
	\end{align*}
	where $\mathcal{T}$ is the soft-thresholding operator, $\mathcal{T}(a,b)= \max(0, |a|-b) \sign(a)$, which is applied  entry-wise. 
	%\[.\]
	%The operator $(\cdot)_+$ returns zero if the input is negative, and the input otherwise. The function $\sign(\cdot)$ returns the sign of the input number.
	
	\item The Lagrange multiplier is updated as follows 
	\[
	\Delta = \Delta + \rho \big(C-A\big).
	\]
\end{enumerate}
These three steps are computed iteratively till some convergence criterion is satisfied. We use the criterion of $\|A-C\|_F^2 \leq \epsilon$ and set $\epsilon = 2 \times 10^{-4}$, following~\cite{elhamifar2013sparse}.

\paragraph{Updating the estimated label matrix $F$}

Optimizing~\eqref{eq:AS-SC} with respect to the matrix $F$ requires to solve 
\begin{align}\label{eq:LP}
	\min_F \ & \sum_{i=1}^{\tilde{n}} \sum_{j=1}^{\tilde{n}} \|F(i,:)-F(j,:)\|_2^2 \ |\tilde{A}(i,j)|+ \gamma_1 \ \tr\left((F-\tilde{Y})^\top U (F-\tilde{Y})\right) \nonumber \\ & + \gamma_2 \sum_{i=1}^{\tilde{n}} \sum_{j=1}^{\tilde{n}} \|F(i,:)-F(j,:)\|_2^2 \ \tilde{S}(i,j),
\end{align}
where $\tilde{A} = \mathbb{R}^{\tilde{n} \times \tilde{n}}$ is defined as 
\[
\tilde{A} = 
\left( \begin{array}{c|c}
	C(1:n,1:n) & \frac{1}{2}C(n+1:\tilde{n},1:n)^\top \\ \hline
	\frac{1}{2}C(n+1:\tilde{n},1:n) & \textbf{0}_{\tilde{n}-n} \\
\end{array}\right). 
\]
The matrix $\textbf{0}_{\tilde{n}-n}$ is of size $\tilde{n}-n \times \tilde{n}-n$ and all entries are set to zero. The square matrix $\tilde{S}$ is constructed from the matrix $S$ similarly.  
Note that we have discarded the constraints $Fe=e$ and $F\geq 0$ because they will be satisfied automatically; see Appendix~\ref{app:F_stochastic}.  
We define the Laplacian matrices $L_{\tilde{A}}$ and $L_{\tilde{S}}$ as $L_{\tilde{A}} = D_{\tilde{A}} - \tilde{A}$ and $L_{\tilde{S}} = D_{\tilde{S}} - \tilde{S}$, where $D_{\tilde{A}}$ and $D_{\tilde{S}}$ are the diagonal matrices of the corresponding vertex degrees. Using this definition, we rewrite~\eqref{eq:LP} as 
\[
\min_F \ \tr\big(F^\top \ L_{\tilde{A}} \ F\big) + \gamma_1 \ \tr\left((F-\tilde{Y})^\top U (F-\tilde{Y})\right) + \ \gamma_2 \ \tr\big(F^\top \ L_{\tilde{S}} \ F\big).
\]
By taking the derivative of the aforementioned problem and setting it to zero, we have the following linear system of equations: 
\begin{equation} \label{eq:matrix_F}
\big(L_{\tilde{A}}+ \ \gamma_1 \ U + \gamma_2 L_{\tilde{S}} \big) \ F = \ \gamma_1 \ U \tilde{Y}.
\end{equation}

The AS-SSC algorithm is summarized in Algorithm~\ref{algo:AS-SSC}.

\begin{algorithm}[ht!]
	\caption{Iterative algorithm for augmented semi-supervised SSC (AS-SSC)  \label{algo:AS-SSC}}
	\begin{algorithmic}[1] 
		\REQUIRE 
		$X \in \mathbb{R}^{d \times n}$, parameters $\lambda$, $\lambda_2$, {$\gamma_1$ and $\gamma_2$}, label information $Y$, the augmented samples $\bar{X}$ and the set $\Phi_j$ (for $j=1,..,n$) which contains the indices of the corresponding augmented samples for each data.   
		\ENSURE Pairwise coefficient matrix $C_f$, estimated label matrix $F$.
		\medskip  
		\STATE Form the self-expressive dictionary $\tilde{X} = [X \ | \ \bar{X}] \in \mathbb{R}^{d \times \tilde{n}}$.
		\STATE Initialization: set the binary matrix $\tilde{Y}$ as an extension of $Y$ defined as: $\tilde{Y} = [Y;0_{(\tilde{n}-n)\times p}]$. Initialize the estimated labels $F$ as $\tilde{Y}$. Set $U$ and $S$ as in~\eqref{define_S} and~\eqref{define_U}, respectively. 
		\WHILE{convergence criterion is not satisfied}
		\STATE Initialization for updating the coefficient matrix $C$ using ADMM~\cite{gabay1976dual}:
		$C=A=\Delta = 0$, 
		$\mu = \frac{\lambda}{\max_{i \neq j} |x_j^\top  x_i|}$, 
		$\rho = \lambda$.
		
		\WHILE{ADMM algorithm has not converged}
		\STATE Update $C$ by optimizing the linear system of equations in \[\big(\lambda \tilde{X}^\top \tilde{X} + \rho I_{\tilde{n}}\big) C = \lambda \tilde{X}^\top X + \rho A - \Delta, \]
		
		\STATE Update $A$ by applying soft-threshoding operator as:
		\begin{align*}
			J &= \mathcal{T}\left(C+\frac{\Delta}{\rho},\frac{\bar{W}}{\rho}\right), \\
			A(i,j) & =\left\lbrace \begin{array}{ll}
				J(i,j), & \ i \notin \Phi_j\\
				0, &  \ i \in \Phi_j
			\end{array}\right. \ \text{for }i=1,\dots,\tilde{n} \ \text{ and } \ j=1,\dots,n, 
		\end{align*}
		where $\bar{W}(i,j)=1+\lambda_2 \| F(:,i)-F(:,j)\|_2^2$ for $i=1,\dots,\tilde{n}$ and $j=1,\dots,n$.
		
		\STATE  Update the Lagrange multiplier $\Delta$ by:
		\[\Delta = \Delta + \rho \big( C - A\big).\]
		\ENDWHILE
		
		\STATE Update the estimated label matrix $F$ by optimizing the linear system of equations in \[
		\big(L_{\tilde{A}}+ \ \gamma_1 \ U + \gamma_2 L_{\tilde{S}} \big) F = \ \gamma_1 \ U \tilde{Y}.
		\]
		
		\ENDWHILE
		
		\STATE Compute the squared coefficient matrix $C_f$ by $C_f(i,j)=\sum_{k \in \Omega(i)} |A(k,j)|$, for $i,j=1,\dots,n$
	\end{algorithmic}  
\end{algorithm}

\subsection{Matrix $F$ is row stochastic at every iteration} \label{app:F_stochastic}

The optimization problem for the matrix $F$ reduces to solving a system of linear equations in~\eqref{eq:matrix_F}, with the solution $F = \big(L_{\tilde{A}}+ \ \gamma_1 \ U + \gamma_2 L_{\tilde{S}} \big)^{-1} \big(\gamma_1 U\tilde{Y}\big)$. Let us show that the solution of this system, $F$, is row-stochastic, that is, $Fe=e$ and $F \geq 0$. Let us first show that $Fe=e$. Since $U$ is diagonal, and $L_{\tilde{A}}$ and $L_{\tilde{S}}$ are Laplacian matrices, 
we have 
\[
\big(L_{\tilde{A}}+ \ \gamma_1 \ U + \gamma_2 L_{\tilde{S}} \big) e \ = \gamma_1 \ U e = \gamma_1 \diag(U), 
\] 
where $e$ is the all-one vector, and $\diag(U)$ is the vector containing the diagonal entries of $U$. Therefore  
\[
\big(L_{\tilde{A}}+ \ \gamma_1 \ U + \gamma_2 L_{\tilde{S}} \big)^{-1}\big( \gamma_1 \diag(U)\big) = e . 
\]
On the other hand, the matrices $U$ and $\tilde{Y}$ are binary matrices where the nonzero entries correspond to the labeled samples. This implies that $U\tilde{Y}=\tilde{Y}$, and that  $\tilde{Y}e$ is equal to a binary vector with nonzero entries corresponding to the labeled samples which is equal to $\diag(U)$. Hence 
\[
\big(\gamma_1 U\tilde{Y}\big)e = \gamma_1 \diag(U). 
\]
Finally 
\[
Fe=\big(L_{\tilde{A}}+ \ \gamma_1 \ U + \gamma_2 L_{\tilde{S}} \big)^{-1} \big(\gamma_1 U\tilde{Y}\big)e = \big(L_{\tilde{A}}+ \ \gamma_1 \ U + \gamma_2 L_{\tilde{S}} \big)^{-1} \gamma_1 \diag(U) = e, 
\] 
so that  $F e = e$. 

To show that $F \geq 0$, observe that the matrix $G = L_{\tilde{A}} + \gamma_1 \ U + \gamma_2 L_{\tilde{S}}$ is a Stieltjes matrix, that is, a real symmetric positive definite matrix with nonpositive off-diagonal entries, since $U$ is diagonal with positive diagonal elements, and $L_{\tilde{A}}$ and $L_{\tilde{S}}$ are Laplacian matrices (hence $G$ is diagonally dominant).  
A Stieltjes matrix is necessarily an M-matrix so that its inverse is nonnegative\footnote{See, e.g., \url{https://en.wikipedia.org/wiki/Stieltjes_matrix}}. 
Note that $U$ and $\tilde{Y}$ are both nonnegative matrices, hence, the matrix $F$ is the multiplication of nonnegative matrices (namely, $G^{-1} \geq 0$ and $U\tilde{Y}$) and is therefore nonnegative as well.

\small 
\nocite{cai2010singular}
\bibliographystyle{spmpsci} 
\bibliography{refs}

\normalsize

\newpage 

\section{Supplementary Material} \label{suppl}

In this supplementary material, we provide further definitions and details for subspace preserving conditions in SSC and optimizing the unsupervised Ak-SC and semi-supervised AkS-SC problems. 

\subsection{Definitions for the subspace preserving condition for SSC} \label{suppl:defpres}

\begin{figure*}[!htbp]
	\centering
	\centerline{\includegraphics[width=10cm]{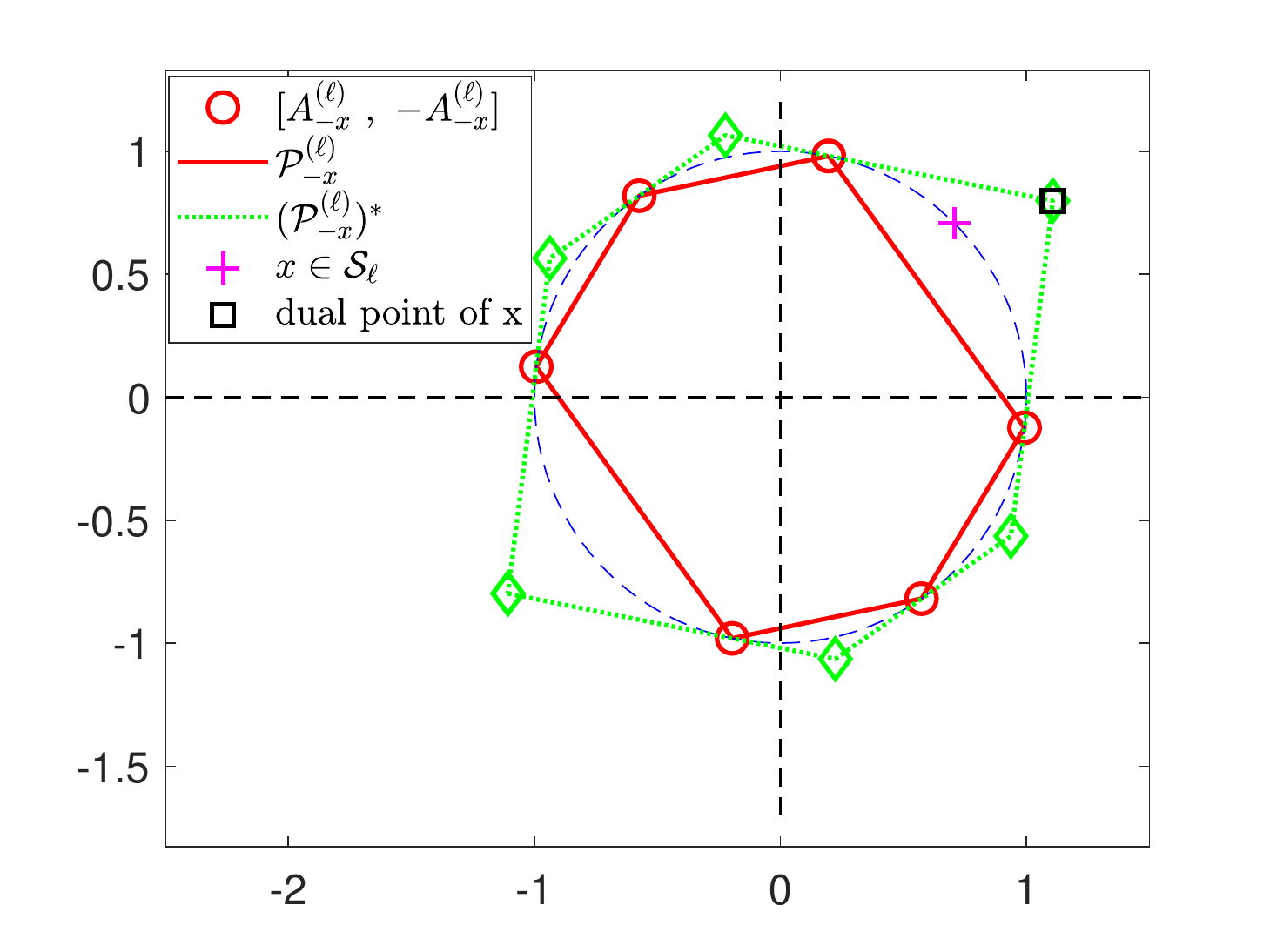}}
	\caption{The dual point corresponding to the sample $x$, shown with square and cross signs, respectively. }
	\label{dual_point}
\end{figure*}

The subspace preserving condition for SSC involves several concepts which are defined as follows: 
\begin{definition}[Dual point~\cite{li2018geometric}]
	Let $X(:,j) \in \mathcal{S}_\ell$. Suppose $X^{(\ell)}_{-j}$ are the normalized samples from the subspace $\mathcal{S}_\ell$, excluding the sample $X(:,j)$ and $U^{(\ell)} \in \mathbb{R}^{d \times d_\ell} $ is an orthonormal basis for this subspace. Let $A^{(\ell)}_{-j} = U^{(\ell)\top} X^{(\ell)}_{-j}$ be the projected coordinates of these samples on the subspace $\mathcal{S}_\ell$. The Lagrangian dual problem of \eqref{SSC} with $A^{(\ell)}$ as the expressive dictionary, for the projected sample $A(:,j)=U^{(\ell)\top}X(:,j)$ is as follows~\cite{li2018geometric,soltanolkotabi2012geometric}:
	\begin{align} \label{SSC_dual}
		W(:,j) = & \argmax_{w} w^\top A(:,j) \quad \text{ such that } \ ||A^{(\ell)\top}_{-j} w||_{\infty} \leq 1,
	\end{align}
	where $||\cdot||_\infty$ returns the the maximum absolute
	value of input vector entries. The solution $W(:,j)$ with minimum Euclidean norm is called the dual point corresponding to the sample $X(:,j)$ within subspace $\mathcal{S}_{\ell}$.
\end{definition}
This is illustrated in Figure~\ref{dual_point} for samples from an exemplary two-dimensional linear subspace. In this Figure, the projected sample $x$ and the corresponding dual point are shown with cross and square signs, respectively. The rest of projected samples and their negative counterparts, that is, $[A^{(\ell)}_{-x} \ , \ -A^{(\ell)}_{-x}]$ are shown in red circles. The convex hull of these samples, that is, $\mathcal{P}_{-x}^{\ell}$ and the corresponding dual convex hull $(\mathcal{P}_{-x}^{\ell})^*$ are shown in red solid and green dotted lines, respectively.

\begin{definition}[Dual direction~\cite{li2018geometric}]
	The dual direction $V(:,j) \in \mathbb{R}^{d}$ corresponding to the sample $X(:,j)$ from the subspace $\mathcal{S}_\ell$ is defined as:
	\begin{equation} \label{dual_direction}
		V(:,j)=U^{(\ell)}\frac{W(:,j)}{||W(:,j)||_2}.
	\end{equation}
\end{definition}

\begin{definition}[Subspace Incoherence~\cite{soltanolkotabi2012geometric}]
	The subspace incoherence of sample $X(:,j) \in \mathcal{S}_\ell$ with respect to samples from other subspaces is defined as:
	\[
	\mu(X(:,j))=\max \left\{||X^{(i)\top} \ \frac{V(:,j)}{||V(:,j)||_2}||_\infty , i =1,\dots,p, i \neq \ell \right\},
	\]
\end{definition}
where $X^{(i)}$ is a matrix which contains the samples from the $i$-th subspace.

\subsection{Optimizing the unsupervised Ak-SC algorithms} \label{suppl:optAkSC}

We have used iterative ADMM for optimizing the unsupervised problem in Ak-SC models. Depending on the regularization function $\mathcal{R}(C)=\{\|C\|_1, \|C\|_F^2, \|C\|_*\}$, we have three algorithms: Ak-SSC, Ak-LSR and Ak-LRR which are summarized in Algorithms~\ref{algo:ak-ssc},~\ref{algo:ak-lsr} and~\ref{algo:ak-lrr}, respectively.

\begin{algorithm}[ht!]
	\caption{ADMM for optimizing Ak-SSC by setting $\mathcal{R}(C)=\|C\|_1$ \label{algo:ak-ssc}}
	\begin{algorithmic}[1] 
		\REQUIRE 
		$X \in \mathbb{R}^{d \times n}$, 
		$m$ predefined augmentation strategies, parameters $\lambda$ and $k$.
		\ENSURE Pairwise coefficient matrix $C_f$.
		\medskip  
		
		\STATE Initialization: 
		Apply $m$ augmentation strategies to obtain $\tilde{X} \in \mathbb{R}^{d \times nm}$,
		$\bar{C}=A=\Delta = 0$, 
		$\mu = \frac{\lambda}{\max_{i \neq j} |x_j^T x_i|}$, 
		$\rho = \lambda$ 
		
		\FOR{each sample $X(:,j)$ in $X$}
		\STATE Set $D_j = \mathcal{N}_k(X(:,j))$ as the dictionary, $\Psi_j$ as the indices of $\mathcal{N}_k(X(:,j))$ with $\Psi_j \cap \{j+(k-1)n\}_{k=1}^{m+1} = \emptyset$.
		
		\WHILE{some convergence criterion is not met }
		
		\STATE  $A(:,j) \leftarrow \left( \mu D_j^\top D_j + \rho I_k \right)^{-1} (\mu D_j^\top X(:,j) + \rho \bar{C}(:,j) - \Delta(:,j))$ where $I_k$ is the identity matrix of dimension $k$.
		
		\STATE  $\bar{C}(:,j) \leftarrow T_{\frac{1}{\rho}} \left( A(:,j) + \Delta(:,j)/\rho \right)$ \\ where 
		$T_{\gamma} ( y ) = \max \left(0,  |y| - \gamma \right) \text{sign}(y)$ is the soft-thresholding operator. 
		
		\STATE $\Delta(:,j) = \Delta(:,j) + \rho (A(:,j)-\bar{C}(:,j))$ 
		
		\ENDWHILE
		
		\STATE Set $\tilde{C}(\Psi_j,j) = \bar{C}(:,j)$
		
		\ENDFOR
		
		\STATE Set $C_f$ as block-wise sum of submatrices in $\tilde{C}$.
	\end{algorithmic}  
\end{algorithm} 

\begin{algorithm}[ht!]
	\caption{ADMM for optimizing Ak-LSR by setting $\mathcal{R}(C)=\|C\|_F^2$ \label{algo:ak-lsr}}
	\begin{algorithmic}[1] 
		\REQUIRE 
		$X \in \mathbb{R}^{d \times n}$, 
		$m$ predefined augmentation strategies, parameters $\lambda$ and $k$.
		\ENSURE Pairwise coefficient matrix $C_f$.
		\medskip  
		
		\STATE Initialization: 
		Apply $m$ augmentation strategies to obtain $\tilde{X} \in \mathbb{R}^{d \times nm}$,
		$\bar{C}= 0$, 
		$\mu = \frac{\lambda}{\max_{i \neq j} |x_j^T x_i|}$, 
				
		\FOR{each sample $X(:,j)$ in $X$}
		\STATE Set $D_j = \mathcal{N}_k(X(:,j))$ as the dictionary, $\Psi_j$ as the indices of $\mathcal{N}_k(X(:,j))$ with ${\Psi_j \cap \{j+(k-1)n\}_{k=1}^{m+1} = \emptyset}$.
		
		\STATE  $\bar{C}(:,j) \leftarrow \left( \mu D_j^\top D_j + I_k \right)^{-1} \left(\mu D_j^\top X(:,j)\right)$ where $I_k$ is the identity matrix of dimension $k$.
		
		\STATE Set $\tilde{C}(\Psi_j,j) = \bar{C}(:,j)$
		
		\ENDFOR
		
		\STATE Set $C_f$ as block-wise sum of submatrices in $\tilde{C}$.
	\end{algorithmic}  
\end{algorithm}

\begin{algorithm}[ht!]
	\caption{ADMM for optimizing Ak-LRR by setting $\mathcal{R}(C)=\|C\|_*$ \label{algo:ak-lrr}}
	\begin{algorithmic}[1] 
		\REQUIRE 
		$X \in \mathbb{R}^{d \times n}$, 
		$m$ predefined augmentation strategies, parameters $\lambda$ and $k$.
		\ENSURE Pairwise coefficient matrix $C_f$.
		\medskip  
		
		\STATE Initialization: 
		Apply $m$ augmentation strategies to obtain $\tilde{X} \in \mathbb{R}^{d \times nm}$,
		$\tilde{C}=A=\Delta = 0$, 
		$\mu = \frac{\lambda}{\max_{i \neq j} |x_j^T x_i|}$, 
		$\rho = \lambda$ 
		
		\WHILE{some convergence criterion is not met }
		
		\FOR{each sample $X(:,j)$ in $X$}
		\STATE Set $D_j = \mathcal{N}_k(X(:,j))$ as the dictionary, $\Psi_j$ as the indices of $\mathcal{N}_k(X(:,j))$ with $\Psi_j \cap \{j+(k-1)n\}_{k=1}^{m+1} = \emptyset$.
		
		\STATE  $A(\Psi_j,j) \leftarrow \left( \mu D_j^\top D_j + \rho I_k \right)^{-1} \left(\mu D_j^\top X(:,j) + \rho \tilde{C}(\Psi_j,j) - \Delta(\Psi_j,j)\right)$ where $I_k$ is the identity matrix of dimension $k$.				
		\ENDFOR
				
		\STATE  $\tilde{C} \leftarrow SVT\left( A + \Delta/\rho \right)$ where 
		$SVT$ is the singular value thresholding operator. 
		
		\STATE $\Delta = \Delta + \rho (A-\bar{C})$ 

		\ENDWHILE

		\STATE Set $C_f$ as block-wise sum of submatrices in $\tilde{C}$.
	\end{algorithmic}  
\end{algorithm}

\subsection{Optimizing the semi-supervised AkS-SC algorithms} \label{suppl:optsemiAkS-SC} 

In this section, we consider the problem of semi-supervised AkS-SC. Similar to AS-SSC algorithm (in Appendix~A), optimizing these algorithms is carried out using an iteratice two-block coordinate decent. In each iteration, we update the estimated label matrix $F$ and coefficient matrix $C$ individually. Optimizing the matrix $F$ is identical to AS-SSC and the main difference lies in the optimization of the coefficient matrix $C$. Hence, in this section, we provide the details for updating the coefficient matrix $C$ for three scalable AkS-SC algorithms: AkS-SSC, AkS-LSR and AkS-LRR.

\subsubsection{Updating the coefficient matrix in AkS-SSC}
The AkS-SSC algorithm minimizes the following problem with respect to the coefficient matrix $C$:
\begin{align}
	\min_C & \ \|C\|_1 + \frac{\lambda}{2} \sum_{j=1}^n\|X(:,j)-\mathcal{N}_k\big(X(:,j)\big)C(\Psi_j,j)\|_F^2 + \lambda_2 \sum_{i=1}^{\tilde{n}} \sum_{j=1}^n \|F(i,:)-F(j,:\|_2^2 \ |C(i,j)| \nonumber \\ 
	& \text{s.t. } C(\Phi_j,j)=0, \ \text{for} \ j=1,\dots,n.
\end{align}
By introducing auxiliary variable $A$, this problem can be written as follows:
\begin{align} 
	\min_{A,C}  \ & \sum_{j=1}^n\|\bar{W}(:,j) \odot A(:,j)\|_1 \ + \  \sum_{j=1}^n\|X(:,j)-\mathcal{N}_k\big(X(:,j)\big)C(\Psi_j,j)\|_2^2 \ + \ \frac{\rho}{2}\sum_{j=1}^n\|C(:,j)-A(:,j)\|_2^2 \ \nonumber \\ 
	&  +\ \sum_{j=1}^n\Delta(:,j)^\top \big(C(:,j)-A(:,j)\big), \nonumber \\ 
	& \text{s.t. }\ A(\Phi_j,j)=0 \ \text{for }i=1,\dots,n, 
\end{align}
where $\bar{W}(i,j) = 1 + \lambda_2 ||F(i,:)-F(j,:)||_2^2$. In fact, optimizing this problem involves minimizing $n$ independent subproblems with respect to each column of the matrix $C$. Let $D_j = \mathcal{N}_k\big(X(:,j)\big)$, for $j=1,\dots,n$, we apply the iterative ADMM algorithm on each column of the matrices $C$, $A$ and the Lagrangian multiplier $\Delta$, separately. In each iteration, the ADMM algorithm contains three main steps for updating the $j$th column of the matrices $C$, $A$ and $\Delta$, for $j=1,\dots,n$::
\begin{enumerate}
	\item Updating the $j$-th column of the coefficient matrix $C$ by solving a scalable linear system of equations:
	\begin{align*}
		\big(\lambda D_j^\top D_j + \rho I_{k}\big) C(\Psi_j,j) = \lambda D_j^\top X(:,j) + \rho A (:,j)- \Delta(:,j),
	\end{align*}
\item Updating the $j$-th column of the auxiliary matrix $A$ while keeping other variables fixed:  	
\begin{align*} 
	\min_A  & \ \sum_{j=1}^n\|\bar{W}(:,j) \odot A(:,j)\|_1 +  \frac{\rho}{2} \|A(:,j)-\left(C(:,j)+\frac{\Delta(:,j)}{\rho}\right)\|_2^2, \nonumber \\ 
	& \text{s.t. }\ A(\Phi_j,j)=0.
\end{align*}
This problem is minimized by the using soft-thresholding operator on each column:
\begin{align*}
	J(:,j) &= \mathcal{T}\left(C(:,j)+\frac{\Delta(:,j)}{\rho},\frac{\bar{W}(:,j)}{\rho}\right),  \\
	A(i,j) & =\left\lbrace \begin{array}{ll}
		J(i,j), & \ i \notin \Phi_j\\
		0, &  \ i \in \Phi_j
	\end{array}\right. \ \text{for }i=1,\dots,\tilde{n} \ \text{ and } \ j=1,\dots,n.
\end{align*}
\item Updating the $j$-th column of the Lagrangian multiplier as:
$\Delta(:,j) = \Delta(:,j) + \rho \big( C(:,j) - A(:,j) \big)$. 
\end{enumerate}

\subsubsection{Updating the coefficient matrix in AkS-LSR}
The optimization problem for AkS-LSR is as follows:
\begin{align}
	\min_C & \ \|C\|_F^2 + \frac{\lambda}{2} \sum_{j=1}^n\|X(:,j)-\mathcal{N}_k\big(X(:,j)\big)C(\Psi_j,j)\|_F^2 + \lambda_2 \sum_{i=1}^{\tilde{n}} \sum_{j=1}^n \|F(i,:)-F(j,:)\|_2^2 \ |C(i,j)| \nonumber \\ 
	& \text{s.t. } C(\Phi_j,j)=0, \ \text{for} \ j=1,\dots,n.
\end{align}
Minimizing AkS-LSR with $\mathcal{R}(C)=\|C\|_F^2$ using ADMM is very similar to AkS-SSC, with few minor differences. The corresponding augmented Lagrangian function is defined as:
\begin{align} 
	\min_{A,C}  \ & \sum_{j=1}^n \|C(:,j)\|_2^2 \ + \  \frac{\lambda}{2}\sum_{j=1}^n\|X(:,j)-\mathcal{N}_k\big(X(:,j)\big)C(\Psi_j,j)\|_2^2 \ + \  \sum_{j=1}^n\|\bar{W}(:,j) \odot A(:,j)\|_1 \nonumber \\ 
	& + \frac{\rho}{2}\sum_{j=1}^n\|C(:,j)-A(:,j)\|_2^2 \ +\ \sum_{j=1}^n \Delta(:,j)^\top\big(C(:,j)-A(:,j)\big), \nonumber \\ 
	& \text{s.t. }\ A(\Phi_j,j)=0 \ \text{for }i=1,\dots,n.
\end{align}
We minimize this problem for each columns of the matrices $C$, $A$ and $\Delta$ separately. The iterative process contains three main steps:
\begin{enumerate}
	\item Updating the $j$-th column of the matrix $C$:
\begin{align}
\big(\lambda D_j^\top D_j + (2+\rho) I_k\big) C(\Psi_j,j) = \lambda D_j^\top X(:,j) + \rho A(:,j) - \Delta(:,j) \qquad \text{for } j=1,\dots,n. 
\end{align}
\item Updating the $j$-th column of the matrix $A$:
\begin{align*}
	\min_A  & \ \sum_{j=1}^n\|\bar{W}(:,j) \odot A(:,j)\|_1 +  \frac{\rho}{2} \|A(:,j)-(C(:,j)+\frac{\Delta(:,j)}{\rho})\|_2^2, \nonumber \\ 
& \text{s.t. }\ A(\Phi_j,j)=0,
\end{align*}
where $\bar{W}(i,j) = \lambda_2 ||F(i,:)-F(j,:)||_2^2$. This problem is minimized by applying the soft-thresholding operator on each column:
\begin{align*}
J(:,j) &= \mathcal{T}\left(C(:,j)+\frac{\Delta(:,j)}{\rho},\frac{\bar{W}(:,j)}{\rho}\right), \\
A(i,j) & =\left\lbrace \begin{array}{ll}
	J(i,j), & \ i \notin \Phi_j\\
	0, &  \ i \in \Phi_j
\end{array}\right. \ \text{for }i=1,\dots,\tilde{n} \ \text{ and } \ j=1,\dots,n.
\end{align*}
\item Updating the $j$-th column of the Lagrangian multiplier as:
$\Delta(:,j) = \Delta(:,j) + \rho \big( C(:,j) - A(:,j) \big)$. 
\end{enumerate}

\subsubsection{Updating the coefficient matrix in  AkS-LRR}

Updating the coefficient matrix in AkS-LRR using ADMM is slightly different from the two previous optimization problems in AkS-SSC and AkS-LSR. We introduce two auxiliary matrices $A,Z \in \mathbb{R}^{\tilde{n} \times n}$ and two Lagrangian multipliers $\Delta_1,\Delta_2 \in \mathbb{R}^{\tilde{n} \times n}$ and rewrite the problem as follows: 
\begin{align} 
	\min_{C,A,Z}  \ & \|C\|_* \ + \  \frac{\lambda}{2} \sum_{j=1}^n\|X(:,j)-\mathcal{N}_k\big(X(:,j)\big)Z(\Psi_j,j)\|_2^2 \ + \  \lambda_2 \sum_{i=1}^{\tilde{n}} \sum_{j=1}^n \|F(i,:)-F(j,:)\|_2^2 \ |A(i,j)| \ \nonumber \\ +
	 & \frac{\rho}{2}\big(\|C-Z\|_F^2 \ + \ \|A-Z\|_F^2 \big) \  
	+\ \tr{(\Delta_1^\top\left(C-Z\right))} \ + \ \tr{(\Delta_2^\top\left(A-Z\right))}, \nonumber \\ 
	& \text{s.t. }\ A(\Phi_j,j)=0 \ \text{for }i=1,\dots,n.
\end{align}
Minimizing this problem using ADMM involves four main steps for updating the matrices $C,A,Z$ and the two Lagrangian multipliers $\Delta_1$ and $\Delta_2$ in each iteration:

\begin{enumerate}
	\item minimizing with respect to the matrix $C$ while keeping other variables fixed:
		
		\begin{align*} 
			\min_{C}  \ & \|C\|_* \ + \ \frac{\rho}{2}\|C-Z\|_2^2 \ + \  \tr{\left(\Delta_1^\top(C-Z)\right)}, \nonumber
		\end{align*}
	This problem has a closed-form solution using singular value thresholding operator (SVT): ${C = SVT\left( Z - \Delta_1/\rho \right)}$.
	\item Updating with respect to the matrix $Z$ by optimizing the following differentiable problem:
	\begin{align*} 
		\min_{Z}  \ & \frac{\lambda}{2} \sum_{j=1}^n\|X(:,j)-\mathcal{N}_k\left(X(:,j)\right)Z(\Psi_j,j)\|_2^2 \ + \ \frac{\rho}{2} \sum_{j=1}^n \left(\|C(:,j)-Z(;,j)\|_2^2 \ + \ \|A(:,j)-Z(:,j)\|_2^2 \right) \nonumber \\  
		+ & \sum_{j=1}^n\Delta_1(:,j)^\top\left(C(:,j)-Z(:,j)\right) \ + \ \sum_{j=1}^n\Delta_2^\top\left(A(:,j)-Z(:,j)\right) \qquad \text{for } j=1,\dots,n.
	\end{align*}
	By setting the derivative to zero, we have the following linear system of equations for updating each column of the matrix $Z$: for $j=1,2,\dots, n$, 
		\[ 
		\big(\lambda D_j^\top D_j + 2\rho I_k\big) Z(\Psi_j,j) = \lambda D_j^\top X(:,j) + \rho C(\Psi_j,j) + \rho A(\Psi_j,j) +  \Delta_1(\Psi_j,j) + \Delta_2(\Psi_j,j). 
		\]

	\item Updating with respect to the matrix $A$ reduces to the following problem:
		\begin{align*} 
			\min_{A}  \ & \|\bar{W} \odot A\|_1 \ + \ \frac{\rho}{2}\|A-Z\|_F^2 \ + \ \tr{\Delta_2^\top\left(A-Z\right)}, \nonumber \\ 
			& \text{s.t. }\ A(\Phi_j,j)=0 \ \text{for }j=1,\dots,n.
		\end{align*}
	where $\bar{W}(i,j) = \lambda_2 ||F(i,:)-F(j,:)||_2^2$. 
	This problem is minimized by the using soft-thresholding operator:
		\begin{align*}
			J &= \mathcal{T}\left(Z-\frac{\Delta_2}{\rho},\frac{\bar{W}}{\rho}\right),   \\
			A(i,j) & =\left\lbrace \begin{array}{ll}
				J(i,j), & \ i \notin \Phi_j\\
				0, &  \ i \in \Phi_j
			\end{array}\right. \ \text{for }i=1,\dots,\tilde{n} \ \text{ and } \ j=1,\dots,n.
		\end{align*}
	\item Updating the Lagrangian multipliers:
	\begin{align*}
		\Delta_1 = \Delta_1 + \rho \big( C - Z \big),  \\
		\Delta_2 = \Delta_2 + \rho \big( A - Z \big). 
	\end{align*}
\end{enumerate}

\newpage 

\subsection{Additional numerical results} \label{suppl:addnumexp} 

In this section, we provide additional numerical results for the proposed Ak-SC algorithms. 

\subsubsection{Sensitivity analysis for the synthetic dataset} \label{suppl:addnumexpsens} 

%\paragraph{The effect of $\lambda_2$ on label propagation}
% in matlab: lambda_2_effect.m
The parameter $\lambda_2$ controls the coupling between $C$ and $F$ and hence the influence of label propagation in AS-SSC. As $\lambda_2$ increases, the algorithm eliminates the links between samples with different estimated labels in fewer iterations, and hence the algorithm converges faster. However, the initial constructed graph might contain a considerable number of wrong connections (see e.g., Figure~\ref{LP_exp}(a)) and too fast propagation of labels over this \emph{noisy} graph can lead to suboptimal clusterings. 
Let us investigate the influence of $\lambda_2$ on the error rate and elimination of the wrong paths between the labeled samples, which is the primary goal of the proposed iterative approach. 

\begin{figure*}[!ht]
	\begin{minipage}[b]{0.5\linewidth}
		\centering
		\centerline{\includegraphics[width=7.5cm]{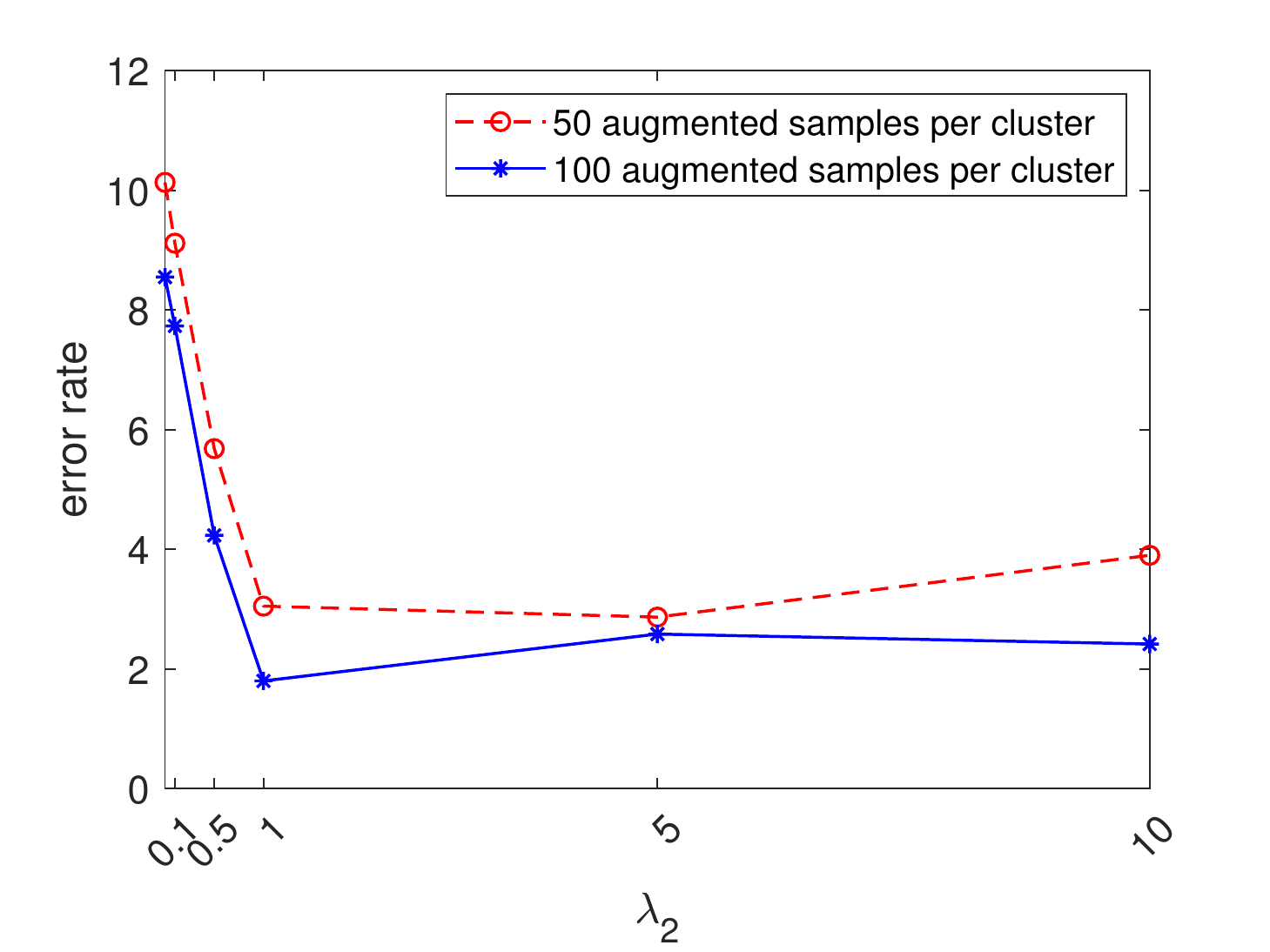}}
		%  \vspace{1.5cm}
		\centerline{(a) $\theta=10^\circ$}\medskip
	\end{minipage}
	\hfill
	\begin{minipage}[b]{0.5\linewidth}
		\centering
		\centerline{\includegraphics[width=7.5cm]{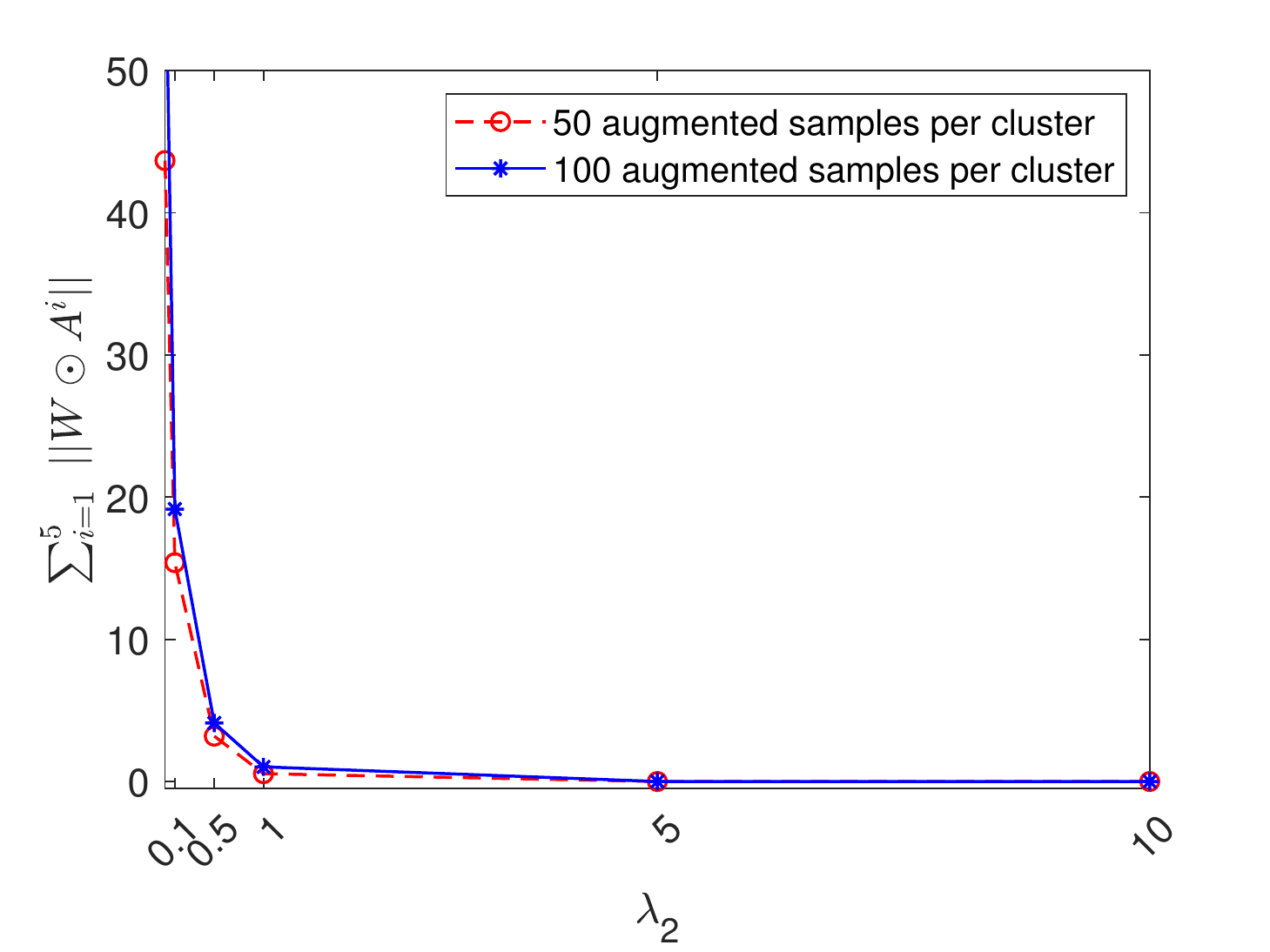}}
		%  \vspace{1.5cm}
		\centerline{(b) $\theta=10^\circ$}\medskip
	\end{minipage}
	\begin{minipage}[b]{0.5\linewidth}
		\centering
		\centerline{\includegraphics[width=7.5cm]{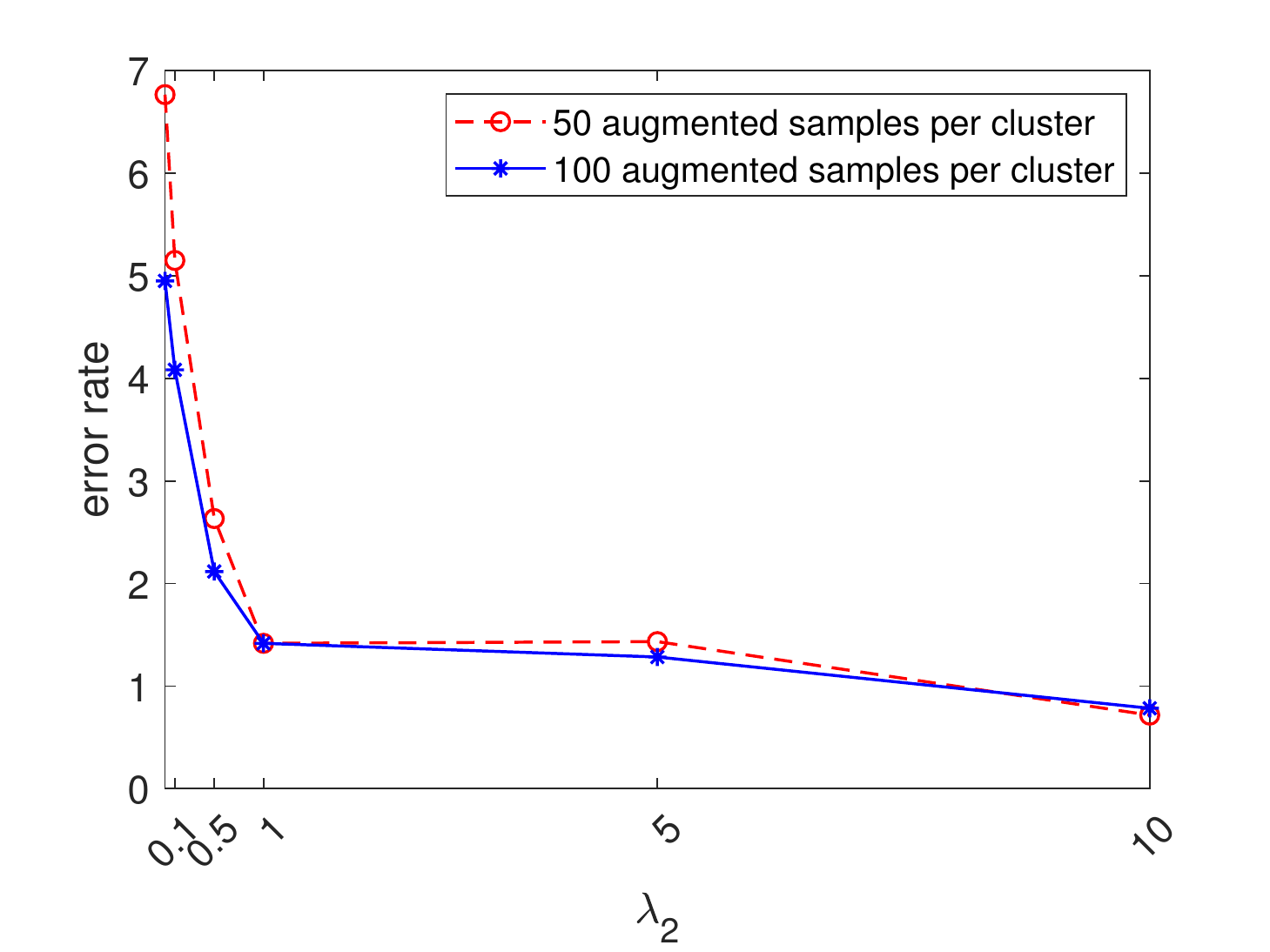}}
		%  \vspace{1.5cm}
		\centerline{(c) $\theta=15^\circ$}\medskip
	\end{minipage}
	\hfill
	\begin{minipage}[b]{0.5\linewidth}
		\centering
		\centerline{\includegraphics[width=7.5cm]{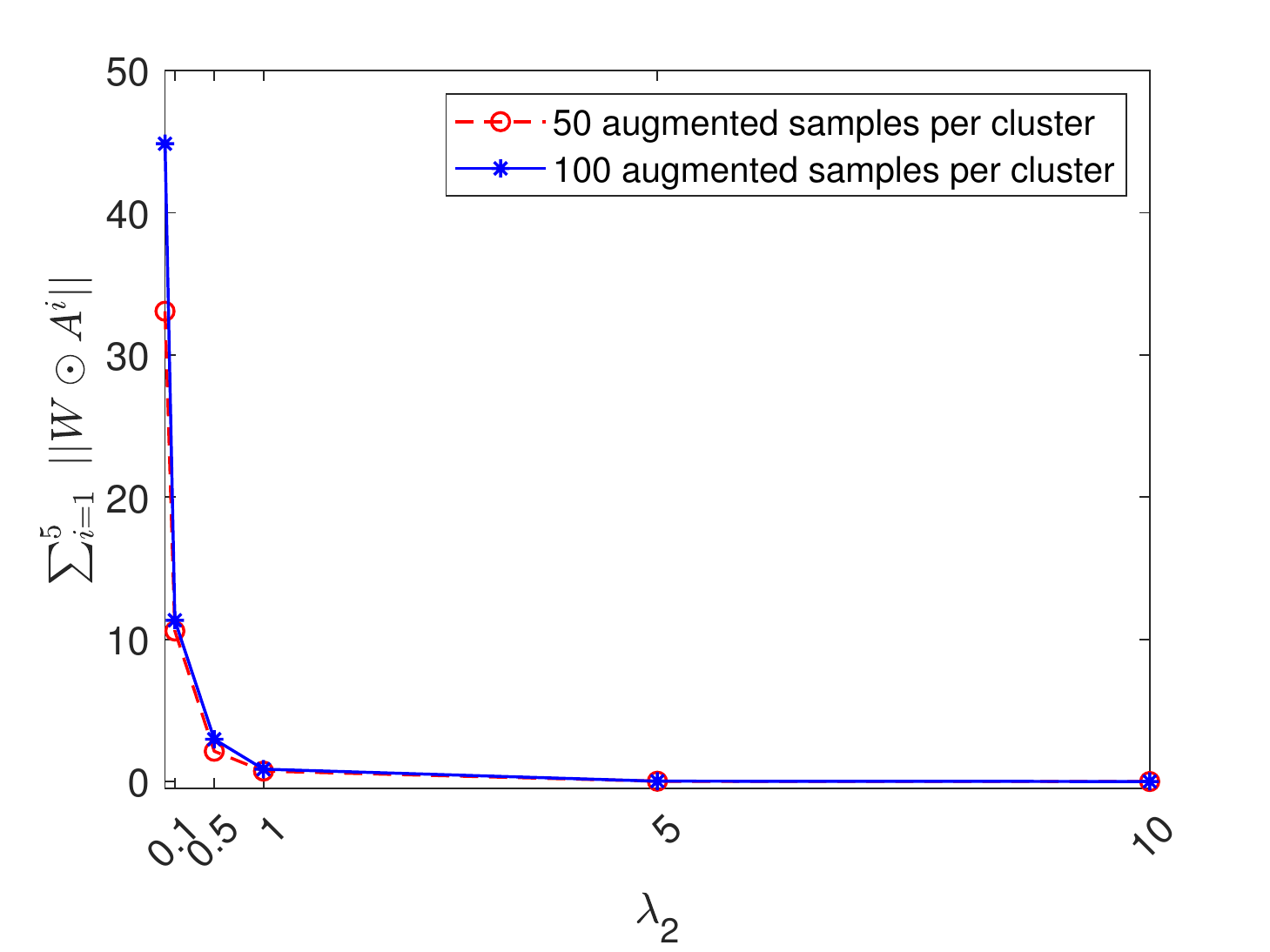}}
		%  \vspace{1.5cm}
		\centerline{(d) $\theta=15^\circ$}\medskip
	\end{minipage}
	\caption{The impact of the value of $\lambda_2 \in \{0,0.1,0.5,1,5,10\}$ on error rate and the elimination of paths between labeled samples from different clusters.
	}
	\label{lambda2_eff}
\end{figure*}

We set the number of augmented samples to either 50 or 100. We consider two cases, $\theta=\{10,15\}^\circ$. The average of error rate and $\sum_{i=1}^5 \|W \odot A^j\|$ over 100 trials are shown in Figure~\ref{lambda2_eff}. The matrix $W$ encodes the cannot-links between labeled samples from different clusters and $\sum_{i=1}^5 \|W \odot A^j\|$ quantifies the \emph{strength} of the paths with length at most equal to five between the labeled samples from different clusters (see Remark~\ref{graph_powers}). Ideally, this quantity should be equal to zero for well-separated samples from different clusters.

We observe that:
\begin{itemize}
	\item Generally, as $\lambda_2$ increases, the number of paths between labeled samples of different clusters decreases. However, this does necessarily lead to lower error rates, as this does not put any restrictions on the unlabeled samples. 
	
	\item More augmented samples provide more information on each subspace (increasing the inradius), increasing   the number of subspace preserving connections from the initial iterations, and hence leading to more robust label propagation.  
	
	\item All in all, for the synthetic data set, AS-SSC is not very sensitive to the value of $\lambda_2$ and the error rate are  similar for a large range of $\lambda_2$, namely $\lambda_2 \in [1,10]$.
\end{itemize}

\subsubsection{Sensitivity to $k$ and to augmentation strategies of Ak-SC algorithms on the COIL-20 dataset} \label{suppl:addnumexpkaug} 

Table~\ref{tab:coil:ak-sc} provides the performance of Ak-SC algorithms with respect to different values of $k$, namely $k=\{10,30,50\}$, 
and different sets of augmentation strategies, namely $\{\text{no augmentation}, \text{flip}, \text{rotation}, \text{scale}\}$. 

All augmentation strategies lead to an improvement in the performance,  for all values of $k$. %compared to the case of no augmented samples.
 Ak-SSC performs better than Ak-LRR and Ak-LSR as the value of $k$ increases. This is due to the fact that Ak-SSC enforces  sparsity, independently of the value of $k$.

\begin{center}
	\begin{table}[!ht]
		\begin{center} \small
			\caption{Evaluation of A-SC on the COIL-20 data set  with respect to different parameters.}
			\label{tab:coil:ak-sc}  
			\begin{tabular}{c||c|c|ccc}
				\hline
				$k$ & Augmentation strategy & Evaluation & Ak-SSC & Ak-LRR & Ak-LSR \\ \hline
				\multirow{8}{*}{10}& \multirow{2}{*}{none} & err & 17.84 & 15.28 & 19.72 \\
				& & NMI & 91.43 & 91.85 & 90.42 \\\cline{2-6}
				& \multirow{2}{*}{\{flip\}}& err & 16.50 & 14.62 & 16.68 \\
				& & NMI & 93.49 & 92.62 & 93.96 \\\cline{2-6}
				& \multirow{2}{*}{\{rotation, scale\}} & err & 0.01$\pm$0.04 & 0$\pm$0 & 0$\pm$0\\
				& & NMI & 99.97$\pm$0.06 & 100$\pm$0 & 100$\pm$0 \\\cline{2-6}
				& \multirow{2}{*}{\{flip, rotation, scale\}} & err & 0.01$\pm$0.04 & 0$\pm$0 & 0.02$\pm$0.06\\
				& & NMI & 99.97$\pm$0.06 & 100$\pm$0 & 99.97$\pm$0.09\\\hline \hline
				\multirow{8}{*}{30}& \multirow{2}{*}{none} & err & 22.15 & 25.28 & 28.13 \\
				& & NMI & 91.67 & 86.06 & 84.00 \\\cline{2-6}
				& \multirow{2}{*}{\{flip\}}& err & 16.32& 17.43 & 17.15\\
				& & NMI & 94.31 & 90.07 & 90.37 \\\cline{2-6}
				& \multirow{2}{*}{\{rotation, scale\}} & err & 5.19$\pm$3.58 & 12.93$\pm$2.85& 12.74$\pm$4.57 \\
				& & NMI & 98.77$\pm$0.84 & 95.70$\pm$0.72 & 95.92$\pm$0.95\\\cline{2-6}
				& \multirow{2}{*}{\{flip, rotation, scale\}} & err & 1.56$\pm$3.33 & 7.15$\pm$2.12 & 6.07$\pm$3.22\\
				& & NMI & 99.58$\pm$0.72 & 97.81$\pm$0.33 & 97.70$\pm$0.53\\\hline \hline
				\multirow{8}{*}{50}& \multirow{2}{*}{none} & err & 20.21 & 24.93 & 27.08\\
				& & NMI & 91.89 & 85.45 & 83.05 \\\cline{2-6}
				& \multirow{2}{*}{\{flip\}}& err & 16.32 & 18.82 & 18.54\\
				& & NMI & 94.25 & 88.99 & 88.42 \\\cline{2-6}
				& \multirow{2}{*}{\{rotation, scale\}} & err & 10.12$\pm$3.46 & 15.55$\pm$0.10 & 18.43$\pm$0.29\\
				& & NMI & 97.41$\pm$1.08 & 94.99$\pm$0.10 & 93.69$\pm$0.32\\\cline{2-6}
				& \multirow{2}{*}{\{flip, rotation, scale\}} & err & 7.47$\pm$0.03 & 15.13$\pm$0.13 & 17.45$\pm$2.70 \\
				& & NMI & 98.25$\pm$0.00 & 94.94$\pm$0.03 & 94.21$\pm$0.39 \\\hline
			\end{tabular}
		\end{center}
	\end{table}
\end{center}

\subsubsection{Example when Ak-SSC improves the quality of the coefficient matrix in the COIL-20 dataset} \label{suppl:addnumexpqualcoef} 

\begin{figure*}[htb]
	%\begin{minipage}[b]{0.5\linewidth}
		%\centering
		%\centerline{\includegraphics[width=6cm]{images/sample_coeff_1_coil_knn_SSC}}
		%  \vspace{1.5cm}
		%\centerline{(a) Coefficients by kNN-SSC}\medskip
	%\end{minipage}
	\begin{minipage}[b]{1\linewidth}
		\centering
		\centerline{\includegraphics[width=14cm]{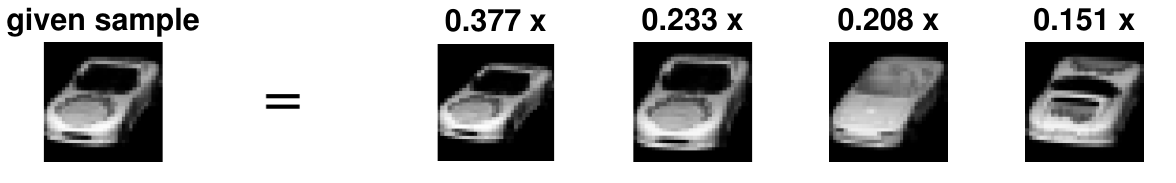}} 		
		\centerline{(a) The top four images utilized for reconstruction using kNN-SSC.} \medskip
	\end{minipage}
	%\begin{minipage}[b]{0.5\linewidth}
	%	\centering
	%	\centerline{\includegraphics[width=6cm]{images/sample_coeff_1_coil_ASSC}}
	%	\centerline{(c) Coefficients by Ak-SSC}\medskip
	%\end{minipage}
	\begin{minipage}[b]{1\linewidth}
		\centering
		\centerline{\includegraphics[width=14cm]{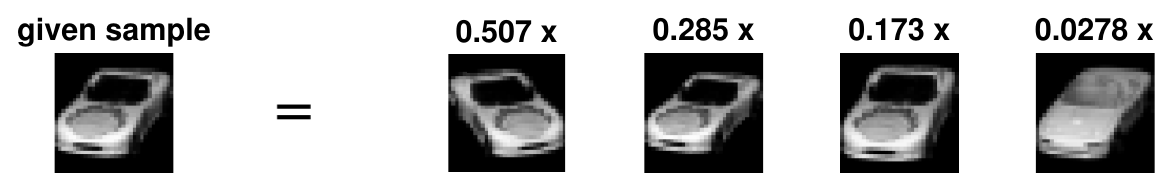}}
		\centerline{(b) The top four images utilized for reconstruction using Ak-SSC.}\medskip
	\end{minipage}
	%\vspace{1cm}
	\caption{Comparing the effect of augmentation on the coefficient matrix between kNN-SSC and Ak-SSC for a specific sample from the COIL-20 data set. The top four images used to reconstruct this sample and their corresponding coefficient values for kNN-SSC and Ak-SSC are illustrated in (a) and (b), respectively.}
	\label{sample_coil20}
\end{figure*}

We now provide an example on how Ak-SSC affects the coefficient vector for a given sample from COIL-20. The given sample is an image of a toy car and we calculate the corresponding coefficient vector using both kNN-SSC (without augmentation) and Ak-SSC (with the aforementioned augmentation strategies). 
Figure~\ref{sample_coil20} illustrates this experiment, showing the top four images corresponding to the highest absolute values in the coefficient vectors. 
We observe that kNN-SSC has a strong wrong connection with a ``visually similar" sample from a different cluster. However, the entry corresponding to this sample has significantly dropped, from 0.208 to 0.02 in Ak-SSC. Hence, augmentation has resulted in strengthening the connectivity between the samples from the same subspaces and weakening the connectivity between the samples from different subspaces. This is due to the increase in the inradius of subspaces which is helpful in clustering \emph{close} subspaces. 

\subsubsection{Effect of $k$ on the performance of Ak-SC for the MNIST dataset} \label{suppl:addnumexpkMNIST}

Table~\ref{tab:mnist:k-analys} provides the performance of Ak-SC algorithms and the corresponding kNN-based algorithms for different values of $k$. We observe that augmented algorithms perform similarly for different values of $k$, 
and all of them improve the corresponding kNN-based algorithms by a large margin. This confirms that the improvement in the performance of Ak-SC algorithms is due to the augmentation and is not sensitive to the value of $k$. 
\begin{center}
	\begin{table}[!ht]
		\begin{center} \small
			\caption{Comparing kNN-based and augmented based SC algorithms on the MNIST data set with respect to different values for $k$ for all the 10 digits.}
			\label{tab:mnist:k-analys} 
			\resizebox{\textwidth}{!}{ 
				\begin{tabular}{c|c||ccc|ccc}
					\hline
					& Evaluation & kNN-SSC & kNN-LRR & kNN-LSR  & Ak-SSC & Ak-LRR & Ak-LSR \\ \hline
					\multirow{2}{*}{k=10}& err & 24.60 & 25.60 & 30.60 & 10.00$\pm$3.44 & 9.76$\pm$3.34 & 10.00$\pm$3.43 \\
					& NMI & 75.73 & 74.42 & 72.42 &85.27$\pm$1.17 & 85.57$\pm$1.04 & 85.01$\pm$0.99 \\\hline
					\multirow{2}{*}{k=50}& err & 16.40 & 22.40 & 18.80 & 9.32$\pm$4.30 & 10.20$\pm$3.96 & 10.18$\pm$3.83 \\
					& NMI & 76.77 & 68.28 & 71.82 & 85.15$\pm$2.06 & 84.08$\pm$1.77 & 83.87$\pm$1.27 \\\hline
					\multirow{2}{*}{k=100}& err & 16.80 & 28.20 & 28.80 & 9.20$\pm$0.63 & 10.63$\pm$0.76 & 10.68$\pm$0.83 \\
					& NMI & 78.31 & 64.18 & 61.89 & 85.19$\pm$0.77 & 82.44$\pm$0.78 & 82.30$\pm$0.86 \\\hline
			\end{tabular}}
		\end{center}
	\end{table}
\end{center}

\end{document}